\documentclass{article}

     \PassOptionsToPackage{numbers, sort&compress}{natbib}
     \newif\ifsubmission
     \newif\ifpreprint
     \newif\iffinal

     \preprinttrue
     
     \ifsubmission
     \usepackage{neurips_2022}
     \fi
     
     \ifpreprint
     \usepackage[preprint]{neurips_2022}
     \fi
     
     \iffinal
     \usepackage[final]{neurips_2022}
     \fi
     
\usepackage[utf8]{inputenc} 
\usepackage[T1]{fontenc}    
\usepackage{hyperref}       
\usepackage{url}            
\usepackage{booktabs}       
\usepackage{nicefrac}       
\usepackage{microtype}      
\usepackage{xcolor}         

\usepackage{graphicx}
\usepackage{mathtools}
\usepackage{amsfonts,dsfont}
\usepackage{mathrsfs}
\usepackage{amssymb}
\usepackage{amsthm}
\usepackage{amsmath} 

\theoremstyle{plain}
\newtheorem{theorem}{Theorem}
\newtheorem{lemma}[theorem]{Lemma}
\newtheorem{proposition}[theorem]{Proposition}
\newtheorem{corollary}[theorem]{Corollary}

\PassOptionsToPackage{hyphens}{url}\usepackage{hyperref}

\hypersetup{
    colorlinks,
    linkcolor={red!50!black},
    citecolor={blue!50!black},
    urlcolor={blue!80!black}
}

\usepackage{tikz}
\usetikzlibrary{shapes,matrix,arrows,calc,positioning,fit,bayesnet,angles}

\usepackage{customCommands}

\usepackage{appendix}

\usepackage{comment}

\usepackage{multirow} 

\newcounter{constnum}
\newcommand{\const}[1]{\refstepcounter{constnum}\label{#1}}

\title{Constrained Langevin Algorithms with L-mixing External Random Variables}
\author{%
  Yuping Zheng \\
  Department of Electrical and Computer Engineering\\
  University of Minnesota, Twin Cities\\
  Minneapolis, MN 55455 \\
  \texttt{zhen0348@umn.edu} \\
   \And
   Andrew Lamperski \\
  Department of Electrical and Computer Engineering\\
  University of Minnesota, Twin Cities\\
  Minneapolis, MN 55455 \\
   \texttt{alampers@umn.edu} \\
}

\begin{document}

\maketitle
\begin{abstract}


  Langevin algorithms are gradient descent methods augmented with additive noise, and are widely used in Markov Chain Monte Carlo (MCMC) sampling, optimization, and machine learning. 
  In recent years, the non-asymptotic analysis of Langevin algorithms for non-convex learning
  has been extensively explored. For constrained problems with non-convex losses over a compact convex domain with IID data variables, the projected Langevin algorithm achieves a deviation of $O(T^{-1/4} (\log T)^{1/2})$ from its target distribution \cite{lamperski2021projected} in $1$-Wasserstein distance. 
  In this paper, we obtain a deviation of $O(T^{-1/2} \log T)$ in $1$-Wasserstein distance for non-convex losses with $L$-mixing data variables and polyhedral constraints (which are not necessarily bounded). This improves on the previous bound for constrained problems and matches the best-known bound for unconstrained problems.

\end{abstract}


\section{Introduction}

Langevin algorithms can be viewed as the simulation of Langevin dynamics  from statistical physics \citep{coffey2012langevin}. They  have been widely studied for Markov Chain Monte Carlo (MCMC) sampling \citep{roberts1996exponential}, non-convex optimization \citep{gelfand1991recursive,borkar1999strong} and machine learning \citep{welling2011bayesian}. In the statistical community, Langevin methods are used to resolve the difficulty of exact sampling from a high dimensional distribution. For non-convex optimization, the additive noise assists the algorithms to escape from local minima and saddles. Since many modern technical challenges can be cast as sampling and optimization problems, Langevin algorithms are a potential choice for the areas of adaptive control, deep neural networks, reinforcement learning, time series analysis, image processing and so on \citep{lekang2021wasserstein_accepted,barkhagen2021stochastic, chau2019stochastic}.  

\paragraph{Related Work.}
In recent years, the non-asymptotic analysis of Langevin algorithms has been extensively studied. The discussion below reviews theoretical studies of Langevin algorithms for MCMC sampling, optimization, and learning.



The non-asymptotic analysis of Langevin algorithms for approximate sampling (Langevin Monte Carlo, or LMC) began with \citep{dalalyan2012sparse,dalalyan2017theoretical}, with more recent relevant work given in \citep{durmus2017nonasymptotic,barkhagen2021stochastic,chau2019stochastic,ma2019sampling,majka2018non, wang2020fast, zou2021faster, li2021sqrt, erdogdu2022convergence, balasubramanian2022towards, nguyen2021unadjusted,lehec2021langevin, chewi2021analysis}. Most works on LMC consider log-concave target distributions, though there exists some work relaxing log-concavity \citep{majka2018non,chau2019stochastic,wang2020fast,nguyen2021unadjusted,chewi2021analysis} and smoothness of the target distribution \citep{nguyen2021unadjusted,lehec2021langevin, chewi2021analysis}. Most LMC work focuses on the unconstrained case.

Constrained problems are less studied, but a variety of works have begun to address constraints in recent years. 
The work \citep{bubeck2015finite,bubeck2018sampling} analyzes the case of log-concave distributions with samples constrained to a convex, compact set. 
Other methods derived from optimization have been introduced to handle constraints, such as  mirror descent \citep{ahn2020efficient,hsieh2018mirrored,zhang2020wasserstein,krichene2017acceleration} and proximal methods \citep{brosse2017sampling}.

Pioneering work on non-asymptotic analysis of Langevin algorithms for unconstrained non-convex optimization with IID external data variables was given in \citep{raginsky2017non}, which was motivated by machine learning applications \citep{welling2011bayesian}. Since then, numerous improvements and variations on unconstrained Langevin algorithms for non-convex optimization have been reported \citep{chau2019stochastic, xu2018global,erdogdu2018global,cheng2018sharp,chen2020stationary}. 

The work \cite{vempala2019rapid} examines the Unadjusted Langevin Algorithms without convexity assumption of the objective function and achieves a convergence guarantee in Kullback-Leibler (KL) divergence assuming that the target distribution satisfies a log-Sobolev inequlity. However, KL divergence is infinite with the deterministic initialization. To mitigate this pitfall, our work measures the convergence bound in 1-Wasserstein distance, which allows the initial condition to be deterministic.

The first  analysis of Langevin algorithms for non-convex optimization with IID external variables constrained to compact convex sets is given in \citep{lamperski2021projected}, and builds upon \citep{bubeck2015finite, bubeck2018sampling}.
However, the convergence rate derived in \citep{lamperski2021projected} is rather slow since it uses a loose result on Skorokhod problems in \citep{tanaka1979stochastic}. Recent work of \citep{sato2022convergence} obtains  $\epsilon$-suboptimality guarantees in $\tilde O(\epsilon^{-1/3})$. However, some extra work would be required to give a direct comparison with the current work, as the results in \citep{sato2022convergence} depend additionally on the spectral gap, which is not computed here.



Most convergence analyses for constrained non-convex optimization require no constraints or bounded constraint sets and IID external random variables or no external variables. In practice, the boundedness of constraint sets and the dependence of external variables do not always hold. The work \citep{chau2019stochastic} gives non-asymptotic bounds with L-mixing external variables and non-convex losses, which achieves tight performance guarantees in the unconstrained case. 
In contrast, our work gets a tight convergence bound (up to logarithmic factors) with L-mixing data streams and applies to arbitrary polyhedral constraints, which may be unbounded.

\paragraph{Contributions.} This paper focuses on the non-asymptotic analysis of constrained Langevin algorithms for a non-convex problem with L-mixing external random variables and polyhedral constraints.
We show the algorithm can achieve a deviation of $O(T^{-1/2} \log T)$ from its target distribution in 1-Wasserstein distance in the polyhedral constraint and with dependent variables. The result from \cite{chau2019stochastic} on unconstrained Langevin algorithms with L-mixing external random variables gives a deviation of $O(T^{-1/2} (\log T)^{1/2})$, and so we see that our results match, up to a factor of $(\log T)^{1/2}$. For constrained problems, our general polyhedral assumption
is not directly comparable to related work of \cite{lamperski2021projected}, which examines compact convex constraints, and \cite{sato2022convergence}, which examines bounded non-convex constraints. In the cases where the domains and random variable assumptions match (i.e. bounded polyhedra with IID external random variables or no external random variables), our paper gives the tightest bounds. In particular, this improves on the bound from \cite{lamperski2021projected}, which gives a deviation of $O(T^{-1/4} (\log T)^{1/2})$ with respect to $1$-Wasserstein distance.
%

A key enabling result in this paper is a new quantitative bound on the deviation between Skorokhod problem solutions over polyhedra, which gives a more explicit variation of an earlier non-constructive result from \citep{dupuis1991lipschitz}. Additionally, we derive a relatively simple approach to averaging out the effect of L-mixing random variables on algorithms.

\section{Problem Setup}
\subsection{Notation and terminology}
$\bbR$ denotes the set of real numbers while $\bbN$ denotes the set of non-negative integers. The Euclidean norm over $\bbR^n$ is denoted by $\|\cdot \|$. 

Random variables will be denoted in bold. If $\bx$ is a random variable, then $\bbE[\bx]$ denotes its expected value and $\cL(\bx)$ denotes its law. IID stands for independent, identically distributed.
The indicator function is denoted by $\indic$. If $P$ and $Q$ are two probability measures over $\bbR^n$, then the $1$-Wasserstein distance between them with respect to the Euclidean norm   is denoted by $W_1(P,Q)$. 

The $1$-Wasserstein distance is defined as:
\begin{equation}
\nonumber
 W_{1}(P,Q) = \inf_{\Gamma \in \mathfrak{C}(P,Q)}  \int_{\cK \times \cK} \|x-y \| d \Gamma(x,y)
\end{equation}   
where $\mathfrak{C}$ is the couplings between $P$ and $Q$.

Let $\cK$ be a convex set. (In this paper, we will assume that $\cK$ is polyhedral with $0$ in its interior.)
 The boundary of $\cK$ is denoted by $\partial \cK$. The normal cone of $\cK$ at a point $x$ is denoted by $N_{\cK}(x)$. The convex projection onto $\cK$ is denoted by $\Pi_{\cK}$. 
 
Let $\cZ$ denote the domain of the external random variables $\bz_k$.

If $\cF$ and $\cG$ are  $\sigma$-algebras, let $\cF\lor\cG$ denote the $\sigma$-algebra generated by the union of $\cF$ and $\cG$.

\subsection{Constrained Langevin algorithm}

For integers $k$ let  $\hat \bw_k \sim\cN(0,I)$ be IID Gaussian random variables and let $\bz_k$ be an L-mixing process whose properties will be described later. Assume that $\bz_i$ is independent of $\hat \bw_j$ for all $i,j\in\bbN$.

Assume that the initial value of $\bx_0\in\cK$ is independent of $\bz_i$ and $\hat \bw_j$. 
Then
the constrained Langevin algorithm has the form:
\begin{equation}
  \label{eq:projectedLangevin}
\bx_{k+1} = \Pi_{\cK}\left(\bx_k -\eta \nabla_x f(\bx_k,\bz_k) +
  \sqrt{\frac{2\eta}{\beta}} \hat\bw_{k}\right),
\end{equation}
with $k$ an integer. Here $\eta>0$ is the step size parameter and $\beta >0$ is the inverse temperature parameter. In the learning context, $f(\bx,\bz)$ is the objective function where $\bx$ are the parameters we aim to learn and $\bz$ is a training data point.

\subsection{L-mixing processes}
\label{ss:L-mixingAssumption}
In this paper, we assume that $\bz_k$ is a sequence of external data variables. The class of $L$-mixing processes was introduced in \citep{gerencser1989class} for applications in system identification and time-series analysis, and gives a means to quantitatively measure how the dependencies between the $\bz_k$ decay over time. 
Formally, $L$-mixing requires two components: 1) M-boundedness, which specifies a global bound on the moments and 2) a measure of the decay of influence over time.

A discrete-time stochastic processes $\bz_k$ is M-bounded if for all $m \ge 1$
\begin{equation}
\label{eq:Mbounded}
	\cM_m(\bz) = \sup_{k\ge 0 } \bbE^{1/m}\left[\|\bz_k\|^{m} \right]<\infty.
\end{equation}

Let $\cF_k$ be an increasing family of $\sigma$-algebras such that $\bz_k$ is $\cF_k$-measurable and $\cF_k^+$ be a decreasing family of $\sigma$-algebras such that $\cF_k$ and $\cF_k^+$ are independent for all $k\ge 0$.
Then, the process $\bz_k$ is L-mixing with respect to $\left( \left( \cF_k\right), \left( \cF_k^+ \right)\right)$ if it is M-bounded and
\begin{subequations}
  \label{eq:LMixing}
\begin{equation}
  \label{eq:totalInfluence}
	\Psi_m(\bz) = \sum_{\tau=0}^\infty \psi_m(\tau,\bz) <\infty 
\end{equation}
with
\begin{equation}
  \label{eq:influenceTau}
	\psi_m(\tau,\bz) = \sup _{k\ge \tau}\bbE^{1/m}\left[\left\|\bz_k - \bbE\left[\bz_k\vert\cF_{k-\tau}^+ \right]\right\|^m\right].
      \end{equation}
\end{subequations}

For a concrete example, consider the order-1 autoregressive model:
\begin{align}
\label{eq:ARmodel1}
    \bz_{k+1} = \alpha \bz_{k} + \boldsymbol{\xi}_{k+1}
\end{align}
where $\alpha$ is a constant with $\left| \alpha \right|<1$ and for all  $k\in \bbZ$, $\boldsymbol{\xi}_k$ are IID standard Gaussian random variables and $\bz_k \in \cZ$, where $\cZ=\bbR$ in this case. It can be observed from \eqref{eq:ARmodel1} that 
\begin{align}
\label{eq:ARmodel2}
    \bz_k = \sum_{j=0}^\infty \alpha^j \boldsymbol{\xi}_{k-j}.
\end{align}

Then, if we specify $\cF_k = \sigma \{ \bxi_i: i\le k\}$ and $\cF_k^+ = \sigma \{ \bxi_i: i> k\}$, it can be verified that $\bz_k$ satisfies \eqref{eq:Mbounded} and \eqref{eq:LMixing} and so is an L-mixing process.
%

\subsection{Assumptions}
\label{ss:assumptions}

We assume that  $\nabla_x f(x,z)$ is $\ell$-Lipschitz in both $x$ and $z$. In particular, this implies that $\|\nabla_x f(x_1,z)-\nabla_x f(x_2,z)\|\le \ell \|x_1-x_2\|$ and $\|\nabla_x f(x,z_1)-\nabla f(x,z_2)\|\le \ell \|z_1-z_2\|$. 

We assume that $\bz_t$ is a stationary $L$-mixing process, and let $\bar{f}(x)=\bbE[f(x,\bz_t)]$ denote the function which averages $f(x,\bz_t)$ with respect to $\bz_t$.

Further, we assume that $\bar{f}(x)$ is $\mu$-strongly convex outside a ball of radius $R>0$, i.e. $(x_1 - x_2)^\top\left( \nabla \bar{f}(x_1) - \nabla \bar{f}(x_2)\right) \ge \mu \left\| x_1 -x_2\right\|^2$ for all $ x_1,x_2 \in \cK$ such that $\| x_1-x_2\| \ge R$.

We assume that the initial second moment is bounded above as $\bbE[\|\bx_0\|^2]\le \varsigma < \infty$. 

Throughout the paper, $\cK$ will denote a polyhedral subset of $\bbR^n$ with $0$ in its interior.



\section{Main results}
\label{sec:results}

\subsection{Convergence of the law of the iterates}

For $\bar f$ defined above, the associated Gibbs measure is defined by:

\begin{equation}
  \label{eq:gibbs}
  \pi_{\beta \bar f}(A) = \frac{\int_{A\cap \cK} e^{-\beta \bar f(x)} dx}{\int_{\cK} e^{-\beta \bar f(x)} dx}.
\end{equation}

The main result of this paper is stated next:

\const{contraction_const1}
\newcommand{\contractionConstOneVal}{2\varphi(R)^{-1} \sqrt{   \frac{2}{\mu} c_{\ref{LyapunovConst}}}}
\const{contraction_const2}
\newcommand{\contractionConstTwoVal}{ 4\varphi(R)^{-1}}
\const{error_polyhedron1}
\newcommand{\errorPolyhedronOneVal}{\left(c_{\ref{AtoC1}}+c_{\ref{BoundCtoD2}}\sqrt{c_{\ref{AlgBound}}}+ \sqrt{2} c_{\ref{BoundCtoD3}} \right) e^{\ell}\left( 1+  \frac{2\varphi(R)^{-1}}{1-e^{-{a}/2}}\right) }
\const{error_polyhedron2}
\newcommand{\errorPolyhedronTwoVal}{ c_{\ref{BoundCtoD2}} e^{\ell}\left( 1+  \frac{2\varphi(R)^{-1}}{1-e^{-{a}/2}}\right)}

\begin{theorem}
  \label{thm:nonconvexLangevin}
  Assume that $\eta \le \min\left\{\frac{1}{4},\frac{\mu}{4\ell^2}\right\}$, $\cK$ is a polyhedron with $0$ in its interior, $\bx_0\in\cK$,  and $\bbE[\|\bx_0\|^2]\le \varsigma$.
  There are constants $a$, $c_{\ref{contraction_const1}}$, $c_{\ref{contraction_const2}}$,  $c_{\ref{error_polyhedron1}}$, and $c_{\ref{error_polyhedron2}}$ such that
 the following bound holds for all integers $k\ge 4$:
\begin{equation*}
  \label{eq:mainBound}
  W_1(\cL(\bx_k), \pi_{\beta \bar{f}}) \le
  (c_{\ref{contraction_const1}} + c_{\ref{contraction_const2}} \sqrt{\varsigma}) e^{-\eta a k}
  +
   (c_{\ref{error_polyhedron1}} + c_{\ref{error_polyhedron2}}\sqrt{\varsigma})\sqrt{\eta\log(\eta^{-1})}.
\end{equation*}
In particular, if $\eta = \frac{\log T}{2aT}$, $T\ge 4$ and $T \ge e^{2a}$, then
\begin{equation*}
  \label{eq:mainBound2}
  W_1(\cL(\bx_T), \pi_{\beta \bar{f}}) \le
  \left( c_{\ref{contraction_const1}} + c_{\ref{contraction_const2}} \sqrt{\varsigma} + \frac{c_{\ref{error_polyhedron1}} + c_{\ref{error_polyhedron2}}\sqrt{\varsigma}}{(2a)^{1/2}} \right)  T^{-1/2} \log T.
\end{equation*}
Furthermore, the constants, $c_{\ref{contraction_const1}}, c_{\ref{contraction_const2}}, c_{\ref{error_polyhedron1}}$, and $c_{\ref{error_polyhedron2}}$ are $O(n)$ with respect to the dimension of $\bx_k$, and $O(e^{\ell \beta R^2/2})$ with respect to the inverse temperature, $\beta$. And for all $\beta>0$, $a \ge \frac{2}{\frac{\beta R^2}{2}+\frac{16}{\mu}}  e^{-\frac{\beta \ell R^2}{4}}$.
\end{theorem}

The constants depend on  the dimension of $\bx_k$, $n$, the noise parameter, $\beta$, the Lipschitz constant, $\ell$, the strong convexity constant $\mu$, the variance bound of the initial states, $\varsigma$, and some geometric properties of the polyhedron, $\cK$. 

The constants shown in Theorem~\ref{thm:nonconvexLangevin} are described  explicitly in Appendix~\ref{app:constants}. 
\subsection{Auxiliary processes for convergence analysis}
\label{ss:processes}

Similar to the previous analyses of Langevin methods, e.g. \citep{raginsky2017non, bubeck2018sampling,chau2019stochastic,lamperski2021projected}, the proof of Theorem~\ref{thm:nonconvexLangevin} uses a collection of auxiliary processes fitting between the algorithms iterates from (\ref{eq:projectedLangevin}) and a stationary distribution given by (\ref{eq:gibbs}).

The algorithm and a variation in which the $\bz_t$ variables are averaged out are respectively given by:
\begin{subequations}
  \begin{align}
    \label{eq:algorithmA}
    \bx_{t+1}^A &= \Pi_{\cK}\left(\bx_t^A-\eta \nabla_x f(\bx_t^A,\bz_t)+\sqrt{\frac{2\eta}{\beta}} \hat \bw_t  \right) \\
    \label{eq:averagedM}
    \bx_{t+1}^M&=  \Pi_{\cK}\left(\bx_t^M-\eta \nabla_x \bar f(\bx_t^M)+\sqrt{\frac{2\eta}{\beta}} \hat \bw_t  \right).  
    \end{align}
\end{subequations}


Here $\bx_t^A$ represents the \underline{A}lgorithm, while $\bx_t^M$ represents a corresponding \underline{M}ean process. 

We embed the mean process in continuous time by setting $\bx_t^M = \bx_{\floor{t}}^M$, where $\floor{t}$ indicates floor function. The Gaussian noise $\hat \bw_k$ can be realized as $\hat \bw_k = \bw_{k+1} - \bw_{k}$ where $\bw_t$ is a Brownian motion.

Let $\bx_t^C$ denote a \underline{C}ontinuous-time approximation of $\bx_t^M$ defined by
the following reflected stochastic differential equation (RSDE):
\begin{equation}
  \label{eq:AvecontinuousProjectedLangevin}
  d\bx^C_t = -\eta \nabla_x \bar{f} (\bx^C_t) dt +
  \sqrt{\frac{2\eta}{\beta}} d\bw_t - \bv_t^C d\bmu^C(t).
\end{equation}

Here $-\int_0^t \bv_s^Cd\bmu^C(s)$ is a bounded variation reflection process that ensures that $\bx_t^C\in\cK$ for all $t\ge 0$, as long as $\bx_0^C\in\cK$.  In particular, the measure $\bmu^C$ is such that $\bmu^C([0,t])$ is finite,  $\bmu^C$ supported on $\{s|\bx_s^C\in\partial \cK\}$, and $\bv_s^C\in N_{\cK}(\bx_s^C)$ where $N_{\cK}(x)$ is the normal cone of $\cK$ at $x$. Lemma~\ref{lem:skorokhodExistence} in Appendix~\ref{app:skorokhod} shows that the reflection process is uniquely defined and $\bx^C$ is the unique solution to the Skorokhod problem for the process defined by:
\begin{equation}
  \label{eq:AveContinuousY}
  \by^C_t = \bx_0^C + \sqrt{\frac{2\eta}{\beta}} \bw_t - \eta \int_0^t \nabla_x
  \bar{f}(\bx^C_s) ds.
\end{equation}
See Appendix~\ref{app:skorokhod} for more details on the Skorokhod problem.

For compact notation, we denote the Skorokhod solution for a given trajectory, $\by$, by 
$\cS(\by)$. So, the fact that $\bx^C$ is the solution to the Skorokhod
problem for $\by^C$ will be denoted succinctly by $\bx^C=\cS(\by^C)$.

The basic idea behind the proof is to utilize the triangle inequality:
\begin{align}
  W_1(\cL(\bx_k^A),\pi_{\beta f})
  \label{eq:wassersteinTriangl1} \le 
  W_1(\cL(\bx_k^A),\cL(\bx_k^C))+W_1(\cL(\bx_k^C),\pi_{\beta \bar f}). 
\end{align}
and then bound each of the terms separately.

The second term is bounded by the following lemma:
\begin{lemma}
  \label{lem:convergeToStationary}
   Assume that $\bx_0\in\cK$ and $\bbE[\|\bx_0^C\|^2] \le \varsigma$. There are positive constants $a$, $c_{\ref{contraction_const1}}$ and $c_{\ref{contraction_const2}}$ such that for all $t \ge 0$
   \begin{equation*}
     W_1(\cL(\bx_t^{C}),\pi_{\beta \bar f})\le \left( c_{\ref{contraction_const1}}   + c_{\ref{contraction_const2}}  \sqrt{\varsigma} \right) e^{-\eta a t}. 
   \end{equation*}
 \end{lemma}
 This result is based on an extension of the contraction results from Corollary~2 of \citep{eberle2016reflection} for SDEs to the case of the reflected SDEs. Appendix~\ref{sec:contraction} steps through the methodology from \citep{eberle2016reflection} in order to derive  $a$, $c_{\ref{contraction_const1}}$ and $c_{\ref{contraction_const2}}$ for our particular problem.

 Most of the novel work in the paper focuses on deriving the following bound on $W_1(\cL(\bx_k^A),\cL(\bx_k^C))$:
 \begin{lemma}
   \label{lem:AtoC}
   Assume that $\bx_0^A=\bx_0^C\in\cK$, $\bbE[\|\bx_0^C\|^2]\le \varsigma$, and $\eta \le \min\left\{\frac{1}{4},\frac{\mu}{8\ell^2}\right\}$. Then there are positive constants $c_{\ref{error_polyhedron1}}$ and $c_{\ref{error_polyhedron2}}$ such that for all integers $k\ge 0$:
   $$
   W_1(\cL(\bx_k^A),\cL(\bx_k^C))\le 
  \left( c_{\ref{error_polyhedron1}} + c_{\ref{error_polyhedron2}} \sqrt{\varsigma} \right) \sqrt{\eta\log(\eta^{-1})}. 
   $$
 \end{lemma}

 \paragraph*{Proof of Theorem~\ref{thm:nonconvexLangevin}}
 Plugging the results of Lemmas~\ref{lem:convergeToStationary} and \ref{lem:AtoC} into the triangle inequality bound from (\ref{eq:wassersteinTriangl1}) proves the first result of the theorem. Specifically, let $\eta = \frac{\log T}{2aT}$, then 
\begin{align*}
 W_1(\cL(\bx_T), \pi_{\beta \bar{f}}) &\le (c_{\ref{contraction_const1}} + c_{\ref{contraction_const2}} \sqrt{\varsigma}) T^{-1/2}
  +
   (c_{\ref{error_polyhedron1}} + c_{\ref{error_polyhedron2}}\sqrt{\varsigma})\sqrt{\frac{\log T}{2aT}\log( \frac{2aT}{\log T})}\\
   & \le (c_{\ref{contraction_const1}} + c_{\ref{contraction_const2}} \sqrt{\varsigma}) T^{-1/2} \log T
    + \frac{c_{\ref{error_polyhedron1}} +   c_{\ref{error_polyhedron2}}\sqrt{\varsigma}}{(2a)^{1/2}} T^{-1/2}\log T
     .
\end{align*} 
This gives the specific bound in the theorem.
The last inequality utilizes the fact that $\log T > 1$ for all $T \ge 4$ and $\frac{2aT}{\log T} \le T$ when $T \ge e^{2a}$.

Furthermore, we examine the bounds of the constants $c_{\ref{contraction_const1}}$, $c_{\ref{contraction_const2}}$,  $c_{\ref{error_polyhedron1}}$, $c_{\ref{error_polyhedron2}}$ and $a$ in Appendix~\ref{app:constants}, where the dependencies of the convergence guarantee on state dimension $n$ and the inverse temperature parameter, $\beta$ can be observed directly.
 \hfill$\blacksquare$

 The rest of the paper focuses on proving Lemma~\ref{lem:AtoC}. 

\subsection{Proof overview for Lemma~\ref{lem:AtoC}}
\label{ss:proofOverviewofLemmaAtoC}
This subsection describes the main ideas in the proof of Lemma~\ref{lem:AtoC}. The results highlighted here, and proved in the appendix, cover the main novel aspects of the current work. The first novelty, captured in Lemmas~\ref{lem:MeanBetween1} and \ref{lem:MeanBetween2}, is a new way to bound stochastic gradient Langevin schemes with L-mixing data from a Langevin method with the data variables averaged out. The key idea is a method for examining a collection of partially averaged processes. The second novelty is a tight quantitative bound on the deviation of discretized Langevin algorithms from their continuous-time counterparts when constrained to a polyhedron. This result is based on a new quantitative bound on Skorokhod solutions over polyhedra.

First we derive time-dependent bounds (i.e. bounds that depend on $k$) for $W_1(\cL(\bx_k^A),\cL(\bx_k^C))$ . This is achieved by introducing a collection of intermediate processes and bounding their differences. Time-uniform bounds are then achieved by exploiting contractivity properties of $\bx_t^C$.

To bound $W_1(\cL(\bx_k^A),\cL(\bx_k^C))$, we first use the triangle inequality: 
\begin{align}
  \label{eq:AtoCTriangle}
  W_1(\cL(\bx_k^A),\cL(\bx_k^C))\le 
  W_1(\cL(\bx_k^A),\cL(\bx_k^M)) + W_1(\cL(\bx_k^M),\cL(\bx_k^C)).
\end{align}

We bound $W_1(\cL(\bx_k^A),\cL(\bx_k^M))$  via a collection of auxiliary processes in which the effect of $\bz_k$ is partially averaged out. We bound $W_1(\cL(\bx_k^M),\cL(\bx_k^C))$ via a specialized discrete-time approximation of $\bx_t^C$. 

Now we construct the collection of partially averaged processes. 
Recall that $\bz_k\in\cZ$ is a stationary $L$-mixing process with respect to the $\sigma$-algebras $\cF_k$ and $\cF_k^+$.
For $k<0$, we set $\cF_k=\{\emptyset,\cZ\}$, i.e. the trivial $\sigma$-algebra.
Let $\cG_t$ be the filtration generated by the Brownian motion, $\bw_t$.

Recall that for $k\in\bbN$, we set $\hat\bw_k = \bw_{k+1}-\bw_k$. Define the following discrete-time processes:
\begin{subequations}
\begin{align}
  \label{eq:averagedIntermediate}    
    \bx_{k+1}^{M,s}&=
       \Pi_{\cK}\left(\bx_k^{M,s}-\eta \bbE[ \nabla_x f(\bx_k^{M,s},\bz_k) | \cF_{k-s}\lor \cG_k ]+\sqrt{\frac{2\eta}{\beta}} \hat\bw_k  \right)             
 \\
 \label{eq:averagedBetween}
     \bx_{k+1}^{B,s} &=   
  \Pi_{\cK}\left(\bx_k^{B,s}-\eta \bbE[ \nabla_x f(\bx_k^{M,s},\bz_k) | \cF_{k-s-1} \lor \cG_k ]   +\sqrt{\frac{2\eta}{\beta}} \hat\bw_k  \right).      
\end{align}
\end{subequations}
Assume that all initial conditions are equal. In other words, $\bx_0^A=\bx_0^M=\bx_0^{M,s}=\bx_0^{B,s}$, for all $s\ge 0$.
The iterations from (\ref{eq:averagedIntermediate}) define a family of algorithms in which the data variables are partially averaged, while $\bx_k^{B,s}$ from (\ref{eq:averagedBetween}) corresponds to an auxiliary process that fits between $\bx_k^{M,s}$ and $\bx_k^{M,s+1}$. 
(Here ``A'' stands for algorithm, ``M'' stands for mean, and ``B'' stands for between.)

Note for $s=0$, we have that $\bx_k^{M,0}=\bx_k^A$ and for $s>k$, we  have that $\bx_k^{M,s}=\bx_k^M$. So, in order to bound $W_1(\cL(\bx_k^A),\cL(\bx_k^M))$,
  it suffices to bound $W_1(\cL(\bx_k^{M,s}),\cL(\bx_k^{B,s}))$ and $W_1(\cL(\bx_k^{B,s}),\cL(\bx_k^{M,s+1}))$ for all $s\ge 0$. These bounds are achieved in the following lemmas, which are proved in Appendix~\ref{app:aveLemmas}.

\begin{lemma}
  \label{lem:MeanBetween1}
  For all $s\ge 0$ and all $k\ge 0$, the following bound holds:
  \begin{equation*}
          \nonumber
          W_1\left(\cL(\bx_k^{M,s}),\cL(\bx_k^{B,s}) \right)
          \le \bbE[\|\bx_k^{M,s}-\bx_k^{B,s}\|] 
    \le   2\ell \psi_2(s,\bz)\eta \sqrt{k} .
  \end{equation*}
\end{lemma}

\begin{lemma}
\label{lem:MeanBetween2}
For all $s\ge 0$ and all $k\ge 0$, the following bound holds
\begin{multline}
 \nonumber
  W_1\left(\cL(\bx_k^{B,s}),\cL(\bx_k^{M,s+1})\right) \le \bbE[\|\bx_k^{B,s}-\bx_k^{M,s+1}\|]  
  \le
  2\ell \psi_2(s,\bz)\eta \sqrt{k} \left( e^{\eta k \ell}-1\right).
\end{multline}
\end{lemma}

Now we define the discretized approximation of $\bx_t^C$. For any initial $\bx_0^D \in \cK$, we define the following iteration on the integers:
\begin{align*}
	\bx_{k+1}^D = \Pi_{\cK} (\bx_k^D + \by_{k+1}^C - \by_k^C) 
	= \Pi_{\cK} \left(\bx_k^D +  \int_{k}^{k+1} \nabla \bar{f} (\bx_{s}^C) ds + \sqrt{\frac{2 \eta}{\beta}} \hat\bw_{k}\right).
\end{align*}
Recall that the process $\by^C$ is defined by \eqref{eq:AveContinuousY}.

Provided that $\bx_0^D = \bx_0^C$, we have that $\bx^D = \cS(\by^D) = \cS(\cD(\by^C))$, where $\cD$ is the discretization operator that sets $\cD(x)_t = x_{\floor{t}}$ for any continuous-time trajectory $x_t$. Recall that $\cS$ is the Skorokhod solution operator.

The approximation, $\bx^D$, was utilized in \citep{bubeck2018sampling,lamperski2021projected} to bound discretization errors. 
The next lemmas show how to bound $W_1(\cL(\bx_k^C),\cL(\bx_k^D))$ and $W_1(\cL(\bx_k^M),\cL(\bx_k^D))$, respectively. In particular, Lemma~\ref{lem:C2D} is analogous to Propositions 2.4 and 3.6 of \citep{bubeck2018sampling} and Lemma 9 of \citep{lamperski2021projected}. These earlier works end up with bounds of $O(\eta^{3/4}k^{1/2}+\sqrt{\eta \log k})$. It is shown in \cite{lamperski2021projected} that such bounds can be translated into time-uniform bounds of the form $\tilde O(\eta^{1/4})$. The bound from Lemma~\ref{lem:C2D} is of the form $O(\eta k^{1/2}+\sqrt{\eta \log k})$, and we will see in the next subsection that this leads to a time-uniform bound of the form $\tilde O(\eta^{1/2})$. 

  \const{BoundCtoD1}
 \newcommand{\BoundCtoDOneVal}{(c_{\ref{diamBound}}+1)\sqrt{\frac{2}{\mu}\ell^2 c_{\ref{LyapunovConst}}  + 2 \| \nabla_x \bar{f}(0)\|^2}} 
  \const{BoundCtoD2}
  \newcommand{\BoundCtoDTwoVal}{(c_{\ref{diamBound}}+1) \sqrt{2 \ell^2}}
  \const{BoundCtoD3}
  \newcommand{\BoundCtoDThreeVal}{ (c_{\ref{diamBound}}+1) n\sqrt{\frac{8}{\beta}} }

\begin{lemma}\label{lem:C2D}
  Assume that $\cK$ is a polyhedron with $0$ in its interior.
  Assume that  $\bx_0^C = \bx_0^D \in \cK$ and that $\bbE[\|\bx_0^C\|] \le \varsigma$. There are constants, $c_{\ref{BoundCtoD1}}$, $c_{\ref{BoundCtoD2}}$ and $c_{\ref{BoundCtoD3}}$ such that for all integers $k\ge 0$, the following bound holds:
\begin{align*}
  W_1\left(\cL(\bx_k^C),\cL(\bx_k^D) \right)\le \bbE\left[\|\bx_k^C - \bx_k^D \| \right] 
  \le  \left( c_{\ref{BoundCtoD1}} + c_{\ref{BoundCtoD2}} \sqrt{\varsigma}\right)  \eta \sqrt{k} + c_{\ref{BoundCtoD3}} \sqrt{\eta \log(4k)}  .
\end{align*}

\end{lemma}

\begin{lemma}
\label{lem:M2D}
  Assume that $\cK$ is a polyhedron with $0$ in its interior.
Assume that  $\bx_0^C = \bx_0^D=\bx_0^M \in \cK$ and that $\bbE[\|\bx_0^C\|] \le \varsigma$. Then for all integers $k\ge0$, the following bound holds
\begin{align*}
  W_1\left(\cL(\bx_k^M),\cL(\bx_k^D) \right) \le \bbE\left[\|\bx_k^M - \bx_k^D \| \right]  
   \le \left(\left( c_{\ref{BoundCtoD1}} + c_{\ref{BoundCtoD2}} \sqrt{\varsigma}\right)  \eta \sqrt{k} + c_{\ref{BoundCtoD3}} \sqrt{\eta \log(4k)}  \right) \left( e^{\eta\ell k} -1 \right).
\end{align*}
\end{lemma}

We highlight that Lemma~\ref{lem:C2D} utilizes the rather tight bounds on solutions to Skorokhod problems over a polyhedral domain shown in Theorem~\ref{thm:skorokhodConst}. The derivation of such tight bounds is one of the novelties of our work. More details will be discussed in Section \ref{sec:SkorokhodTightBounnd} and Appendix {\ref{app:skorokhod}.

With all of the auxiliary processes defined and their differences, we have the following lemma, which gives a time-dependent bound on $W_1(\cL(\bx_k^A),\cL(\bx_k^C))$:
\const{AtoC1}
\newcommand{\AtoCOneVal}{c_{\ref{BoundCtoD1}}+2\ell \Psi_2(\bz) }
\begin{lemma}
  \label{lem:AtoCdependent}
  Assume that $\cK$ is a polyhedron with $0$ in its interior.
  Assume that  $\bx_0^A=\bx_0^C \in \cK$ and that $\bbE[\|\bx_0^A\|] \le \varsigma$. There are constants, $c_{\ref{BoundCtoD2}}$, $c_{\ref{BoundCtoD3}}$ and $c_{\ref{AtoC1}}$, such that for all $k\ge 0$, the following bound holds:
  \begin{align*}
  W_1(\cL(\bx_k^A),\cL(\bx_k^C)) \le \left(
    \left( c_{\ref{AtoC1}} + c_{\ref{BoundCtoD2}} \sqrt{\varsigma}\right)  \eta \sqrt{k} + c_{\ref{BoundCtoD3}} \sqrt{\eta \log(4k)}
    \right)e^{\eta \ell k}.
  \end{align*}
\end{lemma}

\paragraph{Proof of Lemma~\ref{lem:AtoCdependent}}
  Recalling that $\bx_k^{M,0}=\bx_k^A$ and $\bx_k^{M,k+1}=\bx_k^M$ and using the triangle  inequality gives:
  \begin{align}
    \nonumber
    \MoveEqLeft[0]
    W_1(\cL(\bx_k^A),\cL(\bx_k^M))
    \\
    \nonumber
    &\le
      \sum_{s=0}^{k} W_1(\cL(\bx_k^{M,s}),\cL(\bx_k^{M,s+1}))\\
    \nonumber
    &\le \
      \sum_{s=0}^{k} \left(W_1(\cL(\bx_k^{M,s}),\cL(\bx_k^{B,s})) +
      W_1(\cL(\bx_k^{B,s}),\cL(\bx_k^{M,s+1}))\right)\\
    \nonumber
    &\overset{\textrm{Lemmas~\ref{lem:MeanBetween1}~\& ~\ref{lem:MeanBetween2}}}{\le}
      \sum_{s=0}^{k}2\ell\psi_2(s,\bz)\eta \sqrt{k}e^{\eta\ell k}
    \\
    \label{eq:AtoMdependent}
    &\le 2\ell \Psi_2(\bz) \eta \sqrt{k}e^{\eta \ell k}.
  \end{align}
  Here $\psi_2(s,\bz)$ and $\Psi_2(\bz)$ are the terms that bound the decay of probabilistic dependence between the $\bz_k$ variables, as defined in \eqref{eq:LMixing}.

  Similarly, we bound
  \begin{align}
    \nonumber
    \MoveEqLeft[0]
    W_1(\cL(\bx_k^M),\cL(\bx_k^C))\\
    \nonumber
    &\le W_1(\cL(\bx_k^M),\cL(\bx_k^D))+W_1(\cL(\bx_k^D),\cL(\bx_k^C))
    \\
    \label{eq:MtoCdependent}
    &\overset{\textrm{Lemmas}~\ref{lem:C2D}~\&~\ref{lem:M2D}}{\le}
       \left( (c_{\ref{BoundCtoD1}} + c_{\ref{BoundCtoD2}} \sqrt{\varsigma}) \eta \sqrt{k} + c_{\ref{BoundCtoD3}} \sqrt{\eta \log(4k)} \right) e^{\eta\ell k}.
  \end{align}
  Plugging the bounds from (\ref{eq:AtoMdependent}) and (\ref{eq:MtoCdependent}) into (\ref{eq:AtoCTriangle}) proves the lemma, with $c_{\ref{AtoC1}} = \AtoCOneVal $.
  \hfill$\blacksquare$


The proof of Lemma~\ref{lem:AtoC} is completed by showing how the time-dependent bound from Lemma~\ref{lem:AtoCdependent} can be turned into a bound that is independent of $k$. The technique used for this step is based on ideas from \citep{chau2019stochastic}, and is shown in Appendix~\ref{app:switching}.

\section{Quantitative bounds on Skorokhod solutions over polyhedra} \label{sec:SkorokhodTightBounnd}

In this section, we present a result that enables our new bound between the continuous-time process $\bx_t^C$ and the discretized process $\bx_t^M$ when constrained to the set $\cK$ defined by:
\begin{equation}
\label{eq:halfspaceRep}
\cK = \{x | a_i^\top x \le b_i \textrm{ for } i=1,\ldots,m\},
\end{equation}
where $a_i$ are unit vectors.

As discussed in Section \ref{ss:proofOverviewofLemmaAtoC}, the bound in Lemma ~\ref{lem:C2D} improves upon the corresponding results in earlier works \cite{bubeck2018sampling,lamperski2021projected}. The improvement arises from the use of Theorem~\ref{thm:skorokhodConst} below, which utilizes the explicit polyhedral structure of $\cK$ to achieve a tighter bound than could be obtained for general convex constraint sets. It is a variation on an earlier result from \cite{dupuis1991lipschitz}. The main distinction is that the proof in \cite{dupuis1991lipschitz} is non-constructive, and so there is no way to calculate the constants, whereas the proof in Appendix~\ref{app:skorokhod} is fully constructive and the constants can be computed explicitly.


\begin{theorem}
  \label{thm:skorokhodConst}
  There are constants $c_{\ref{diamBound}}$ and $\alpha \in (0, 1/2]$ such that 
  if $x = \cS(y)$ and $x'=\cS(y')$ are Skorokhod solutions on the polyhedral set $\cK$ defined by (\ref{eq:halfspaceRep}), then for all $t\ge 0$, the following bound holds:
 $$
\sup_{0\le s \le t} \|x_s-x_s'\| \le (c_{\ref{diamBound}}+1) \sup_{0\le s\le t} \|y_s-y'_s\|.
$$
Here
 $$
  c_{\ref{diamBound}} = 6 \left(\frac{1}{\alpha}\right)^{\rank(A)/2}
  $$
and $A = \begin{bmatrix} a_1 & \cdots a_m\end{bmatrix}^\top$ whose rows are the $a_i^\top$ vectors.
\end{theorem}

\section{Limitations}
\label{sec:limitations}

Our current work is restricted to polyhedral sets. In particular, Theorem~\ref{thm:skorokhodConst} requires the polyhedral assumption, and it is unclear if Skorokhod problems satisfy similar bounds on any more general classes of constraint sets. As a result, it is unclear if our main results on projected Langevin algorithms can be extended beyond polyhedra. We also only considered constant step sizes, but in many cases decreasing or adaptive step sizes are used in practice. 
Finally, the dependence of the external data variables is limited to the class of L-mixing processes, which does not include all the real-world dependent data streams. Furthermore, it can be  difficult to check that a data stream is L-mixing without requiring strong assumptions or knowledge about how it is generated.


\section{Conclusions and future work}
In this paper, we derived non-asymptotic bounds in 1-Wasserstein distance for a constrained Langevin algorithm applied to non-convex functions with dependent data streams satisfying L-mixing assumptions. Our convergence bounds match the best known bounds of the unconstrained case up to logarithmic factors, and improve on all existing bounds from the constrained case. 
The tighter bounds are enabled by a constructive and explicit bound on Skorokhod solutions, which builds upon an earlier non-constructive bound from
\citep{dupuis1991lipschitz}.
The analysis of L-mixing variables followed by a comparatively simple averaging method. 
Future work will examine extensions beyond polyhedral domains, higher-order Langevin algorithms, alternative approaches to handling constraints, such as mirror descent,  and more sophisticated step size rules.
More specifically, future work will examine whether the projection step, and thus Skorkhod problems, can be circumvented by utilizing different algorithms, such as those based on proximal LMC \citep{brosse2017sampling}.
Additionally, applications to real-world problems such as time-series analysis and adaptive control will be studied. 



\section{Acknowledgments}
This work was supported in part by NSF CMMI-2122856. The authors thank the reviewers for helpful suggestions for improving the paper.


\bibliography{cool-refs}

\newpage

\appendix


\section{Background and results on Skorokhod problems}
\label{app:skorokhod}

In this section, we will show that when the domain is a polyhedron, rather tight bounds on solutions to Skorokhod problems can be obtained.

\subsection{Background on Skorokhod problems}

Let $\cK$ be a convex subset of $\bbR^n$ with non-empty interior.  
Let $y:[0,\infty)\to \bbR^n$
be a trajectory which is right-continuous with left limits and has $y_0\in \cK$. For each $x\in\bbR^n$, let
$N_{\cK}(x)$ be the normal cone at $x$. 
Then the functions $x_t$ and
$\phi_t$ solve the \emph{Skorokhod problem} for $y_t$ if the
following conditions hold:
\begin{itemize}
\item $x_t = y_t+\phi_t \in \cK$ for all $t\in [0,T)$.
\item The function $\phi$ has the form $\phi(t) = -\int_0^t v_s
  d\mu(s)$, where $\|v_s\|\in \{0,1\}$ and $v_s\in N_{\cK}(x_s)$ for
  all $s\in [0,T)$, while the measure, $\mu$, satisfies
  $\mu([0,T))<\infty$ for any $T>0$. 
\end{itemize}

It can be shown that if a solution exists, it is unique. See \cite{tanaka1979stochastic}. However, existence of solutions typically relies on extra requirements beyond just convexity. For example, \cite{tanaka1979stochastic} showed the existence of solutions in the case that $y$ is continuous and $\cK$ is compact. Below, we will utilize results from \cite{anulova1991diffusional} to prove existence in the case that $\cK$ is a polyhedron. Whenever solutions are guaranteed to exist, uniqueness implies that we may view the Skorokhod solution as a mapping: $x=\cS(y)$.

\subsection{Existence of solutions over polyhedra}

The following is a consequence of Theorem 4 from \cite{anulova1991diffusional}.

\begin{lemma}
  \label{lem:skorokhodExistence}
  Let $\cK$ be a polyhedron with non-empty interior.
  If $y_t$ is a trajectory in $\bbR^n$ which is right-continuous with left-limits, then $x=\cS(y)$ exists, is unique, and is right-continuous with left-limits.
\end{lemma}

\paragraph{Proof}
  To verify the conditions of Theorem 4 from \cite{anulova1991diffusional}, we just need to show that $\cK$ satisfies condition $\beta$ of that paper, which states that there exist constants $\epsilon >0 $ and $\bar \delta >0$ such that for all $x\in\partial \cK$, there exist $x_0\in \cK$ such that  $\|x-x_0\|\le \bar \delta$ and $\{y|\|y-x_0\| <\epsilon\} \subset \cK$.  We will show how to construct $\epsilon$, $\bar \delta$, and we will see that a suitable vector, $x_0$, exists for any $x\in\cK$. 

  Note that since $\cK$ is a polyhedron, there are vectors $u_1,\ldots,u_p$ such that $x\in\cK$ if and only if it  can be expressed as
  $$
  x = \sum_{i=1}^k \lambda_i u_i + \sum_{i=k+1}^p \lambda_i u_i
  $$
  with $\lambda_i\ge 0$ for $i=1,\ldots,p$ and $\sum_{i=1}^k\lambda_i = 1$. See~\cite{rockafellar2015convex}. (If $p=k$, then $\cK$ is a compact polytope, while if $k=0$, then $\cK$ is a convex cone.)

  Let $x^\star$ be an arbitrary point in the interior of $\cK$ and let $\epsilon >0$ be such that $\{y|\|y-x^\star\| <\epsilon\} \subset\cK$. 

  Pick $\bar \delta$ such that $\|u_i-x^\star\| \le \bar\delta$ for $i=1,\ldots,k$.
  
  For any $x=\sum_{i=1}^p \lambda_i u_i \in \cK$, let $x_0 = x^\star + \sum_{i=k+1}^p \lambda_i u_i$. It follows that
  \begin{align*}
    \|x-x_0\| &= \left\| \sum_{i=1}^k \lambda_i u_i - x^\star \right\| \\
              &= \left\| \sum_{i=1}^k \lambda_i (u_i - x^\star) \right\| \\
              &\le \sum_{i=1}^k \lambda_i \|u_i-x^\star\| \\
              &\le \bar \delta.
  \end{align*}

  Also, if $y\in \{y|\|y-x_0\| < \epsilon\}$, then there is a vector, $v$, with $\|v\|< \epsilon$, such that
  \begin{align*}
    y = x_0 + v = (x^\star + v) + \sum_{i=k+1}^p \lambda_i u_i.
  \end{align*}
  Now note that $x^\star + v\in \cK$, so there must be numbers $\lambda'_i\ge 0$ such that $\sum_{i=1}^k\lambda_i'=1$ and $x^\star + v = \sum_{i=1}^p \lambda_i' u_i$.
  It follows that
  $$
  y = \sum_{i=1}^p \lambda_i' u_i + \sum_{i=k+1}^p \lambda_i u_i = \sum_{i=1}^k \lambda_i' u_i + \sum_{i=k+1}^p (\lambda_i+\lambda_i')u_i \in \cK.
  $$
\hfill$\blacksquare$

%
%

\subsection{Proof of Theorem~\ref{thm:skorokhodConst}}
\label{app:diamProof}
In this subsection, we provide a short proof of Theorem \ref{thm:skorokhodConst}. A supporting Lemma is firstly presented to complete the proof.

The technical work in this subsection relies on some notation about the vectors defining $\cK$ from (\ref{eq:halfspaceRep}). 
  Let $A = \begin{bmatrix} a_1 & \cdots a_m\end{bmatrix}^\top$ be the matrix whose rows are the $a_i^\top$ vectors. For $\cI\subset \{1,\ldots,m\}$ let $A_{\cI}$ be the matrix whose rows are $a_i^\top$ for $i\in\cI$. Let $\begin{bmatrix} W_{\cI} &  V_{\cI}\end{bmatrix}$ be an orthogonal matrix such that $\cN(A_{\cI})=\cR(W_{\cI})$. Here $\cN(A_{\cI})$ denotes the null space of $A_{\cI}$ and $\cR(W_{\cI})$ denotes the range space of $W_{\cI}$. Let $P_{\cI}=W_{\cI}W_{\cI}^\top$, which is the orthogonal projection onto $\cN(A_{\cI})$. We  will use the convention that $A_{\emptyset}$ is a $1\times n$ matrix of zeros, so that $\cN(A_{\emptyset})=\bbR^n$, and thus $P_{\emptyset}=I$.

The following lemma is a quantitative and explicit version of Theorem 2.1 of \cite{dupuis1991lipschitz}:
\const{diamBound}
\newcommand{\diamBoundVal}{6 \left(\frac{1}{\alpha}\right)^{\rank(A)/2}}
\begin{lemma}
  \label{lem:boundingExistence}
  If  $\cK$ is a polyhedron defined by (\ref{eq:halfspaceRep}), then there is a compact, convex set $\cB$ with $0\in\mathrm{int}(\cB)$ such that if $z\in\partial \cB$, $v\in N_{\cB}(z)$, and $a_j$ is a unit vector from (\ref{eq:halfspaceRep}) with $a_j^\top v\ne 0$, then
  \begin{enumerate}
  \item \label{item:largeProduct}
    $|a_j^\top z| \ge 1$
  \item \label{item:sameSign}
    $\mathrm{sign}(a_j^\top z) = \mathrm{sign}(a_j^\top v)$.
  \end{enumerate}
  Furthermore, the diameter of $\cB$ is at most $c_{\ref{diamBound}}$, defined by
  $$
  c_{\ref{diamBound}} = 6 \left(\frac{1}{\alpha}\right)^{\rank(A)/2}
  $$
  where 
\begin{align*}
\alpha=\frac{1}{2}\min\left\{\|P_{\cI} a_j\|^2 \middle| P_{\cI}a_j\ne 0, 
\: \cI\subset \{1,\ldots,m\},\: j\in\{1,\ldots,m\}  \right\}, 
\end{align*}  
 and $\alpha \in (0,1/2] $.
\end{lemma}

A non-constructive proof of the existence of $\cB$ was given in \cite{dupuis1991lipschitz}. While that paper shows that $\cB$ is compact, it does not quantitatively bound its diameter. The diameter of $\cB$ is precisely the quantity that is used to bound the difference between Skorokhod solutions. 

 \paragraph*{Proof of Theorem \ref{thm:skorokhodConst}.}
 Theorem 2.2 of \cite{dupuis1991lipschitz} shows that if a compact convex set with $0\in\mathrm{int}(B)$ satisfying conditions \ref{item:largeProduct} and \ref{item:sameSign} exists, then
 $$
\sup_{0\le s \le t} \|x_s-x_s'\| \le (\mathrm{diameter}(\cB)+1) \sup_{0\le s\le t} \|y_s-y'_s\|.
 $$
 The result now follows since $c_{\ref{diamBound}}$ is an upper bound on the diameter of the set $\cB$ constructed in Lemma~\ref{lem:boundingExistence}.
\hfill$\blacksquare$

  \paragraph*{Proof of Lemma \ref{lem:boundingExistence}.}
  We will focus on constructing a compact, convex $\cB$ with $0\in\mathrm{int}(\cB)$ which satisfies condition \ref{item:largeProduct}. Lemma 2.1 of \cite{dupuis1991lipschitz} shows that condition \ref{item:sameSign} must also hold. (Note that  the sign is opposite of what appears in \cite{dupuis1991lipschitz}, because that paper examines inward normal vectors, while we are examining outward normal vectors.)

  We will find numbers $\epsilon \in (0,1)$ and $r_{\cI}\in (0,1)$ for $\cI\subset \{1,\ldots,m\}$ such that
  $$
  \cB = \{x |  \|P_{\cI} x\| \le \epsilon^{-1} r_{\cI} \textrm{ for } \cI \in \{1,\ldots,m\} \}
  $$
  has the desired properties. By construction, $\cB$ is compact and convex, $0\in\mathrm{int}(\cB)$, and the diameter is at most $2\epsilon^{-1} r_{\emptyset} < 2\epsilon^{-1}$, since every $x\in\cB$ satisfies $\|P_{\emptyset}x\|=\|x\|\le \epsilon^{-1}r_{\emptyset}$. Furthermore, $\cB = \epsilon^{-1} \hat\cB$, where
  $$
  \hat \cB = \{x | \|P_{\cI} x\| \le r_{\cI} \textrm{ for } \cI \subset \{1,\ldots,m\}\}.
  $$
A similar construction for $\cB$ was utilized in \cite{dupuis1991lipschitz}. The main distinction is that this proof will give an explicit procedure for determining the values of $\epsilon$ and $r_{\cI}$.

  Note that $z\in \hat \cB$ if and  only if $\epsilon^{-1}z \in \cB$, $z\in\partial \hat\cB$ if and only if $\epsilon^{-1}z\in \partial \cB$, and 
  $N_{\hat\cB}(z)=N_{\cB}(\epsilon^{-1} z)$. Thus, Condition \ref{item:largeProduct} holds for $\cB$ if and only if

  \begin{equation}
    \label{eq:revisedImplication}
    z\in\partial\hat\cB,\: v\in N_{\hat\cB}(z),\: \textrm{ and } a_j^\top v\ne 0 \implies |a_j^\top z|\ge \epsilon>0.
  \end{equation}

  Note that if $x\in\partial\hat \cB$, then
  \begin{align}
    \nonumber
  N_{\hat\cB}(x) &= \mathrm{cone}\{P_{\cI}x | \|P_{\cI}x\|=r_{\cI}\} \\
  \label{eq:normalConeRep}
  &= \left\{\sum_{\{\cI | \|P_{\cI}x\|=r_{\cI} \}} \lambda_{\cI}P_{\cI}x \middle| \lambda_{\cI}\ge 0 \right\}. 
  \end{align}
  See Corollary 23.8.1 of \cite{rockafellar2015convex}.

  The representation in (\ref{eq:normalConeRep}) implies that 
   if $x\in\partial\hat\cB$, $v\in N_{\cB}(x)$, and $a_j^\top v\ne 0$, then there must be a set $\cI$ such that, $\|P_{\cI}x\|=r_{\cI}$, $\lambda_{\cI}>0$, and $a_j^\top P_{\cI}\ne 0$.
  We will choose $\epsilon$ such that for all $\cI$ and $j$ with $P_{\cI}a_j\ne 0$, $\epsilon$ is a lower bound on the optimal value of the following (non-convex) optimization problem:
  \begin{subequations}
    \label{eq:epsDefProblem}
  \begin{align}
    &\min_{x} && |a_j^\top x| \\
    &\textrm{subject to} && \|P_{\cI}x\| \ge r_{\cI} \\
    &&& \|P_{\cI\cup\{j\}} x\| \le r_{\cI\cup \{j\}} \\
    &&& \|x\|\le 1.
  \end{align}
\end{subequations}

By construction, if $x\in\partial\hat \cB$, $v\in N_{\cB}(x)$, and $a_j^\top v\ne 0$, there must be some $\cI$ such that $x$ is feasible for (\ref{eq:epsDefProblem}). As a result, we must have that $|a_j^\top x|\ge \epsilon$. Thus, the implication from (\ref{eq:revisedImplication}) will hold, provided that the values of $r_{\cI}$ can be chosen so that all of the problems of the form (\ref{eq:epsDefProblem}) have strictly positive optimal values.

The rest of the proof proceeds as follows. First we derive conditions on $r_{\cI}$ that ensure that the problems from  (\ref{eq:epsDefProblem}) always have positive optimal values. Next, we compute specific values of  $r_{\cI}$  that satisfy these conditions. Finally, we use those values of $r_{\cI}$ to compute $\epsilon$, the desired lower bound on the optimal value of (\ref{eq:epsDefProblem}).

We now assume that $r_{\cI},r_{\cI\cup\{j\}}\in (0,1)$ and derive sufficient conditions to make the optimal value in (\ref{eq:epsDefProblem}) strictly positive.

To derive the optimal value of (\ref{eq:epsDefProblem}), we need a few basic facts:
\begin{itemize}
\item If $\cI\subset \cJ$, then $P_{\cJ}P_{\cI}=P_{\cJ}$ and $P_{\cI}P_{\cJ}=P_{\cJ}$.
\item The matrix $\begin{bmatrix}W_{\cI\cup\{j\}} & \frac{P_{\cI}a_j}{\|P_{\cI}a_j\|} & V_{\cI} \end{bmatrix}$ is orthogonal.
\end{itemize}

First we show that $\cI\subset \cJ$ implies that $P_{\cJ}P_{\cI}=P_{\cJ}$. Symmetry of the projection matrices would then imply that $P_{\cI}P_{\cJ}=P_{\cJ}$. Note that $P_{\cI}=I-V_{\cI}V_{\cI}^\top$, where
$$
\cR(V_{\cI}) = \cR(P_{\cI})^{\perp} = \cN(A_{\cI})^\perp \subset \cN(A_{\cJ})^\perp = \cR(P_{\cJ})^\perp.
$$
It follows that $P_{\cJ}V_{\cI}=0$ and thus $P_{\cJ}P_{\cI}=P_{\cJ}$.

Now we will show that $P_{\cI} a_j \in \cR(P_{\cI})\setminus \cR(P_{\cI\cup\{j\}})$. By construction, $P_{\cI}a_j \in \cR(P_{\cI})$. Also, we have that $P_{\cI\cup \{j\}}P_{\cI}a_j = P_{\cI\cup \{j\}} a_j = 0$, where the second equality follows because $a_j \in \cN(A_{\cI\cup\{j\}})^\perp = \cR(P_{\cI\cup\{j\}})^\perp$.  Thus, we have that $P_{\cI}a_j\ne P_{\cI\cup\{j\}} P_{\cI} a_j$.
Now, if $P_{\cI}a_j\in \cR(P_{\cI\cup\{j\}})$, then $P_{\cI}a_j = P_{\cI\cup \{j\}} z$ for some vector $z$. But then $P_{\cI\cup \{j\}}^2 = P_{\cI\cup\{j\}}$ would imply that $P_{\cI\cup \{j\}}P_{\cI}a_j = P_{\cI\cup\{j\}}z = P_{\cI}a_j$, which gives a contradiction.
Thus, $P_{\cI}a_j\notin \cR(P_{\cI\cup\{j\}})$.

Now the rank nullity theorem implies that
\begin{align*}
\rank(A_{\cI})&=n-\dim(\cN(A_{\cI}))\\
  \rank(A_{\cI\cup\{j\}})&=n-\dim(\cN(A_{\cI\cup\{j\}})).
\end{align*}

Now since $A_{\cI\cup\{j\}}$ has only one more row than $A_{\cI}$, we must have that $\rank(A_{\cI})\le \rank(A_{\cI\cup\{j\}})\le \rank(A_{\cI})+1$. Also, $\cN(A_{\cI\cup\{j\}})\subset \cN(A_{\cI}) $ by construction, and we just saw that $P_{\cI}a_j \in \cN(A_{\cI})\setminus \cN(A_{\cI\cup \{j\}})$, so the inclusion is strict. 
It follows that
\begin{align}
  \label{eq:nullspaceDimensions}
  \nonumber
\dim(\cN(A_{\cI}))&=\dim(\cR(P_{\cI}))\\
&=\dim(\cN(A_{\cI\cup\{j\}}))+1   
\nonumber
\\ 
&=\dim(\cR(P_{\cI\cup\{j\}}))+1.
\end{align}

Now, since $\cR(W_{\cI\cup\{j\}})=\cN(A_{\cI\cup\{j\}})$, we must have that
  $$
  \cR\left(\begin{bmatrix}W_{\cI\cup\{j\}} & \frac{P_{\cI}a_j}{\|P_{\cI}a_j\|} \end{bmatrix}\right) = \cN(A_{\cI}).
  $$
Furthermore, since $\cR(W_{\cI\cup\{j\}})=\cR(P_{\cI\cup\{j\}})$ and $P_{\cI\cup \{j\}} P_{\cI}a_j=0$, we must  have that 
$$
\begin{bmatrix}W_{\cI\cup\{j\}}^\top \\ \frac{(P_{\cI}a_j)^\top}{\|P_{\cI}a_j\|} \\ V_{\cI}^\top \end{bmatrix}
\begin{bmatrix}W_{\cI\cup\{j\}} & \frac{P_{\cI}a_j}{\|P_{\cI}a_j\|} & V_{\cI} \end{bmatrix}
=I
$$

Now we use this orthogonal matrix to perform a change of coordinates. In particular, let $y_1$, $y_2$, and $y_3$ be such that
$$
x = W_{\cI\cup \{j\}} y_1 + \frac{P_{\cI}a_j}{\|P_{\cI}a_j\|} y_2 + V_{\cI}y_3.
$$
In these new coordinates, (\ref{eq:epsDefProblem}) is equivalent to
\begin{subequations}
\label{eq:epsDefChanged}
\begin{align}
  & \min_{y} && |\|P_{\cI}a_j\|y_2 + a_j^\top V_{\cI} y_3| \\
  \label{eq:partialProj1}
  &\textrm{subject to}&& \|y_1\|^2 + y_2^2 \ge r_{\cI}^2 \\
  &&& \|y_1\| \le r_{\cI\cup\{j\}} \\
  \label{eq:totalNormBound}
  &&& \|y_1\|^2 + y_2^2 + \|y_3\|^2 \le 1.
\end{align}
\end{subequations}

The equivalence arises because
\begin{align*}
  a_j^\top x &= \|P_{\cI}a_j\| y_2 + a_j^\top V_{\cI}y_3  \\
  P_{\cI}x &= W_{\cI\cup \{j\}} y_1 + \frac{P_{\cI}a_j}{\|P_{\cI}a_j\|}y_2 \\
  P_{\cI\cup\{j\}} x &= W_{\cI\cup \{j\}} y_1 
\end{align*}
along with orthogonality of the corresponding transformation from $y$ to $x$.

If we choose $r_{\cI} > r_{\cI\cup \{j\}}$, then we must have
$$
y_2^2 \ge r_{\cI}^2 - \|y_1\|^2 \ge r_{\cI}^2-r_{\cI\cup \{j\}}^2 > 0.
$$
Now, if $y$ is feasible, $-y$ is also feasible, and they have the same objective value in (\ref{eq:epsDefChanged}). So, without loss of generality, we may assume that $y_2 > 0$.

The Cauchy-Schwartz inequality, combined with (\ref{eq:totalNormBound}),  implies that
\begin{equation}
  \|P_{\cI}a_j\|y_2 +  a_j^\top V_{\cI} y_3   
    \label{eq:csLower}
  \ge  \|P_{\cI}a_j\|y_2 -  \|V_{\cI}^\top a_j\| \sqrt{1-\|y_1\|^2-y_2^2}. 
\end{equation}
Note that this bound is achieved by setting $y_3 = -\frac{V_{\cI}^\top a_j}{\|V_{\cI}^\top a_j\|}\sqrt{1-\|y_1\|^2-y_2^2}$. 

The right side of (\ref{eq:csLower}) is monotonically increasing in $y_2$. So, (\ref{eq:partialProj1}) implies that it is minimized over $y_2$ by setting $y_2=\sqrt{r_{\cI}^2-\|y_1\|^2}$. This leads to a lower bound of the form:
\begin{equation*}
  \|P_{\cI}a_j\|y_2 -  \|V_{\cI}^\top a_j\| \sqrt{1-\|y_1\|^2-y_2^2}
  \ge
  \|P_{\cI}a_j\|\sqrt{r_{\cI}^2-\|y_1\|^2} - \|V_{\cI}^\top a_j\| \sqrt{1-r_{\cI}^2}.
\end{equation*}
The right  side is now monotonically decreasing with respect to $\|y_1\|$, and so it is minimized by setting $\|y_1\|=r_{\cI\cup\{j\}}$. This leads to the characterization: 
\begin{align}
  &\textrm{Optimal Value of (\ref{eq:epsDefChanged})}  \nonumber \\
  &= 
  \|P_{\cI}a_j\|\sqrt{r_{\cI}^2-r_{\cI\cup\{j\}}^2} - \|V_{\cI}^\top a_j\| \sqrt{1-r_{\cI}^2} \nonumber \\
  \label{eq:optVal}  
  &= \|P_{\cI}a_j\|\sqrt{r_{\cI}^2-r_{\cI\cup\{j\}}^2} - \sqrt{1-\|P_{\cI}^\top a_j\|^2} \sqrt{1-r_{\cI}^2}. 
\end{align}
The second equality follows because
$$
\|V_{\cI}^\top a_j\|^2 = a_j^\top V_{\cI}V_{\cI}^\top a_j = a_j^\top (I-P_{\cI})a_j = 1-\|P_{\cI}a_j\|^2.
$$

Now, we have that 
the right side of (\ref{eq:optVal}) is positive if and only if:
\begin{subequations}
\label{eq:inductiveRbound}
\begin{align}
  \MoveEqLeft
\|P_{\cI}a_j\|^2\left(r_{\cI}^2-r_{\cI\cup\{j\}}^2 \right) > \left(1-\|P_{\cI} a_j\|^2\right)\left(1-r_{\cI}^2\right)
  \\
  \label{eq:inductiveRboundIterative}
  &\iff r_{\cI}^2 > 1-\|P_{\cI}a_j\|^2 + \|P_{\cI}a_j\|^2 r_{\cI\cup\{j\}}^2\\
  \label{eq:inductiveRboundMonotone}
  &\iff r_{\cI}^2 > 1-\|P_{\cI}a_j\|^2 (1-r_{\cI\cup\{j\}}^2) \\
  \label{eq:inductiveRboundFinal}
  & \iff r_{\cI}^2 > r_{\cI\cup\{j\}}^2 +(1-\|P_{\cI}a_j\|^2)(1-r_{\cI\cup\{j\}}^2).
\end{align}
\end{subequations}


Note that (\ref{eq:inductiveRboundFinal}) implies that $r_{\cI} > r_{\cI\cup\{j\}}$ holds.

Also note that any collection of $r_{\cI}$ values in $(0,1)$ that satisfy (\ref{eq:inductiveRbound}) will ensure that  the corresponding set, $\hat\cB$, satisfies the implication from (\ref{eq:revisedImplication}). In that case, we have that $\cB$ has the desired properties.

Now we seek a simpler, more explicit formula for the $r_{\cI}$ values which satisfy (\ref{eq:inductiveRbound}). 
Note that (\ref{eq:inductiveRboundMonotone}) implies that the right side is monotonically decreasing with respect to $\|P_{\cI}a_j\|^2$. So, if $\alpha>0$ is a number such that $\alpha \le \frac{1}{2}\|P_{\cI}a_j\|^2$ for all $\cI$ and $j$ with $P_{\cI}a_j\ne 0$ we obtain a sufficient condition for (\ref{eq:inductiveRbound}): 
\begin{subequations}
  \label{eq:relaxedRbound}
\begin{align}
  r_{\cI}^2& = 1-\alpha(1-r_{\cI\cup\{j\}}^2) \\
  r_{\cI}^2& =   (1-\alpha) + \alpha r_{\cI\cup\{j\}}^2.
\end{align}
\end{subequations}

Now we use (\ref{eq:relaxedRbound}) to derive the desired formula for $r_{\cI}$. In particular, consider the recursion
$$
x_{k+1}=(1-\alpha) + \alpha x_k.
$$
This has an explicit solution given by
$$
x_{k} = \alpha^k x_0 + 1-\alpha^k = 1-\alpha^k(1-x_0).
$$
In particular, if $x_0\in (0,1)$, we have that $x_k\in (0,1)$ for all $k\ge 0$.

We define $r_{\cI}$ by fixing a value $x_0\in (0,1)$, which will be defined explicitly later, and setting $r_{\cI}^2=x_k=1-\alpha^k(1-x_0)$ if $\rank(A)-\rank(A_{\cI})=k$. 

To see that this definition satisfies (\ref{eq:relaxedRbound}), first note that $r_{\cI}^2= x_0$ for all $\cI$ with $\rank(A)=\rank(A_{\cI})$. Now, recall that if $P_{\cI}a_j\ne 0$, then (\ref{eq:nullspaceDimensions}) implies that $\rank(A_{\cI\cup\{j\}})=\rank(A_{\cI})+1$. The converse is also true: If $\rank(A_{\cI\cup\{j\}})=\rank(A_{\cI})+1$, then we must have that  $a_j\notin \cR(A_{\cI}^\top)=\cR(V_{\cI})=\cR(P_{\cI})^\perp$. It follows that $P_{\cI}a_j\ne 0$. Thus, if $\rank(A)-\rank(A_{\cI\cup\{j\}})=k\ge 0$, we have that $P_{\cI}a_j\ne 0$ precisely when $\rank(A)-\rank(A_{\cI})=k+1$. So we see that setting $r_{\cI}^2=x_{k+1}=1-\alpha(1-x_k)$ gives the same value as specified in (\ref{eq:relaxedRbound}). 

The final step in the proof requires finding a lower bound, $\epsilon$, for the optimal value from (\ref{eq:optVal}). Let $r_{\cI}^2=x_k$ and $r_{\cI\cup\{j\}}^2=x_{k-1}$. Then we have that
\begin{align*}
  r_{\cI}^2-r_{\cI\cup\{j\}}^2 &= (1-\alpha)\alpha^{k-1}(1-x_0)\\
  1-r_{\cI}^2 &= \alpha \alpha^{k-1} (1-x_0).
\end{align*}
Also note that the right side of (\ref{eq:optVal}) is monotonically increasing with respect to $\|P_{\cI}a_j\|^2$ and that $\|P_{\cI}a_j\|^2 \ge 2\alpha$ by our choice of $\alpha$. So, plugging in this lower bound gives  
\begin{align*}
  \MoveEqLeft
  \|P_{\cI}a_j\|\sqrt{r_{\cI}^2-r_{\cI\cup\{j\}}^2} - \sqrt{1-\|P_{\cI}^\top a_j\|^2} \sqrt{1-r_{\cI}^2}
  \\
  &\ge
 \left(\sqrt{2\alpha}\sqrt{1-\alpha}-\sqrt{1-2\alpha}\sqrt{\alpha}\right)\sqrt{\alpha^{k-1}(1-x_0)}
  \\
  &=\left(\sqrt{2-2\alpha}-\sqrt{1-2\alpha}\right)\sqrt{\alpha^k(1-x_0)}
  \\
  &\ge \left(\sqrt{2}-1\right) \sqrt{\alpha^{\rank(A)} (1-x_0)}
\end{align*}
The final inequality follows because $k\le \rank(A)$ and the minimum value of $\sqrt{2-2\alpha}-\sqrt{1-2\alpha}$ over $\alpha \in [0,\|P_{\cI}a_j\|^2/2] \subset [0,1/2]$ occurs at $\alpha=0$.

To simplify the final formula for $\epsilon$, note that $\sqrt{2}-1>1/3$, and thus we can choose $x_0\in(0,1)$ so that
$$
(\sqrt{2}-1)\sqrt{1-x_0}=\frac{1}{3} \iff x_0 = 1-\frac{1}{9\left(\sqrt{2}-1\right)^2}\approx 0.352.
$$
Plugging in this value for $x_0$ gives the bound:
\begin{equation*}
  \textrm{Optimal Value of (\ref{eq:epsDefChanged})} \ge \frac{1}{3}\alpha^{\frac{\rank(A)}{2}}=:\epsilon
\end{equation*}
Now recalling that the diameter of $\cB$ is at most $2/\epsilon$ completes the proof.
\hfill$\blacksquare$

\section{Invariance of the Gibbs measure}
\label{sec:gibbs}

\begin{lemma}
  \label{lem:gibbs}
 The Gibbs measure, (\ref{eq:gibbs}), is stationary under the dynamics of the reflected SDE from (\ref{eq:AvecontinuousProjectedLangevin}). 
\end{lemma}

\paragraph{Proof}
Before showing invariance of the Gibbs measure, we first remark that it is well-defined. In particular, we have that $\int_{\cK}e^{-\beta \bar f(x)} dx<\infty$.

To see this, let $\|x\| \ge R/\theta$, where $\theta\in (0,1)$ is a number to be chosen later. Note that for $t\in [\theta,1]$, we have that $\|\theta x\|\ge R$. So, we can use strong convexity outside a ball of radius $R$ to show
\begin{align}
  \nonumber
  \bar f(x)&\ge \bar f(0)+\int_0^1 \nabla \bar f(tx)^\top x dt \\
  \nonumber
           &=\bar f(0) + \nabla \bar f(0)^\top x + \int_0^\theta \left(\nabla \bar f(tx)-\nabla\bar f(0) \right)^\top x dt +
             \int_{\theta}^1 \left(\nabla \bar f(tx)-\nabla\bar f(0) \right)^\top x dt  \\
  \nonumber
           &\ge \bar f(0)-\|\nabla \bar f(0)\| \|x\| -\ell \|x\|^2 \int_0^\theta t dt
             +\mu \|x\|^2 \int_{\theta}^1 t dt \\
  \nonumber
           & \ge \bar f(0)-\|\nabla \bar f(0)\| \|x\| + \frac{1}{2} \|x\|^2\left(
             - \ell \theta^2 + \mu(1-\theta^2) 
             \right)
\end{align}

The coefficient $-\ell^2 \theta^2 + \mu (1-\theta^2)$ is positive, as long as $\theta < \sqrt{\frac{\mu}{\mu+\ell}}$. In particular, choosing $\theta^2 = \frac{1}{2}\frac{\mu}{\mu+\ell}$ gives
\begin{equation}
  \label{eq:quadraticLower}
  \bar f(x) \ge \bar f(0)-\|\nabla \bar f(0)\| \|x\| +\frac{1}{4}\mu \|x\|^2.
\end{equation}

  It follows that $Z=\int_{\cK}e^{-\beta \bar f(x)}dx < \infty$.

  In \cite{lamperski2021projected}, it was shown in that the Gibbs measure is invariant under (\ref{eq:AvecontinuousProjectedLangevin}) when $\cK$ is compact. We will extend the result to non-compact $\cK$ via a limiting argument.

  Let $\cK_i = \cK \cap \{x\in\bbR^n | \|x\|_{\infty}\le i\}$. Let $Z_i=\int_{\cK_i}e^{-\beta \bar f(x)}dx$. Note that $\lim_{i\to\infty}Z_i=Z$, by monotone convergence. We choose $\|x\|_\infty = \max\{|x_1|,\ldots,|x_n|\} \le i$ so that $\cK_i$ becomes a compact polyhedron for $i\ge 1$.
  
  Let $\bx_t^C$ be a solution to the original form of (\ref{eq:AvecontinuousProjectedLangevin}) and let $\bx_t^{C,i}$ be a solution to the RSDE from (\ref{eq:AvecontinuousProjectedLangevin}), with $\cK_i$ used in place of $\cK$. Since $\cK_i$ is polyhedral, Lemma~\ref{lem:skorokhodExistence} in Appendix~\ref{app:skorokhod} shows that $\bx_t^{C,i}$ is uniquely defined. 
  Define the diffusion operators $P$ and $P^i$ by 
  \begin{align*}
    (P_tg)(x) &= \bbE[g(\bx_t^C)|\bx_0=x] \\
    (P_t^i g)(x) &= \bbE[g(\bx_t^{C,i})|\bx_0=x]
  \end{align*}
  
  Let $L_2(\cK,\pi_{\beta \bar f})$ be the set of functions $g:\cK\to \bbR$ which are square integrable with respect to the measure $\pi_{\beta \bar f}$. We will show that $\pi_{\beta \bar f}$ is invariant for (\ref{eq:AvecontinuousProjectedLangevin}) by showing that for all $g\in L_2(\cK,\pi_{\beta \bar f})$ the following equality holds for all $t\ge 0$:
  \begin{equation}
  \label{eq:stationaryTransition}
  \frac{1}{Z}\int_{\cK} g(x)e^{-\beta \bar f(x)}dx = \frac{1}{Z}\int_{\cK} (P_tg)(x)e^{-\beta \bar f(x)}dx. 
  \end{equation} 

  The subset of bounded, compactly supported functions in $L_2(\cK,\pi_{\beta \bar f})$ is a dense subset.
  Fix an arbitary bounded, compactly supported $g\in L_2(\cK,\pi_{\beta \bar f})$. It suffices to show that \eqref{eq:stationaryTransition} holds for $g$.

  Lemma 19 of \cite{lamperski2021projected} shows that for all $i\ge 1$, the following holds:
    \begin{equation}
  \label{eq:stationaryTransitionCompact}
  \frac{1}{Z_i}\int_{\cK_i} g(x)e^{-\beta \bar f(x)}dx = \frac{1}{Z_i}\int_{\cK_i} (P_t^ig)(x)e^{-\beta \bar f(x)}dx. 
  \end{equation} 

  We saw earlier that $Z_i\to Z$. Furthermore, since $g$ is compactly supported, there is a number, $m$, such that $i\ge m$ implies that
$$
\int_{\cK_i} g(x)e^{-\beta \bar f(x)}dx = \int_{\cK}g(x)e^{-\beta \bar f(x)}dx.
$$
It follows that the left of \eqref{eq:stationaryTransitionCompact} converges to the left of \eqref{eq:stationaryTransition}.

The proof will be completed if we can show that for $t>0$,
\begin{align}
  \label{eq:doubleIntegralSeq}
  \lim_{i\to\infty} \int_{\cK_i} (P_t^ig)(x)e^{-\beta \bar f(x)}dx 
  =&\lim_{i\to\infty}\int_{\cK_i}\bbE[g(\bx_t^{C,i})|\bx_0=x] e^{-\beta \bar f(x)}dx \\
  \label{eq:doubleIntegralFinal}
  =&
\int_{\cK}\bbE[g(\bx_t^{C})|\bx_0=x] e^{-\beta \bar f(x)}dx \\
  \nonumber
  = &\int_{\cK}(P_tg)(x)e^{-\beta \bar f(x)}dx. 
\end{align}

We assumed that $g$ was bounded, and so there is a number, $b$, such that $|g(x)|\le b$ for all $x\in\cK$. It follows from the definition of $P_t$ and $P_t^i$ that $|P_tg(x)|\le b$ and $|P_t^i g(x)|\le b$ for all $t$.

Fix any $t>0$. The Brownian motion, $\bw$, is continuous, and so for each $t$, $\bw_s$ is bounded for $s\in [0,t]$. Now the form of (\ref{eq:AveContinuousY}) shows that $\by^C$, and thus $\bx^C$ must also be continuous, and thus also bounded for $s\in [0,t]$. Thus, for each realization, we see that there is a number $m$ such that $\bx_s^C\in \cK_i$ for all $s\in [0,t]$ and all $i\ge m$. Thus, we see that $\bx_s^{C}=\bx_s^{C,i}$ for $s\in [0,t]$. This argument shows that the integrand on the right of \eqref{eq:doubleIntegralSeq} converges pointwise to the integrand of \eqref{eq:doubleIntegralFinal}. So, the desired equality follows by the dominated convergence theorem. 
\hfill$\blacksquare$

\section{Bounded variance of the processes}
\label{app:bounded}

In this section, we derive variance bounds on all of the main processes, $\bbE[\|\bx_k^A\|^2]$ and $\bbE[\|\bx_t^C\|^2]$. The bound on $\bbE[\|\bx_t^C\|^2]$ is used to prove bounds on the discretization error from $\bx_t^M$ to $\bx_t^C$. The bound on $\bbE[\|\bx_k^A\|^2]$ is used to derive the  time-uniform bounds on $W_1(\cL(\bx_k^A),\cL(\bx_k^C))$ from Lemma~\ref{lem:AtoC}.

\subsection{Continuous-time bounds}
\label{app:contBounds}
In this section, we show that the assumption that $\bar f$  is strongly convex outside a ball implies that $\cV(x)=\frac{1}{2}\|x\|^2$ can be used as a Lyapunov function for $\bx_t^C$. In turn, we use this Lyapunov function to derive bounds on $\bbE[\|\bx_t^C\|^2]$. 

\const{LyapunovConst}
\newcommand{\LyapunovConstVal}{(\ell+\mu)R^2 + R\|\nabla_x \bar f(0)\| + \frac{n}{\beta}}

\begin{lemma}
\label{lem:GeometricDrift}
If $\bar f(x)$ is $\mu$-strongly convex outside a ball with radius $R$, then  $\cV(x) = \frac{1}{2} x^\top x$ satisfies the following the geometric drift condition:
\begin{align*}
	\cA \cV(x) \le  -2\eta \mu \cV(x) + c_{\ref{LyapunovConst}} \eta.
\end{align*}
Here $c_{\ref{LyapunovConst}}$ is defined by
\begin{align*}
	c_{\ref{LyapunovConst}} &= \LyapunovConstVal.
\end{align*}
\end{lemma}
\paragraph{Proof}
By Ito's formula, we have
\begin{align*}
	d\cV(\bx^C_t) &= \nabla_x \cV ^\top d\bx^C_t 	+	
	\frac{1}{2} d(\bx^C_t)^\top (\nabla_x^2\cV)d\bx^C_t\\
	&= (\bx^C_t)^\top (-\eta \nabla_x \bar{f}(\bx^C_t)dt 
	+ \sqrt{\frac{2\eta}{\beta}}d\bw_t 
	- \bv_t d\bmu_t) + \frac{1}{2}d(\bx^C_t)^\top d\bx^C_t\\
	&= - \eta (\bx^C_t)^\top \nabla_x \bar{f}(\bx^C_t)dt
	+ \sqrt{\frac{2\eta}{\beta}} (\bx^C_t)^\top d\bw_t  - (\bx^C_t)^\top \bv_t d\bmu_t + \frac{\eta}{\beta} \Tr(d\bw_t d\bw_t^\top) \\
	&= (- \eta (\bx^C_t)^\top \nabla_x \bar{f}(\bx^C_t)+ \frac{n \eta}{\beta})dt 
	+ \sqrt{\frac{2\eta}{\beta}}  (\bx^C_t)^\top d\bw_t - (\bx^C_t)^\top \bv_t d\bmu_t. 
\end{align*}
The third equality holds because $\int_0^t \bv_s d \bmu_s$ has bounded variation. The last equality is based on the fact that $d\bw_t d\bw_t^\top =  dt \:I$.

Since $\bv_t \in N_\cK(\bx_t^C)$, $\bmu_t$ is a nonnegative measure, and $0\in\cK$, we have that $- (\bx_t^C)^\top \bv_t d\bmu_t  \le 0 $. Thus, the generator of the Lyapunov function satisfies 
\begin{align}
	\label{eq:generalLyapunovGen}
	\cA \cV(x) &\le - \eta x^\top \nabla_x \bar{f}(x)+ \frac{n \eta}{ \beta}.
\end{align}

If $\|x\|\ge R$, strong convexity outside a ball of radius $R$, along with the Cauchy-Schwartz inequality imply that
\begin{align}
  \nonumber
  x^\top \nabla_x \bar f(x) & = (x-0)^\top (\nabla_x \bar f(x)-\nabla_x \bar f(0)) + x^\top \nabla_x \bar f(0) \\
  \label{eq:bigXInnerProduct}
  &\ge \mu \|x\|^2 - R\|\nabla_x \bar f(0)\|
\end{align}

It follows that when $\|x\|\ge R$, we have that
\begin{align*}
  \cA\cV(x) &\le -\eta\mu \|x\|^2 + \eta \left( R\|\nabla_x \bar f(0)\| + \frac{n}{\beta} \right)  \\
  &= -\eta 2\mu \cV(x) + \eta \left( R\|\nabla_x \bar f(0)\| + \frac{n}{\beta} \right).
\end{align*}

If $\|x\|\le R$, then  the Cauchy-Schwartz inequality and the Lipschitz continuity imply that
\begin{align}
  \nonumber
  - x^\top \nabla_x \bar{f}(x) &= -  x^\top \left(\nabla_x \bar{f}(x)-\nabla_x \bar{f}(0) +\nabla_x \bar{f}(0) \right)\\
  \nonumber
  &\le  \|x \| \|\nabla_x \bar{f}(x)-\nabla_x \bar{f}(0)\| + R \|\nabla_x \bar{f}(0)\|\\
  \nonumber
                               &\le \ell \|x \|^2 + R \|\nabla_x \bar{f}(0)\| \\
  \nonumber
                               &= -\mu \|x\|^2 + (\ell+\mu) \|x\|^2 + R \|\nabla_x \bar{f}(0)\| \\
  \label{eq:smallXInnerProduct}
  &\le -\mu \|x\|^2 + (\ell+\mu)R^2 + R\|\nabla_x \bar f(0)\|.
\end{align}

Note that \eqref{eq:bigXInnerProduct} implies that \eqref{eq:smallXInnerProduct} also holds whenever $\|x\|\ge R$. So, combining  (\ref{eq:smallXInnerProduct}) with (\ref{eq:generalLyapunovGen}) shows that for all $x\in\cK$, 
\begin{align*}
  \cA\cV(x)&\le \eta \left(  -\mu \|x\|^2 + (\ell+\mu)R^2 + R\|\nabla_x \bar f(0)\| \right) + \frac{n\eta}{\beta} \\
  &=-\eta 2\mu \cV(x) + \eta \left(
     (\ell+\mu)R^2 + R\|\nabla_x \bar f(0)\| + \frac{n}{\beta}
    \right)
\end{align*}
\hfill$\blacksquare$

\begin{lemma}
  \label{lem:continuousBound}
  If $\bbE[\|\bx^C_0\|^2]\le \varsigma$, then for all $t\ge 0$, we have that
  $$
  \bbE[\|\bx^C_t\|^2]\le \varsigma +\frac{1}{\mu} c_{\ref{LyapunovConst}},
  $$
  where $c_{\ref{LyapunovConst}}$ is defined in Lemma~\ref{lem:GeometricDrift}.
\end{lemma}
\paragraph{Proof}
Recall that Lyapunov generator $\cA$ is defined as below
\begin{align*}
	\cA\cV(x) = \lim_{t\downarrow0} \bbE\left[ \frac{1}{t}(\cV(\bx^C_t) - \cV(\bx^C_0))\vert \bx^C_0 = x\right].
\end{align*}
Using Dynkin's formula and Lemma $\ref{lem:GeometricDrift}$ gives
\begin{align*}
	\bbE\left[ \cV(\bx^C_t) - \cV(\bx^C_0) \right] 
	&= \int_0^t \bbE\left[ \cA \cV(\bx^C_s)\right] ds\\
	&\le - 2\eta \mu \int_0^t \bbE\left[ \cV(\bx^C_s)\right]ds + c_{\ref{LyapunovConst}}\eta t.
\end{align*}


Let $u_t = \bbE\left[ \cV(\bx^C_t)\right]$, $u_0 = \bbE\left[ \cV(\bx^C_0)\right]$.  By Gr\"{o}nwall's inequality, we get
\begin{align*}
  u_t &\le e^{-2\eta \mu t} u_0 + \eta c_{\ref{LyapunovConst}} \int_0^t e^{-2\eta \mu s} ds \\
      &= e^{-2\eta \mu t}  u_0 + \frac{c_{\ref{LyapunovConst}}}{2\mu}\left(1 -e^{-2\eta \mu t}\right) \\
      &\le u_0 + \frac{c_{\ref{LyapunovConst}}}{2\mu}.
\end{align*}
Recalling that $u_t = \frac{1}{2}\bbE[\|\bx_t\|^2]$ and $\bbE[\|\bx_0\|^2]\le \varsigma$ completes the proof. 
\hfill$\blacksquare$

\subsection{Discrete-time bounds}
\label{app:dis-timeBounds}
Here we derive a uniform bound on $\bbE[\|\bx_k^A\|^2]$.

\const{AlgBound}
\newcommand{\AlgBoundVal}{\frac{4}{\mu} \left(\frac{ n}{\beta}+(\ell+\mu)R^2 + (2+R)\|\nabla_x \bar f(0)\|   + \left(8 \ell^2 + \frac{1}{\mu}\right)\ell^2\cM_2(\bz)\right)}
\begin{lemma}
  \label{lem:algBound}
  Assume that $\bbE[\|\bx_0^A\|^2] \le \varsigma$ and that $\eta\le \min\left\{1, \frac{\mu}{4\ell^2} \right\}$. There is a constant, $c_{\ref{AlgBound}}$ such that for all $k\ge 0$, we have that
  $$
  \bbE[\|\bx_k^A\|^2]\le \varsigma + c_{\ref{AlgBound}}.
  $$
  The constant is given by
  $$
  c_{\ref{AlgBound}}=\AlgBoundVal
  $$
\end{lemma}

\paragraph{Proof}
Using non-expansiveness of the projection and then expanding the square of the norm gives:
\begin{align*}
\bbE \left[ \|\bx_{t+1}^A\|^2 \right]
  &=
    \bbE\left[
    \left\|
    \Pi_{\cK}\left(
    \bx_t^A - \eta \nabla_x f(\bx_t^A,\bz_t) + \sqrt{\frac{2\eta}{\beta}}\hat \bw_t
    \right) - \Pi_{\cK}(0)
    \right\|^2 \right] \\
  &\le
    \bbE\left[
    \left\|
    \bx_t^A - \eta \nabla_x f(\bx_t^A,\bz_t) + \sqrt{\frac{2\eta}{\beta}}\hat \bw_t
    \right\|^2 \right] \\
  &= \bbE\left[\|\bx_t^A\|^2 + \eta^2 \|\nabla_x f(\bx_t^A,\bz_t)\|^2 
    -2\eta (\bx_t^A)^\top \nabla_x f(\bx_t^A,\bz_t)\right] + \frac{2n\eta}{\beta}.
\end{align*}

Now we bound the term $\bbE\left[\|\nabla_x f(\bx_t^A,\bz_t)\|^2\right]$. 
For any $x\in\cK$, we have that
\begin{align}
\label{eq:gradientSquareBound1}
\nonumber
  \|\nabla_x f(x,z)\|^2 &= \|\nabla_x f(x,z)-\nabla_x f(0,z)+ \nabla_x f(0,z)\|^2\\
  &\le 2\|\nabla_x f(0,z)\|^2 + 2\ell^2\|x\|^2.
\end{align}

This leads to:
\begin{multline}
  \nonumber
    \bbE\left[\|\bx_{t+1}^A\|^2\right] \le \left(1+2\ell^2\eta^2\right) \bbE\left[ \|\bx_{t}^A\|^2 \right]
    \\
    +\left(\frac{2\eta n}{\beta} + 2\eta^2\bbE[ \|\nabla_x f(0,\bz_t)\|^2] \right)
    -2\eta \bbE\left[
      (\bx_t^{A})^\top \nabla_x  f(\bx_t^A,\bz_t)
      \right].
\end{multline}

To bound the term $\bbE[\|\nabla_x f(0,\bz_t)\|^2]$, note that $\nabla_x \bar f(0)=\bbE[\nabla_x f(0,\hat\bz_t)]$, where $\hat \bz_t$ is identically distributed to $\bz_t$ and independent of $\bz_t$.
\begin{align}
\label{eq:gradientSquareBound2}
\nonumber
  \bbE[\|\nabla_x f(0,\bz_t)\|^2] &= \bbE\left[
                                    \|
          \nabla_x \bar f(0)+\nabla_x f(0,\bz_t)-\bbE[\nabla_x f(0,\hat\bz_t)]\|^2
                                    \right] \\
                                    \nonumber
  & \le 2\|\nabla_x \bar f(0)\|^2+ 2\bbE[\|\nabla_x f(0,\bz_t)-\bbE[\nabla_x f(0,\hat\bz_t)]\|^2] \\
  \nonumber
  & \overset{\textrm{Jensen}}{\le} 
   2\|\nabla_x \bar f(0)\|^2+ 2\bbE[\|\nabla_x f(0,\bz_t)-\nabla_x f(0,\hat\bz_t)\|^2] \\
   \nonumber
  & \le 2\|\nabla_x \bar f(0)\|  + 2\ell^2 \bbE[\|\bz_t-\hat\bz_t\|^2] \\
  \nonumber
  &\le 2\|\nabla_x \bar f(0)\|  + 4\ell^2 \bbE[\|\bz_t\|^2+\|\hat\bz_t\|^2] \\
  &\le 2\|\nabla_x \bar f(0)\|  + 8 \ell^2 \cM_2(\bz),
\end{align}
where $\cM_2(\bz)$ is a bound on $\bbE[\|\bz_t\|^2]$ from (\ref{eq:Mbounded}).

So, we have a bound of the form
\begin{multline}
    \label{eq:meanSquareIntermediateZ}
    \bbE\left[\|\bx_{t+1}^A\|^2\right] \le \left(1+2\ell^2\eta^2\right) \bbE\left[ \|\bx_{t}^A\|^2 \right]
    \\
    +\left(\frac{2\eta n}{\beta} + \eta^2 \left(4\|\nabla_x \bar f(0)\|  + 16 \ell^2 \cM_2(\bz)\right) \right)
    -2\eta \bbE\left[
      (\bx_t^{A})^\top \nabla_x  f(\bx_t^A,\bz_t)
      \right].
\end{multline}

To bound the inner product term, note that
\begin{align*}
  \bbE\left[
  (\bx_t^A)^\top \nabla_x f(\bx_t^A,\bz_t)
  \right]
  &=
  \bbE\left[
  (\bx_t^A)^\top \left(\nabla_x f(\bx_t^A,\bz_t) - \nabla_x f(\bx_t^A,\hat\bz_t)\right)
    \right]  +  \bbE\left[
  (\bx_t^A)^\top \nabla_x f(\bx_t^A,\hat\bz_t)
  \right]
  \\
  &=
    \bbE\left[
  (\bx_t^A)^\top \left(\nabla_x f(\bx_t^A,\bz_t) - \nabla_x f(\bx_t^A,\hat\bz_t)\right)
    \right]  +  \bbE\left[
  (\bx_t^A)^\top \nabla_x \bar f(\bx_t^A)
  \right].
\end{align*}
The second equality follows because $\hat\bz_t$ is independent of $\bx_t^A$ and identically distributed to $\bz_t$. So, we can use the Cauchy-Schwartz inequality on the first term on the right and (\ref{eq:smallXInnerProduct}) on the second term to give:
\begin{align*}
  \bbE\left[
  (\bx_t^A)^\top \nabla_x f(\bx_t^A,\bz_t)
  \right]
  \ge -\ell \bbE[\|\bx_t^A\| \|\bz_t-\hat\bz_t\|] +\mu\bbE[\|\bx_t^A\|^2]
-\left((\ell+\mu)R^2 + R\|\nabla_x \bar f(0)\|\right).
\end{align*}

  Using a completing-the-squares argument shows that for any numbers $a$ and $b$
  \begin{align*}
    \frac{\mu}{2}a^2 -\ell ab
    &=
      \frac{\mu}{2}\left(a - \frac{\ell}{\mu}b\right)^2 - \frac{\ell^2}{2\mu}b^2  \\
    &\ge -\frac{\ell^2}{2\mu}b^2.
  \end{align*}

 Setting $a = \|\bx_t^A\|$ and $b = \|\bz_t-\hat\bz_t\|$ leads to a bound of the form
\begin{align}
\label{eq:innerProductBound}
\nonumber
  \bbE\left[
  (\bx_t^A)^\top \nabla_x f(\bx_t^A,\bz_t)
  \right]&\ge \frac{\mu}{2}\bbE[\|\bx_t^A\|^2]-\frac{\ell^2}{2\mu}\bbE[\|\bz_t-\hat\bz_t\|^2]
-\left((\ell+\mu)R^2 + R\|\nabla_x \bar f(0)\|\right) \\
         &\ge
 \frac{\mu}{2}\bbE[\|\bx_t^A\|^2]-\frac{\ell^2}{\mu} \cM_2(\bz) 
-\left((\ell+\mu)R^2 + R\|\nabla_x \bar f(0)\|\right).
\end{align}

Plugging the new bounds into (\ref{eq:meanSquareIntermediateZ}) gives
\begin{multline}
  \nonumber
  \bbE\left[\|\bx_{t+1}^A\|^2\right]\le  
  \left(1-\mu\eta+2\ell^2\eta^2\right) \bbE\left[ \|\bx_{t}^A\|^2 \right] 
  \\
  + \left(\frac{2\eta n}{\beta} + \eta^2\left( 4\|\nabla_x \bar f(0)\|  + 16 \ell^2 \cM_2(\bz)\right) \right)
  +2\eta \left(  \frac{\ell^2}{\mu}\cM_2(\bz) +\left((\ell+\mu)R^2 + R\|\nabla_x \bar f(0)\|\right)\right)
\end{multline}

Note that if $\eta \le \frac{\mu}{4\ell^2}$, then
$$
1-\mu\eta+2\ell^2\eta^2\le 1-\frac{\mu\eta}{2}.
$$
Furthermore, if $\eta \le 1$, we get the simplified bound:
\begin{multline}
  \label{eq:algIterateBound}
    \bbE\left[\|\bx_{t+1}^A\|^2\right]\le  
  \left(1-\frac{\mu\eta}{2}\right) \bbE\left[ \|\bx_{t}^A\|^2 \right] 
  \\
  + 2\eta\left(\frac{ n}{\beta} + 2\|\nabla_x \bar f(0)\|  + 8 \ell^2 \cM_2(\bz) +  \frac{\ell^2}{\mu}\cM_2(\bz) +
(\ell+\mu)R^2 + R\|\nabla_x \bar f(0)\|
  \right).
\end{multline}

Now for any $a\in [0,1)$ and any $b\ge 0$, if $u_t\ge 0$ satisfies
$$
u_{t+1} \le au_t  +b
$$
then
\begin{align*}
  u_{t} &\le a^t u_0 + b\sum_{k=0}^{t-1} a^k \\
        &= a^t u_0 + b\frac{1-a^t}{1-a} \\
        &\le u_0 + \frac{b}{1-a}
\end{align*}
Applying this bound  to $\bbE[\|\bx_t^A\|^2]$ and using that $\bbE[\|\bx_0^A\|^2]\le \varsigma$ gives

\begin{align*}
  \nonumber
  \bbE[\|\bx_t^A\|^2] \le \varsigma + 
  \frac{4}{\mu} \left(\frac{ n}{\beta}+(\ell+\mu)R^2 + (2+R)\|\nabla_x \bar f(0)\|   + \left(8 \ell^2 + \frac{1}{\mu}\right)\ell^2\cM_2(\bz)\right).
\end{align*}
\hfill$\blacksquare$


\section{Stochastic contraction analysis}
\label{sec:contraction}
In this Appendix, we prove Lemma~\ref{lem:convergeToStationary}.
\subsection{Contraction for the reflected SDEs}

We extend the analysis of standard SDEs from \cite{eberle2016reflection} to the case of reflected SDEs. The main idea of \cite{eberle2016reflection} is to construct a specialized metric over $\bbR^n$ and corresponding Wasserstein distance under which contraction rates can be computed. In the context of this paper, we only use Euclidean norm to construct the metric, whereas in \cite{eberle2016reflection}, both Euclidean and a second norm were used to construct the specilized metric. Using just one norm leads to some simplifications. Our choice of reflection term in the coupling process is also slightly different, leading to further simplifications. 

In the following, we firstly examine the contractivity properties of the generalized reflected SDEs and then associate the generalized process with the original process from (\ref{eq:projectedLangevin}).

Let $\cK$ be a closed convex subset of $\bbR^n$ and consider a reflected stochastic differential equations of the form:
\begin{equation}\label{eq:diffX}
d\bx_t = H(\bx_t)dt + G d\bw_t - \bv_td\bmu(t) ,
\end{equation}
where $G$ is an invertible $n\times n$ matrix with minimum singular value $\sigma_{\min}(G)$, $\bw_t$ is a standard Brownian motion, and $-\int_0^t \bv_sd\bmu(s)$ is a reflection term that ensures that $\bx_t\in\cK$ for all $t\ge 0$. (We are slightly abusing notation, since here $\bx_t$ denotes the solution to a general RSDE, and is not the iterates of the original algorithm from (\ref{eq:projectedLangevin}).)

Following \cite{eberle2016reflection},  we construct a function $\delta: [0,+\infty) \rightarrow \bbR $ such that $\delta(0) = 0$, $\delta'(0) = 1$, $\delta'(r) >0$,  and $\delta''(r) \le 0$ for all $r \ge 0 $. With these properties, it can be shown that $\delta(\|x-y\|)$ forms a metric over $\cK$. The particular metric is constructed so that the dynamics are contractive with respect to the corresponding Wasserstein distance. 

Assume there exists a continuous function $\kappa(r): [0, +\infty) \rightarrow \bbR$ such that for any $ x, y \in \bbR^n, x \neq y$, 
\begin{equation}
  \label{eq:oneSided}
  (x-y)^\top\left( H(x)-H(y)\right) \le \kappa(\|x-y\|)\|x-y\|^2.
\end{equation}
Also, assume that 
\begin{equation}
\lim \sup \kappa(r) < 0.
\end{equation}
This implies that there is a postive constant, $R_0$, and a negative constant $\bar \kappa$, such that $  \kappa(r) \le \bar \kappa <0$ for all $r>R_0$.

We choose 
$$R_1 = \frac{R_0}{2} + \frac{1}{2}\sqrt{R_0^2 - \frac{16 \sigma_{min}(G)^2e^{h(R_0)}}{\bar \kappa}} > R_0,$$
and define $\delta$ via the following chain of definitions:
\begin{subequations}
    \label{eq:deltaDef}
    \begin{align}
    \label{eq:delta}
\delta(r) &= \int_0^{r} \varphi(s) g(s) ds\\
\label{eq:g}
g(r) &= 1- \frac{\xi}{2} \int_0^{r \wedge R_1} \Phi(s)\varphi(s)^{-1} ds \\
\label{eq:xi}
\xi^{-1} &= \int_0^{R_1} \Phi(s)\varphi(s)^{-1} ds \\
\label{eq:Phi}
\Phi(r) &= \int_0^r \varphi(s) ds \\
      \label{eq:phi}
      \varphi(r) &= e^{-h(r)}\\
      \label{eq:lipschitzIntegral}
h(r) &= \frac{1}{2\sigma_{\min}(G)^2}\int_0^r s(\kappa(s) \vee 0) ds.
  \end{align}
  \end{subequations}
In the above definition, we use the shorthand notation $a \wedge b = \min \{a,b \}$ and $a \lor b = \max \{a,b \}$.

The details on the choices of $R_0$ and $R_1$ will be presented during the proof of Theorem \ref{thm:contraction_general} for the general reflection coupling related to \eqref{eq:diffX} and Corollary \ref{cor:contraction} for the specific reflection coupling related to \eqref{eq:AvecontinuousProjectedLangevin}.
 
%
%
As discussed above, $\delta(\|x-y\|)$ is a metric. See \cite{eberle2016reflection} for details.
The corresponding Wasserstein distance is defined by 
\begin{equation}
\nonumber
 W_{\delta}(P,Q) = \inf_{\Gamma \in \mathfrak{C}(P,Q)}  \int_{\cK \times \cK} \delta(\|x-y \|) d \Gamma(x,y)
\end{equation}   
Here, $\mathfrak{C}$ is the couplings between $P$ and $Q$.

To get an explicit form of the constant factor in Lemma~\ref{lem:convergeToStationary}, we  use the following theorem, which is analogous to Corollary 2 of \cite{eberle2016reflection}.

\begin{theorem}
  \label{thm:contraction_general}
  If $\bx_t^1$ and $\bx_t^2$ are two solutions to (\ref{eq:diffX}), then for all $0\le s\le t$, their laws satisfy 
  $$
  W_{\delta}(\cL(\bx_t^1),\cL(\bx_t^2))\le e^{-\tilde a (t-s)} W_{\delta}(\cL(\bx_s^1),\cL(\bx_s^2))
  $$
  where $\tilde a = \xi \sigma_{\min}(G)^2$.
\end{theorem}

\paragraph{Proof}
The proof closely follows the proof of Theorem 1 from \cite{eberle2016reflection} with constraints handled similar to works in \cite{lamperski2021projected,lekang2021wasserstein_accepted}. 
The key is to create an explicit coupling between $\bx_t^1$ and $\bx_t^2$, which is known as a \emph{reflection coupling} \cite{lindvall1986coupling}.

To define the reflection coupling, let $\btau$ be coupling time: $\btau = \inf \left\{ t \vert \bx_t^1 = \bx_t^2 \right\}$.
Let $\br_t = \|\bx_t^1 -\bx_t^2\|$, $\bu_t = (\bx_t^1 -\bx_t^2)/{\br_t}$.  Then the reflection coupling between $\bx_t^1$ and $\bx_t^2$ is defined by:
\begin{subequations}
\label{eq:reflectionCoupling}
	\begin{align}
	d\bx_t^1 &= H(\bx_t^1)dt + G d\bw_t - \bv^1_td\bmu^1(t)\\
	d\bx_t^2 &= H(\bx_t^2)dt + (I - 2\bu_t \bu_t^\top \indic (t<\btau)) G d\bw_t - \bv^2_td\bmu^2(t)
	\end{align}
\end{subequations}
where $-\int_0^t \bv^1_sd\bmu^1(s)$ and $-\int_0^t \bv^2_s d\bmu^2(s)$ are reflection terms that ensure that $\bx_t^1\in\cK$ and $\bx_t^2\in\cK$ for all $t\ge 0$.

The processes from (\ref{eq:reflectionCoupling}) define a valid coupling since $\int_0^T (I - 2\bu_t \bu_t^\top \indic(t<\btau)) G d\bw_t$ is a Brownian motion by L\'evy's characterization. 

The main idea is to show that with the specially constructed metric (\ref{eq:deltaDef}), there will be a constant $\tilde{a}$ such that $e^{\tilde{a}t}\delta(\br_t)$ is a supermartingale.
Then, the definition of $W_{\delta}$ and the supermartingale property shows that
\begin{align*}
W_{\delta} (\cL(\bx_t^1),\cL(\bx_t^2)) \le \bbE\left[ \delta(\br_t)\right] 
\le e^{-\tilde{a}(t-s)} \bbE[\delta( \br_s )]
\end{align*}  
Since this bound holds for all couplings of the laws $ \cL(\bx_s^1)$ and $\cL(\bx_s^2)$, it must hold for the optimal coupling, and so
\begin{align*}
W_{\delta} (\cL(\bx_t^1),\cL(\bx_t^2)) \le  e^{-\tilde{a}(t-s)} W_{\delta}(\cL(\bx_s^1),\cL(\bx_s^2)),
\end{align*}
which is the desired conclusion.

Therefore, to complete the proof, we must show that $e^{\tilde{a}t}\delta(\br_t)$ is a supermartingale, which is to ensure that this process is non-increasing on average. Recall that $\btau$ is the coupling time, so that $e^{\tilde{a}t}\delta(\br_t) =0$ for $t \ge \btau$. So we want to bound the behavior of the process for all $t <\btau$. Specifically, it is required to show that non-martingale terms of $d \left( e^{\tilde{a} t } \delta(\br_t)\right)$  are non-positive. 
By It\^o's formula, we have that
\begin{align*}
d \left( e^{\tilde{a} t } \delta( \br_t)\right) =  e^{\tilde{a} t } \left( \tilde{a} \delta(r)dt + \delta'(r) d \br_t 
+ \frac{1}{2} \delta''( r) (d \br_t)^2 \right).
\end{align*}

To achieve the desired differential, we have to derive the terms $d \br_t$ and $(d \br_t)^2$.
%
\begin{align*}
d \br_t &=  \bu_t^\top \left( d\bx_t^1 - d \bx_t^2\right)  \\
&= \bu_t^\top \left( \left( H(\bx_t^1)- H(\bx_t^2)\right) dt + 2 \bu_t \bu_t^\top G d\bw_t  - \bv^1_td\bmu^1(t) +  \bv^2_td\bmu^2(t) \right)
\end{align*}

The above equation is simplified because $(d\bx_t^1 - d \bx_t^2)^\top (\nabla^2 \br_t) (d\bx_t^1 - d \bx_t^2)=0$.

Also, by assumption we have  
\begin{equation}
  (\bx_t^1-\bx_t^2)^\top\left( H(\bx_t^1)-H(\bx_t^2)\right) \le \kappa(\|\bx_t^1-\bx_t^2\|)\|\bx_t^1-\bx_t^2\|^2.
\end{equation}


By the definition of $\bu_t$ and the facts that $\bv_t^1\in N_{\cK}(\bx_t^1)$ and $\bv_t^2\in N_{\cK}(\bx_t^2)$ imply that 
$- \bu_t^\top \bv^1_td\bmu^1(t) \le 0$ and $\bu_t^\top \bv^2_td\bmu^2(t) \le 0$. It follows that
and the assumption (\ref{eq:oneSided})
gives
\begin{align*}
d \br_t \le \kappa(r) r dt + 2 \bu_t^\top G d\bw_t. 
\end{align*}

Now, since the terms that were dropped in the inequality have bounded variation, we  have that
\begin{align*}
(d \br_t)^2 = 4\bu_t G G^\top \bu_t dt \ge 4 \sigma_{\min}(G)^2 dt.
\end{align*}

By construction $\delta'(r)\ge 0$ and $\delta''(r)\le 0$, and so It\^o's formula gives
\begin{align*}
d \left( e^{\tilde{a} t } \delta(\br_t)\right) 
&\le dt e^{\tilde{a}t } \left(  \tilde{a} \delta(r) + \delta'(r)\kappa(r) r + \delta''(r) 2 \sigma_{\min}(G)^2 \right)   + \bm_t\\
&= 2 \sigma_{\min}(G)^2  e^{\tilde{a} t } dt \left( \frac{\tilde{a}}{2 \sigma_{\min}(G)^2} \delta(r) + \frac{k(r) r}{2 \sigma_{\min}(G)^2} \delta'(r) +\delta''(r) \right) + \bm_t,
\end{align*}
where $\bm_t$ denotes a local martingale.

So it suffices to pick certain $\tilde{a}$ and $R_1$ to ensure that for all $r \ge 0 $, the following holds:
\begin{align}
\label{eq:driftTerm}
\frac{\tilde{a}}{2 \sigma_{\min}(G)^2} \delta(r) + \frac{\kappa(r) r}{2 \sigma_{\min}(G)^2} \delta'(r) +\delta''(r) \le 0.
\end{align}

Recall that 
\begin{subequations}
\begin{align*}
\delta''(r) &= \varphi'(r)g(r) + g'(r)\varphi(r) \\
&= -\frac{1}{2\sigma_{\min} (G)^2} r(\kappa(r) \vee 0 )\delta'(r) -\frac{\xi}{2} \Phi(r) \indic(r<R_1).
\end{align*}
\end{subequations}
So if we set $\tilde{a} = \xi \sigma_{\min}(G)^2$,  then $\delta(r)\le \Phi(r)$ implies that (\ref{eq:driftTerm}) holds for all $r<R_1$.

The remaining work is to find a sufficient condition under which (\ref{eq:driftTerm}) holds when $r \ge R_1$.

Recall that we assume that there exists $0<R_0$, such that $k(r) <0$ for all $r\ge R_0$. So, if we choose $R_1 > R_0$, we have for all $r\ge R_1$ that $\varphi(r) = \varphi(R_0)$. By definition, $g(r) = \frac{1}{2}$ for all $r\ge R_1$, and so we must also have $\delta'(r) = \frac{1}{2} \varphi(R_0)$.

Therefore, for $r\ge R_1$, (\ref{eq:driftTerm}) becomes
\begin{align*}
\label{driftInft}
\frac{\tilde{a}}{2 \sigma_{\min}(G)^2} \delta(r) + \frac{\kappa(r) r}{2 \sigma_{\min}(G)^2} \frac{1}{2} \varphi(R_0) \le 0.
\end{align*}



So, a sufficient condition for (\ref{eq:driftTerm}) to hold when $r \ge R_1$ is given by:
\begin{subequations}
\label{eq:sufficientarguement}
\begin{align}
&\frac{\tilde a \delta(r)}{2 \sigma_{\min}(G)^2} + \frac{\kappa(r)r}{2 \sigma_{\min}(G)^2} \frac{1}{2} \varphi(R_0) \le 0 \\
\iff & \tilde a \delta(r) + \kappa(r)r \frac{1}{2} \varphi(R_0) \le 0 \\
\iff & \kappa(r)r \frac{1}{2} \varphi(R_0) \le -\tilde a \delta(r) \\
\iff & \kappa(r)r \frac{1}{2} \varphi(R_0) \le -\xi \sigma_{min}(G)^2 \delta(r) \\
\iff  & \kappa(r)r \frac{1}{2} \varphi(R_0) \le -\frac{\sigma_{min}(G)^2 }{\int_0^{R_1} \Phi(s)\varphi(s)^{-1} ds}\delta(r)  
\label{eq:sufficient1}\\
\impliedby & \kappa(r)r \frac{1}{2} \varphi(R_0) \le -\frac{\sigma_{min}(G)^2 }{(R_1-R_0)\Phi(R_1) \varphi(R_0)^{-1} /2} \delta(r) \label{eq:sufficient2}\\
\impliedby  & \kappa(r)r \frac{1}{2} \varphi(R_0) \le -\frac{\sigma_{min}(G)^2 }{(R_1-R_0)\Phi(R_1) \varphi(R_0)^{-1} /2} r \label{eq:sufficient3}\\
\iff & \kappa(r) \le -\frac{4 \sigma_{min}(G)^2 }{(R_1-R_0)\Phi(R_1) } \\
\iff & (R_1-R_0)\Phi(R_1)   \ge -\frac{4 \sigma_{min}(G)^2 }{\kappa(r)} \label{eq:sufficient4}\\
\impliedby & (R_1-R_0)R_1 e^{-h(R_0)}   \ge -\frac{4 \sigma_{min}(G)^2 }{\kappa(r)} \label{eq:sufficient5} \\
\impliedby & (R_1-R_0)R_1 e^{-h(R_0)}   \ge -\frac{4 \sigma_{min}(G)^2 }{\bar \kappa} \label{eq:sufficient6}
\end{align}
\end{subequations}
Note
(\ref{eq:sufficient1}) is implied by (\ref{eq:sufficient2}) because 
for $r>R_0$, $\varphi(r) = \varphi(R_0)$, therefore, $\Phi(r) = \Phi(R_0) + \varphi(R_0)(r-R_0)$ which gives
\begin{align}
\nonumber
\int_0^{R_1} \Phi(s)\varphi(s)^{-1} ds & \ge \int_{R_0}^{R_1}\Phi(s)\varphi(s)^{-1} ds \\ 
\nonumber
&= \int_{R_0}^{R_1} \left( \Phi(R_0) +\varphi(R_0)(s-R_0) \right) \varphi(R_0)^{-1} ds \\
\nonumber
 &= \Phi(R_0)\varphi(R_0)^{-1}(R_1-R_0) + \frac{(R_1-R_0)^2}{2} \\
 \nonumber
 & \ge \frac{\Phi(R_0)\varphi(R_0)^{-1}(R_1-R_0)}{2} + \frac{(R_1-R_0)^2}{2}\\
 \nonumber
 &= (R_1-R_0) \left( \Phi(R_0) + (R_1-R_0)\varphi(R_0)\right) \varphi(R_0)^{-1}/2\\
 &= (R_1-R_0)\Phi(R_1) \varphi(R_0)^{-1} /2.
 \label{eq:xiInverseBound}
\end{align}

Also, (\ref{eq:sufficient2}) is implied by (\ref{eq:sufficient3}) because $\delta(r) < r$.

From (\ref{eq:sufficient4}) to (\ref{eq:sufficient5}), we use:
\begin{align}
\label{eq:PhiBound}
\nonumber
\Phi(R_1) &= \int_0^{R_1} \varphi(s) ds\\
\nonumber
&= \int_0^{R_1} e^{-h(s)}ds \\
\nonumber
&\ge  \int_0^{R_1} e^{-h(R_0)}ds \\
&= R_1 e^{-h(R_0)}.
\end{align}
The implication (\ref{eq:sufficient6}) $\implies$ 
(\ref{eq:sufficient5}) arises because of the assumption that $ \kappa(r) \le \bar \kappa <0$ for all $r>R_0$.

Therefore,  (\ref{eq:driftTerm}) will hold all $r\ge R_1$, as long as $R_1$ satisfies (\ref{eq:sufficient6}). The smallest such $R_1$ is given by
\begin{equation}
\label{R_1}
R_1 = \frac{R_0}{2} + \frac{1}{2}\sqrt{R_0^2 - \frac{16 \sigma_{min}(G)^2e^{h(R_0)}}{\bar \kappa}} > R_0.
\end{equation} 
\hfill$\blacksquare$

We choose our reflection term as $(I - 2\bu_t \bu_t^\top \indic (t<\btau)) G d\bw_t$, while \cite{eberle2016reflection} uses $G(I - 2\be_t \be_t^\top \indic (t<\btau)) d\bw_t$, with  $\be_t = \frac{G^{-1}(\bx_t^1-\bx_t^2)}{\|G^{-1}(\bx_t^1-\bx_t^2)\|}$. Our form of the reflection term leads to mild simplification of some  formulas. 

Now we specialize the result from the previous theorem to the specific case of this paper:

\begin{corollary}
 \label{cor:contraction}
  If $\bx_t^1$ and $\bx_t^2$ are two solutions to \eqref{eq:AvecontinuousProjectedLangevin},
then for all $0\le s\le t$, their laws satisfy 
  $$       
  W_{\delta}(\cL(\bx_t^1),\cL(\bx_t^2))\le e^{-\tilde a (t-s)} W_{\delta}(\cL(\bx_s^1),\cL(\bx_s^2))
  $$
  where $\tilde a = \xi \frac{2\eta}{\beta}$, $R_0=R$, and
  $R_1 = \frac{R}{2} +\frac{1}{2} \sqrt{R^2 + \frac{32}{\mu
      \beta}e^{\frac{\beta \ell R^2}{8}}}$ in the construction of
  $\delta$.
\end{corollary}

\paragraph{Proof}
We can see that \eqref{eq:AvecontinuousProjectedLangevin} is a special case of \eqref{eq:diffX} with
  \begin{align*}
H(x) &= -\eta \nabla_x \bar f(x) \\
G &= \sqrt{\frac{2\eta}{\beta}}I .
\end{align*}

Since we assume that $\bar f$ is $\ell$-Lipschitz and convex outside a ball with radius $R$, we have that \eqref{eq:oneSided} holds with $\kappa(s)=\eta \ell$ for $0\le s<R$ and $\kappa(s) = - \eta \mu$ for $s\ge  R$. Therefore, we can pick $R_0 = R$ to construct the metric \eqref{eq:deltaDef}.

Now, $\sigma_{\min}(G)^2 =\frac{2\eta}{\beta}$ implies that $\tilde a = \xi \frac{2\eta}{\beta}$. Furthermore, the choice of $\kappa(r)$ implies that $h(R_0)=h(R) = \frac{\beta \ell R^2}{8}$.

The choice of $\kappa(r)$ also implies that $\bar\kappa = -\eta \mu$. Thus, the form of $R_1$ is given by plugging terms into \eqref{R_1}.
\hfill$\blacksquare$

%
%

\begin{corollary}
\label{cor:contractionW1}
If $\bx_t^1$ and $\bx_t^2$ are two solutions to \eqref{eq:AvecontinuousProjectedLangevin},
then for all $0\le s\le t$, their laws satisfy 
  $$
  W_{1}(\cL(\bx_t^1),\cL(\bx_t^2))\le 2 \varphi(R)^{-1}e^{-\tilde a (t-s)} W_{1}(\cL(\bx_s^1),\cL(\bx_s^2)).
  $$
\end{corollary}

\paragraph{Proof}
From the special constructed of $\delta$, we that $\delta'(r)$ is monotonically decreasing, and also $\delta'(r)=\delta(R_1)$ for all $r\ge R_1$.  Furthermore, $\delta(r)=\int_0^r \delta'(s) ds \ge \delta'(r) \int_0^r ds = r\delta'(r)$. Thus, for all $r\ge 0$, the following bounds hold:
$$\delta'(R_1)r \le \delta'(r)r \le \delta(r) \le  r$$

These bounds are now used to relate the $W_{\delta}$ and $W_1$ distances:
\begin{align}
  \label{eq:W1Upper}
\delta'(R_1) W_1(\cL(\bx_t^1),\cL(\bx_t^2))\le W_{\delta}(\cL(\bx_t^1),\cL(\bx_t^2)) 
\le W_1(\cL(\bx_t),\cL(\by_t)).
\end{align}
In particular, 
\begin{align}
\label{eq: deltaR1}
\delta'(R_1) = \varphi(R_1)g(R_1) = \frac{1}{2} \varphi(R).
\end{align}
Plugging (\ref{eq: deltaR1}) into the first inequality of (\ref{eq:W1Upper}) gives
\begin{align}
W_1(\cL(\bx_t^1),\cL(\bx_t^2))\le 2 \varphi(R)^{-1} W_{\delta}(\cL(\bx_t^1),\cL(\bx_t^2)) 
\end{align}
And combining with Corollary \ref{cor:contraction} gives 
\begin{align}
W_1(\cL(\bx_t^1),\cL(\bx_t^2))\le 2 \varphi(R)^{-1} e^{-\tilde a (t-s)} W_{\delta}(\cL(\bx_s^1),\cL(\bx_s^2)) 
\end{align}
Finally, utilizing the second inequality of \eqref{eq:W1Upper} gives the desired result.
\hfill$\blacksquare$

\subsection{Proof of Lemma~\ref{lem:convergeToStationary}}

\label{ss:proofLemmaconvergeToStationary}

In Lemma~\ref{lem:gibbs} of Appendix~\ref{sec:gibbs}, we showed that the Gibbs distribution, $\pi_{\beta \bar f}$, defined in (\ref{eq:gibbs}) is invariant for the dynamics of $\bx_t^C$. Thus, setting $\cL(\bx_t^1)=\cL(\bx_t^C)$ and $\cL(\bx_t^2)=\pi_{\beta\bar f}$ in Corollary ~\ref{cor:contractionW1} gives
\begin{equation}
  \label{eq:convergeW1}
  W_{1}(\cL(\bx_t^C),\pi_{\beta \bar f})\le 2 \varphi(R)^{-1} e^{-\tilde a t} W_{1}(\cL(\bx_0^C),\pi_{\beta \bar f}).
\end{equation}


Let $\by$ be distributed according to $\pi_{\beta \bar f}$. For any joint distribution over $(\bx_0^C,\by)$ whose marginals are $\cL(\bx_0^C)$ and $\pi_{\beta \bar f}$, we  have that
\begin{align}
  \nonumber
  W_{1}(\cL(\bx_0^C),\pi_{\beta \bar f})
  \nonumber
  &\le 
  \nonumber
    \bbE[ \|\bx_0^C -\by \| ] \\
    \nonumber
&\le \sqrt{\bbE[ \|\bx_0^C -\by \|^2 ]}\\
\nonumber
&\le \sqrt{\bbE[ 2\|\bx_0^C\|^2 +2\|\by \|^2 ]}\\
\nonumber
&= \sqrt{2\bbE[ \|\bx_0^C\|^2 ] +2\bbE[ \|\by \|^2 ]}\\
\nonumber
&\le \sqrt{ 2\varsigma  + 2\left(\varsigma+ \frac{1}{\mu} c_{\ref{LyapunovConst}}\right)}\\
&\le  \sqrt{\frac{2}{\mu} c_{\ref{LyapunovConst}}} + 2 \sqrt{\varsigma}.
  \label{eq:WrhoBound}
\end{align}

The second to last inequality uses Lemma~\ref{lem:continuousBound}.


Combining \eqref{eq:convergeW1}, \eqref{eq:WrhoBound} shows that
\begin{equation*}
  W_1(\cL(\bx_t^C),\pi_{\beta \bar f}) \le 
  2 \varphi(R)^{-1} e^{-\tilde a t}   \left(   \sqrt{\frac{2}{\mu} c_{\ref{LyapunovConst}}} +  2\sqrt{\varsigma}\right).
\end{equation*}
Thus, the lemma is proved and the constants are given by:
\begin{subequations}
  \label{eq:contractionConstants}
\begin{align}
a &=  \frac{2\xi}{\beta}\\
c_{\ref{contraction_const1}} &=  \contractionConstOneVal  \\
  c_{\ref{contraction_const2}} &= \contractionConstTwoVal
\end{align}
where $\xi$ is given in \eqref{eq:xi}.
\end{subequations}
\hfill$\blacksquare$

\section{Proofs of averaging lemmas}
\label{app:aveLemmas}

\paragraph{Proof of Lemma~\ref{lem:MeanBetween1}}
  Non-expansiveness of the projection and the definitions of $\bx_t^{M,s}$ and  $\bx_t^{B,s}$ show that:
  \begin{align}
    \nonumber
    \MoveEqLeft
    \|\bx_{t+1}^{M,s}-\bx_{t+1}^{B,s}\|^2
   \\ 
    \nonumber
    &  \le
      \left\|
      \bx_t^{M,s}-\bx_t^{B,s} +\eta \left(
      \bbE[ \nabla_x f(\bx_t^{M,s},\bz_t) | \cF_{t-s-1}\lor \cG_t ] -
      \bbE[ \nabla_x f(\bx_t^{M,s},\bz_t) | \cF_{t-s} \lor \cG_t]
      \right)
      \right\|^2 \\
    \nonumber
    &= \|\bx_t^{M,s}-\bx_t^{B,s}\|^2 
     +
      2\eta \left( \bx_t^{M,s}-\bx_t^{B,s}\right)^\top \\ \nonumber
      & \quad{} \left(
      \bbE[ \nabla_x f(\bx_t^{M,s},\bz_t) | \cF_{t-s-1} \lor \cG_t]   -
      \bbE[ \nabla_x f(\bx_t^{M,s},\bz_t) | \cF_{t-s} \lor \cG_t ]
      \right)  \\ 
      &\quad{} \quad{} + \eta^2 \left\|\bbE[ \nabla_x f(\bx_t^{M,s},\bz_t) | \cF_{t-s-1}\lor \cG_t ] 
    -
      \bbE[ \nabla_x f(\bx_t^{M,s},\bz_t) | \cF_{t-s}\lor \cG_t ] \right\|^2.
      \label{eq:meanBetweenExpand}
  \end{align}

 We will show that the second term on the right of (\ref{eq:meanBetweenExpand}) has mean zero, and then we will bound the mean of the third term on the right of (\ref{eq:meanBetweenExpand}).  

  By construction, we have that $\bx_t^{M,s}$ is $\cF_{t-s-1}\lor \cG_{t}$-measurable, while $\bx_t^{B,s}$ is $\cF_{t-s-2}\lor \cG_{t}$-measurable. Thus, the only part of the second term on the right of  (\ref{eq:meanBetweenExpand}) which is not $\cF_{t-s-1}\lor \cG_t$-measurable is $\bbE[ \nabla_x f(\bx_t^{M,s},\bz_t) | \cF_{t-s} \lor \cG_t ]$.
  Therefore, the tower-property gives:
  \begin{align*}
    \MoveEqLeft
    \bbE\left[
\left( \bx_t^{M,s}-\bx_t^{B,s}\right)^\top  \left(
      \bbE[ \nabla_x f(\bx_t^{M,s},\bz_t) | \cF_{t-s-1}\lor \cG_t ]  -
      \bbE[ \nabla_x f(\bx_t^{M,s},\bz_t) | \cF_{t-s}\lor \cG_t ]
    \right) \right] \\
    &=
\bbE\left[
      \left( \bx_t^{M,s}-\bx_t^{B,s}\right)^\top \left(
      \bbE[ \nabla_x f(\bx_t^{M,s},\bz_t) | \cF_{t-s-1}\lor \cG_t ]  \right. \right. \\
      & \left. \left. \quad \quad \quad \quad\quad \quad-
      \bbE\left[\bbE[ \nabla_x f(\bx_t^{M,s},\bz_t) | \cF_{t-s}\lor \cG_t ]
      \right)\middle | \cF_{t-s-1}\lor \cG_t \right] \right] \\
    &=0.
  \end{align*}

  Now we focus on bounding the mean of the third term on the right of (\ref{eq:meanBetweenExpand}). Recall that $\bx_t^{M,s}$ is $\cF_{t-s-1}\lor \cG_t$-measurable. Furthermore, since $\cF_{t-s}^+$ is independent of $\cF_{t-s}\lor \cG_t$, it must also be independent of $\cF_{t-s-1}\lor \cG_t$ because $\cF_{t-s-1}\subset \cF_{t-s}$. It follows that
  \begin{align*}
\bbE[\nabla_x f(\bx_t^{M,s},\bbE[\bz_t|\cF_{t-s}^+]) | \cF_{t-s}\lor \cG_t] =  \bbE[\nabla_x f(\bx_t^{M,s},\bbE[\bz_t|\cF_{t-s}^+]) | \cF_{t-s-1}\lor \cG_t].
  \end{align*}

  Thus, adding and subtracting $\bbE\left[\nabla_x f(\bx_t^{M,s},\bbE[\bz_t|\cF_{t-s}^+])\middle|\cF_{t-s}\lor \cG_t \right]$ gives
  
  \begin{align}
    \nonumber
    \MoveEqLeft[0]
    \left\|\bbE[ \nabla_x f(\bx_t^{M,s},\bz_t) | \cF_{t-s-1}\lor \cG_t ]  -
    \bbE[ \nabla_x f(\bx_t^{M,s},\bz_t) | \cF_{t-s}\lor \cG_t ] \right\|^2 \\
    \nonumber
    &\le 2  \left\|\bbE\left[ \nabla_x f(\bx_t^{M,s},\bz_t)- \nabla_x f(\bx_t^{M,s},\bbE[\bz_t | \cF_{t-s}^+] )  \middle| \cF_{t-s-1}\lor \cG_t \right] \right\|^2 \\
    \label{eq:conditionalSplit}
    &\quad \quad + 2 
      \left\|\bbE\left[ \nabla_x f(\bx_t^{M,s},\bz_t)  - \nabla_x  f(\bx_t^{M,s},\bbE[\bz_t|\cF_{t-s}^+])
      \middle| \cF_{t-s}\lor \cG_t \right]
      \right\|^2.
  \end{align}
  
  To bound the second term on the right of (\ref{eq:conditionalSplit}), we have
  \begin{align*}
    \MoveEqLeft
    \bbE\left[
\left\|\bbE\left[ \nabla_x f(\bx_t^{M,s},\bz_t)  - \nabla_x  f(\bx_t^{M,s},\bbE[\bz_t|\cF_{t-s}^+])
      \middle| \cF_{t-s}\lor \cG_t \right]
      \right\|^2
    \right]
 \\
    &\overset{\textrm{Jensen}}{\le}
       \bbE\left[   \left\|\nabla_x f(\bx_t^{M,s},\bz_t)  -
\nabla_x f(\bx_t^{M,s},\bbE[\bz_t|\cF_{t-s}^+])
      \right\|^2 \right]  \\
    &\overset{\textrm{Lipschitz}}{\le} \ell^2 \bbE\left[\|
      \bz_t - \bbE[\bz_t|\cF_{t-s}^+\|^2
      \right] \\
    &\le \ell^2 \psi_2(s,\bz)^2.
  \end{align*}
  Here  $\psi_2(s,\bz)$ was defined in \eqref{eq:influenceTau}.

  The first term on the right  of (\ref{eq:conditionalSplit}) is bounded by analogous calculations with $\cF_{t-s-1}$ used in place of $\cF_{t-s}$, and gives rise to the same bound of $\ell^2 \psi(s,\bz)^2$.

  Plugging these bounds into (\ref{eq:meanBetweenExpand}) shows that 

  \begin{equation*}
    \label{eq:meanBetweenAve}
    \bbE\left[
    \|\bx_{t+1}^{M,s}-\bx_{t+1}^{B,s}\|^2
  \right]
  \le     \bbE\left[
    \|\bx_{t}^{M,s}-\bx_{t}^{B,s}\|^2
  \right]
  +4\eta^2 \ell^2 \psi_2(s,\bz)^2
  \end{equation*}

  Iterating \eqref{eq:meanBetweenAve} $t$ times and using the fact that $\bx_0^{B,s}=\bx_0^{M,s}$, shows that
  \begin{equation*}
    \bbE\left[\|\bx_t^{M,s}-\bx_t^{B,s}\|^2 \right] \le 4\eta^2 t \ell^2 \psi_2(s,\bz)^2.   
  \end{equation*}
  
 Using the fact that $$\bbE[\|\bx_t^{M,s}-\bx_t^{B,s}\|]\le \sqrt{\bbE\left[\|\bx_t^{M,s}-\bx_t^{B,s}\|^2 \right]}$$ gives the result. 
\hfill$\blacksquare$


\paragraph{Proof of Lemma~\ref{lem:MeanBetween2}}
Non-expansiveness of the projection and the definitions of $\bx_t^{B,s}$ and  $\bx_t^{M,s+1}$, shows that
  \begin{multline}
    \nonumber
    \|\bx_{t+1}^{B,s}-\bx_{t+1}^{M,s+1}\|
    \\
    \nonumber
    \le
      \left\|\bx_{t}^{B,s}-\bx_t^{M,s+1}+\eta \left( \bbE[ \nabla_x f(\bx_t^{M,s+1},\bz_t) | \cF_{t-s-1} \lor \cG_t ] 
      - \bbE[ \nabla_x f(\bx_t^{M,s},\bz_t) | \cF_{t-s-1} \lor \cG_t ]
      \right)
    \right\|.
    \end{multline}
Let $\|\bx\|_2=\sqrt{\bbE[\|\bx\|^2]}$ denote the $2$-norm over random vectors. The triangle inequality then implies that 
\begin{align}
\nonumber
  \hspace{-10pt}
  \nonumber
      &\left\|\bx_{t}^{B,s}-\bx_t^{M,s+1}+\eta \left( \bbE[ \nabla_x f(\bx_t^{M,s+1},\bz_t) | \cF_{t-s-1} \lor \cG_t ] - \bbE[ \nabla_x f(\bx_t^{M,s},\bz_t) | \cF_{t-s-1} \lor \cG_t ]
      \right)
    \right\|_2  \\ 
    \nonumber &\le
       \left\|\bx_{t}^{B,s}-\bx_t^{M,s+1}\right\|_2
       +\eta \left\|\bbE[ \nabla_x f(\bx_t^{M,s+1},\bz_t) -\nabla_x f(\bx_t^{M,s},\bz_t)  | \cF_{t-s-1} \lor \cG_t ]
    \right\|_2.
             \label{eq:twoNormSplit}
  \end{align}
  For any random vector, $\bx$, and any $\sigma$-algebra, $\cF$, Jensen's inequality followed by the tower property implies that $\bbE[\|\bbE[\bx|\cF]\|^2]\le \bbE[\|\bx\|^2]$. Applying this fact to the second term on the right of (\ref{eq:twoNormSplit}) and then using the Lipschitz property shows that 
  \begin{align*}
    \left\|\bbE[ \nabla_x f(\bx_t^{M,s+1},\bz_t) -\nabla_x f(\bx_t^{M,s},\bz_t)  | \cF_{t-s-1} \lor \cG_t ]
    \right\|_2
    \le \ell \|\bx_t^{M,s+1}-\bx_t^{M,s}\|_2. 
  \end{align*}
  Plugging this bound into (\ref{eq:twoNormSplit}) then adding and subtracting $\bx_t^{B,s}$ gives:
  \begin{align}
    \nonumber
    \MoveEqLeft[0]
    \|\bx_{t+1}^{B,s}-\bx_{t+1}^{M,s+1}\|_2
    \\
    \nonumber
    &\le \|\bx_{t}^{B,s}-\bx_t^{M,s+1}\|_2+\eta \ell \|\bx_t^{M,s+1}-\bx_t^{M,s}\|_2 \\
    \nonumber
    &\le \|\bx_{t}^{B,s}-\bx_t^{M,s+1}\|_2+\eta \ell \|\bx_t^{M,s+1}-\bx_t^{B,s}\|_2 + \eta \ell \|\bx_t^{B,s}-\bx_t^{M,s}\|_2
    \\
    &= (1+\eta\ell) \|\bx_{t}^{B,s}-\bx_t^{M,s+1}\|_2 + \eta \ell \|\bx_t^{B,s}-\bx_t^{M,s}\|_2. 
  \end{align}
  Using the fact that $\bx_0^{B,s}=\bx_0^{M,s+1}$ and iterating this inequality shows that:
  \begin{align*}
   & \|\bx_t^{B,s}-\bx_t^{M,s+1}\|_2 \\
   &\le \eta \ell \sum_{k=0}^{t-1}(1+\eta \ell)^{k} \|\bx_{t-k}^{B,s}-\bx_{t-k}^{M,s}\|_2\\
                                        &\overset{\textrm{Lemma~\ref{lem:MeanBetween1}}}{\le}  \left(2\ell \psi_2(s,\bz) \eta \sqrt{t}\right) \eta \ell \sum_{k=0}^{t-1}(1+\eta\ell)^k\\
    &=
\left(2\ell \psi_2(s,\bz) \eta \sqrt{t}\right) \left((1+\eta \ell)^t-1\right) \\
    &\le
\left(2\ell \psi_2(s,\bz) \eta \sqrt{t}\right)  \left(e^{\eta t \ell}-1\right).
  \end{align*}
  The final inequality follows by taking logarithms and using the fact that $\log(1+\eta \ell)\le \eta \ell$. 
\hfill$\blacksquare$


\section{Discretization bounds}
\label{app:disBounds}
\paragraph{Proof of Lemma~\ref{lem:C2D}} 
Recall that $\by_t^D= \by_{\floor{t}}^C$ and so for all $k \in \bbN$, $\by_k^D= \by_{k}^C$. 
By the construction of Skorokhod solutions to the process $\bx_t^C$ and $\bx_t^D$, and 
using Theorem $\ref{thm:skorokhodConst}$, we have for all $  k \in \bbN$
\begin{align*}
	\left\|\bx_k^C - \bx_k^D \right\| 
	\le (c_{\ref{diamBound}}+1) \sup_{0 \le s \le k } \left\| \by_s^C - \by_{\lfloor s \rfloor}^C \right\|.
\end{align*}
Since
\begin{align*}
\by_t^C = \bx_0^C - \eta \int_0^t \nabla_x \bar{f}(\bx_s^C) ds + \sqrt{\frac{2 \eta}{\beta} } \bw_t,
\end{align*}
the triangle inequality implies that
\begin{align*}
	\left\|\bx_k^C - \bx_k^D \right\|  
	&\le 
	(c_{\ref{diamBound}}+1) \eta \sup_{s \in [0,k]} \left\| \int_{\lfloor s \rfloor}^s \nabla_x \bar{f}(\bx_{\tau}^C) d\tau \right\| 
	+ (c_{\ref{diamBound}}+1) \sqrt{\frac{2 \eta}{\beta} } \sup_{s \in [0,k]} \left\|  \bw_s - \bw_{\lfloor s \rfloor}\right\|.
\end{align*}
$\bbE\left[  \sup_{s \in [0,k]} \left\|  \bw_s - \bw_{\lfloor s \rfloor}\right\|\right]$ is upper bounded by $2n\sqrt{\log(4k)}$. See  Lemma 9 in \cite{lamperski2021projected}. So, the remaining work is to bound the first term on the right.
%

Take the expectation of the first term, we have
\begin{align*}
\label{eq:FirstSupremeBound}
	&\bbE\left[ \sup_{s \in [0,k]}\left\| \int_{\lfloor s \rfloor}^s \nabla_x \bar{f}(\bx_{\tau}^C) d\tau \right\|\right]\\
	&= \bbE\left[ \max_{i=0,\cdots, k-1}\sup_{s \in [i,i+1]}\left\| \int_i^s \nabla_x \bar{f}(\bx_{\tau}^C) d\tau \right\|\right]\\
	&\le \bbE\left[ \left( \sum_{i=0}^{k-1}\left(\sup_{s \in [i,i+1]}\left\| \int_i^s \nabla_x \bar{f}(\bx_{\tau}^C) d\tau \right\|\right)^2 \right)^{1/2}\right]\\
	&\overset{\textrm{Jensen}}{\le} 
	\left( \bbE\left[  \sum_{i=0}^{k-1}\left(\sup_{s \in [i,i+1]}\left\| \int_i^s \nabla_x \bar{f}(\bx_{\tau}^C) d\tau \right\|\right)^2 \right]\right)^{1/2}\\
	&\overset{}{=} 
	\left(   \sum_{i=0}^{k-1}\bbE\left[\left(\sup_{s \in [i,i+1]}\left\| \int_i^s \nabla_x \bar{f}(\bx_{\tau}^C) d\tau \right\|\right)^2 \right]\right)^{1/2}.
\end{align*}
So we want to upper bound the supremum  inside the expectation operation.

We can show for all $ s \in [0,k]$,
\begin{align*}
	\left\| \int_{\lfloor s \rfloor}^s \nabla_x \bar{f}(\bx_{\tau}^C) d\tau \right\|
	&\overset{\textrm{triangle inequality}}{\le}
	\int_{\lfloor s \rfloor}^s \left\| \nabla_x \bar{f}(\bx_{\tau}) \right\| d\tau \\
	&\le  \int_{\lfloor s \rfloor}^{\lfloor s \rfloor+1} \left\| \nabla_x \bar{f}(\bx_{\tau}^C) \right\| d\tau \\
	&\overset{\textrm{Jensen}}{\le} \left( \int_{\lfloor s \rfloor}^{\lfloor s \rfloor+1} \| \nabla_x \bar{f}(\bx_{\tau}^C) \|^2 d\tau \right)^{1/2}.
\end{align*}

Therefore, 
\begin{align*}
	 \bbE\left[ \sup_{s \in [0,k]}\left\| \int_{\lfloor s \rfloor}^s \nabla_x \bar{f}(\bx_{\tau}^C) d\tau \right\|\right]
	&{\le} 
	\left(   \sum_{i=0}^{k-1}\bbE\left[\int_{i}^{i+1} \| \nabla_x \bar{f}(\bx_{\tau}^C) \|^2 d\tau \right]\right)^{1/2}\\
	&\overset{\textrm{Fubini}}{=}
	\left(   \sum_{i=0}^{k-1}\int_{i}^{i+1} \bbE\left[\| \nabla_x \bar{f}(\bx_{\tau}^C1) \|^2 \right]d\tau \right)^{1/2}.
\end{align*}

Here, we can see it suffices to bound $\bbE\left[  \| \nabla_x \bar{f}(\bx_{t}) \|^2 \right]$.

We have assumed that $0 \in \cK$, and so we have
\begin{align*}
	\left\| \nabla_x \bar{f}(\bx_{t}^C) \right\|^2 
	&= \left\| \nabla_x \bar{f}(\bx_{t}^C) - \nabla_x \bar{f}(0) + \nabla_x \bar{f}(0)\right\|^2 \\
	&\le 2\left\| \nabla_x \bar{f}(\bx_{t}^C) - \nabla_x \bar{f}(0)\right\|^2 + 2  \left\| \nabla_x \bar{f}(0)\right\|^2 \\
	&\le 2\ell^2\left\| \bx_{t}^C  \right\|^2 + 2  \left\| \nabla_x \bar{f}(0) \right\|^2.
\end{align*}

Plugging in the bound from Lemma~\ref{lem:continuousBound} shows that 
\begin{align*}
	\bbE\left[  \| \nabla_x \bar{f}(\bx_{t}^C) \|^2 \right] &\le \bbE\left[ 2\ell^2 \| \bx_{t}^C\|^2 + 2  \| \nabla_x \bar{f}(0)\|^2\right]\\
	&= 2 \ell^2 \bbE\left[ \| \bx_{t}^C\|^2 \right]+ 2  \| \nabla_x \bar{f}(0)\|^2\\
	&\le 2\ell^2 \left( \varsigma  +\frac{1}{\mu} c_{\ref{LyapunovConst}} \right) + 2  \| \nabla_x \bar{f}(0)\|^2.
\end{align*}


Therefore, we have 
\begin{align*}
	\bbE\left[ \sup_{s \in [0,k]}\left\| \int_{\lfloor s \rfloor}^s \nabla_x \bar{f}(\bx_{\tau}^C) d\tau \right\|\right]
	&{\le}
	\left(   \sum_{i=0}^{k-1}\int_{i}^{i+1} \bbE\left[\| \nabla_x \bar{f}(\bx_{\tau}^C) \|^2 \right]d\tau \right)^{1/2}\\
	&\le \sqrt{2\ell^2 \left(\varsigma+\frac{1}{\mu} c_{\ref{LyapunovConst}} \right) + 2  \| \nabla_x \bar{f}(0)\|^2 } \sqrt{k} \\
	&\le \left(\sqrt{\frac{2}{\mu}\ell^2 c_{\ref{LyapunovConst}}  + 2 \| \nabla_x \bar{f}(0)\|^2} + \sqrt{2 \ell^2} \sqrt{\varsigma}\right) \sqrt{k}.
\end{align*}

Setting
\begin{align*}
c_{\ref{BoundCtoD1}} &= \BoundCtoDOneVal\\
 c_{\ref{BoundCtoD2}} &= \BoundCtoDTwoVal \\
  c_{\ref{BoundCtoD3}} &= \BoundCtoDThreeVal 
\end{align*}
 where
 $c_{\ref{LyapunovConst}}$ is defined in Lemma~\ref{lem:continuousBound} and $c_{\ref{diamBound}}$ is defined in Theorem~\ref{thm:skorokhodConst}
and combining the bound on the second supreme term gives the desired result.
\hfill$\blacksquare$

\paragraph{Proof of Lemma~\ref{lem:M2D}} 

The argument of bounding $\bx_t^M$ and $\bx_t^D$ closely follows the proof of Lemma 10 in \citep{lamperski2021projected}.
Recall that $\bx_t^M$ is a discretized process and $\bx_t^M = \bx_{\floor{t}}^M$. We also have $\bx_t^M = \cS(\cD(\by_t^M))$, where $\by_t^M$ is defined by
\begin{align*}
 \by_t^M = \bx_0^M - \eta \int_0^t \nabla \bar{f} (\bx_{\floor{s}}^M) ds + \sqrt{\frac{2 \eta}{\beta}} \bw_t.
\end{align*}

The intermediate process $\bx_t^D$ satisfies $\bx_t^D = \cS(\cD(\by_t^C))$, where 
\begin{align*}
 \by_t^C = \bx_0^C - \eta \int_0^t \nabla \bar{f} (\bx_s^C) ds + \sqrt{\frac{2 \eta}{\beta}} \bw_t.
\end{align*}

So in particular,
\begin{align*}
	\bx_{k+1}^M &= \Pi_{\cK} \left(\bx_k^M + \by_{k+1}^M - \by_k^M\right) \\
	&= \Pi_{\cK} \left(\bx_k^M -\eta \nabla \bar{f} (\bx_{k}^M)  + \sqrt{\frac{2 \eta}{\beta}} (\bw_{k+1}- \bw_{k}) \right)\\
	\bx_{k+1}^D &= \Pi_{\cK} \left(\bx_k^D + \by_{k+1}^C - \by_k^C \right) \\
	&= \Pi_{\cK} \left(\bx_k^D -\eta  \int_{k}^{k+1} \nabla \bar{f} (\bx_{s}^C) ds + \sqrt{\frac{2 \eta}{\beta}} (\bw_{k+1}- \bw_{k})\right).
\end{align*}

Define a difference process $$\brho_t = \left( \bx_t^M + \by_t^M - \by_{\floor{t}}^M \right) - \left( \bx_t^D + \by_t^C - \by_{\floor{t}}^C \right).$$

Note that at integers
$k \in \bbN$, $\brho_k =\bx_k^M - \bx_k^D$ and for $t \in [k, k+1)$, we have
$$
\brho_t =\left(\bx_k^M-\by_k^M - \bx_k^D + \by_k^D\right) + \by_t^M - \by_t^C.
$$
It follows that
\begin{align*}
	d \brho_t = d(\by_t^M - \by_t^C) = \eta \left(\nabla \bar{f}(\bx_t^C) - \nabla \bar{f}(\bx_t^M)\right)
\end{align*}

By construction, $\brho_t$ is a continuous bounded variation process on the interval $[k, k+1)$. Thus, when $\brho_t \neq 0 $, we can calculate $d\|\brho_t\|$ using the chain rule.
\begin{align*}
	d \left\|\brho_t \right\| &\overset{\textrm{chain rule}}{=} \left( \frac{\brho_t}{\left\| \brho_t \right\|} \right)^\top d \brho_t \\
	 &= \left( \frac{\brho_t}{\left\| \brho_t \right\|} \right)^\top \eta \left(  \nabla \bar{f}(\bx_t^C) - \nabla \bar{f}(\bx_t^M)\right) dt\\
	& \overset{\textrm{Cauchy-Schwarz}}{\le} \eta \left\| \nabla \bar{f}(\bx_t^C) - \nabla \bar{f}(\bx_t^M) \right\| dt\\
	& \overset{\textrm{Lipschitz}}{\le} \eta \ell \left\| \bx_t^C - \bx_t^M \right\| dt \\
	& = \eta \ell \left\| \bx_t^C - \bx_t^D + \bx_t^D - \bx_t^M \right\| dt \\
	& \overset{\textrm{triangle}}{\le} \eta \ell \left( \left\| \bx_t^C - \bx_t^D  \right\| + \left\|  \bx_t^D - \bx_t^M \right\| \right) dt.
\end{align*}

To include the case that $\brho_t = 0$, we use the Lemma 19 from \citep{lamperski2021projected}. The analysis is as below:

For $t \in [k,k+1)$,
\begin{align*}
\MoveEqLeft
	\left\| \brho_t \right\| = \left\| \brho_k \right\| + \int_{k}^{t} d \left\|\brho_t \right\|\\
	& =\left\| \brho_k \right\| +
	\lim_{\epsilon \downarrow 0} \int_k^{t} \indic \left( \left\| \brho_s \right\| \ge \epsilon \right)d \left\|\brho_s \right\| \\
	& \overset{}{\le} \left\| \brho_k \right\| 
	 + \lim_{\epsilon \downarrow 0} \int_k^{t} \indic \left( \left\| \brho_s \right\| \ge \epsilon \right) \eta \ell \left( \left\| \bx_s^C - \bx_s^D  \right\| + \left\|  \bx_s^D - \bx_s^M \right\| \right) dt \\
	& = (1+ \eta \ell )\left\| \brho_k \right\| +  \eta \ell\int_k^t  \left( \left\| \bx_s^C - \bx_s^D  \right\| \right)  ds. 
\end{align*}
The second equality follows from Lemma 19 from \cite{lamperski2021projected}.
The last equality holds because that $ \brho_k =  \bx_s^M - \bx_s^D  $, $\forall s \in [k, k+1 )$.

Non-expansiveness of the convex projection implies that 
\begin{align}
	\left\| \brho_k \right\| = \left\| \bx_s^M - \bx_s^D \right\| \le \lim_{t\uparrow k } \left\| \brho_t \right\|.
\end{align}

Letting $t=k+1$ gives
\begin{align*}
	\left\| \brho_{k+1} \right\| \le (1+ \eta \ell )\left\| \brho_k \right\| +  \eta \ell\int_k^{k+1}  \left\| \bx_s^C - \bx_s^D  \right\|  ds. 
\end{align*}

Iterating this inequality, and using the assumption that $\bx_0^M = \bx_0^D$ gives
\begin{align*}
	\left\| \brho_k \right\| \le \sum_{i=0}^{k-1} \eta \ell (1+\eta \ell )^{k-i-1} \int_i^{i+1} \left\| \bx_s^C - \bx_s^D  \right\|  ds.
\end{align*}

Taking expectation, and using Lemma \ref{lem:C2D} gives
\begin{align*}
	\bbE\left[ \left\| \brho_k \right\|\right] &\le \sum_{i=0}^{k-1} \eta \ell (1+\eta \ell )^{k-i-1} \int_i^{i+1} \left( (c_{\ref{BoundCtoD1}} + c_{\ref{BoundCtoD2}} \sqrt{\varsigma} ) \eta \sqrt{s} + c_{\ref{BoundCtoD3}} \sqrt{\eta \log(4s)}  \right) ds \\
	&\le \eta\ell    \left( (c_{\ref{BoundCtoD1}} + c_{\ref{BoundCtoD2}} \sqrt{\varsigma} ) \eta \sqrt{k} + c_{\ref{BoundCtoD3}} \sqrt{\eta \log(4k)}  \right) \sum_{i=0}^{k-1}  (1+\eta \ell )^{k-i-1} \\
	&\le  \left( (c_{\ref{BoundCtoD1}} + c_{\ref{BoundCtoD2}} \sqrt{\varsigma} ) \eta \sqrt{k} + c_{\ref{BoundCtoD3}} \sqrt{\eta \log(4k)}  \right) \left( (1+\eta \ell)^k -1 \right)\\
	&\le  \left( (c_{\ref{BoundCtoD1}} + c_{\ref{BoundCtoD2}} \sqrt{\varsigma} ) \eta \sqrt{k} + c_{\ref{BoundCtoD3}} \sqrt{\eta \log(4k)}  \right) \left( e^{\eta \ell k} -1 \right).
\end{align*}
The last inequality is based on the fact that $(1+\eta \ell)^k \le e^{ \eta \ell k}$ for all $\eta \ell  >0$.

Recall that for all $k \in \bbN$, $\brho_k =\bx_k^M - \bx_k^D$, which gives the desired result.
\hfill$\blacksquare$

\section{Conclusion of the proof of Lemma~\ref{lem:AtoC}}
\label{app:switching}

This subsection uses a ``switching'' trick to derive a bound on $W_1(\cL(\bx_k^A),\cL(\bx_k^C))$ that is uniform in time. The essential idea is to utilize a family of processes that switch from the dynamics of $\bx_k^A$ to the dynamics of $\bx_k^C$, and utilize contractivity of the law of $\bx_k^C$ to derive the uniform bounds.  
A similar methodology was utilized in \cite{chau2019stochastic}. 

For $s\ge 0$, 
let $\bx_{s,t}^{A,C}$ be the process such that $\bx_{s,t}^{A,C}=\bx_t^A=\bx_{\floor*{t}}^A$ for $t\le s$ and for $t\ge s$, $\bx_{s,t}^{A,C}$ follows:
$$
  d\bx^{A,C}_{s,t} = -\eta \nabla_x \bar{f} (\bx^{A,C}_{s,t}) dt +
  \sqrt{\frac{2\eta}{\beta}} d\bw_t - \bv_{s,t}^{A,C} d\bmu_s^{A,C}(t).
  $$
  In other words, $\bx_{s,t}^{A,C}$ follows the algorithm for $t\le s$, and then switches to the dynamics of the continuous-time approximation from (\ref{eq:AvecontinuousProjectedLangevin}) at $t= s$.


  Now let $0\le s\le \hat s \le t$ where $s, \hat{s} \in \bbN$, then Corollary~\ref{cor:contractionW1} from Appendix~\ref{sec:contraction} shows that
  \begin{equation}
  \label{eq:switchContraction}
    W_{1}(\cL(\bx_{s,t}^{A,C}),\cL(\bx_{\hat s,t}^{A,C}))\le  2 \varphi(R)^{-1} e^{-\tilde a (t-\hat s)} W_{1}(\cL(\bx_{s,\hat s}^{A,C}),\cL(\bx_{\hat s,\hat s}^{A,C})).
  \end{equation}

%

  By starting the analysis of the processes $\bx^A$ and $\bx^C$ at time $s$, rather than time $0$, Lemma~\ref{lem:AtoCdependent} implies the following bound:
  \begin{align}
  W_{1}(\cL(\bx_{s,\hat s}^{A,C}),\cL(\bx_{\hat s,\hat s}^{A,C})) &= W_1(\cL(\bx_{s,\hat s}^{A,C}),\cL(\bx_{\hat{s}}^{A})) \nonumber\\
    &\le \left(
      \left(c_{\ref{AtoC1}}+c_{\ref{BoundCtoD2}}\sqrt{\bbE[\|\bx_s^{A}\|^2]}\right) \eta \sqrt{\hat{s}-s}+
    c_{\ref{BoundCtoD3}} \sqrt{\eta \log(4(\hat{s}-s))}
    \right)e^{\eta \ell (\hat{s}-s)} \nonumber\\
        \label{eq:basicSwitch}
    &\le
    \left(
      \left(c_{\ref{AtoC1}}+c_{\ref{BoundCtoD2}}\sqrt{\varsigma + c_{\ref{AlgBound}}}\right) \eta \sqrt{\hat{s}-s}+
    c_{\ref{BoundCtoD3}} \sqrt{\eta \log(4(\hat{s}-s))}
    \right)e^{\eta \ell (\hat{s}-s)}.
  \end{align}



The second inequality is based on Lemma~\ref{lem:algBound}.

Let $H=\floor*{1/\eta}$ and $t \in [\hat{k}H, (\hat{k}+1)H)$ where $\hat{k} \in \bbN$, we have $\bx_{0,t}^{A,C} = \bx_{t}^{C}$ and $\bx_{(\hat{k}+1)H,t}^{A,C} = \bx_{t}^{A} = \bx_{\floor*{t}}^{A}$.  Then, the triangle inequality implies that
   \begin{align*}
  W_1(\cL(\bx_{t}^A),\cL(\bx_{t}^C)) \le 
  \sum_{i=0}^{\hat{k}} W_{1}(\cL(\bx_{iH,t}^{A,C}),\cL(\bx_{(i+1)H,t}^{A,C})). 
  \end{align*}

For $i<\hat{k}$, setting $s=iH$, $\hat s = (i+1)H$ in \eqref{eq:switchContraction} gives that
  \begin{align*}
    W_{1}(\cL(\bx_{iH,t}^{A,C}),\cL(\bx_{(i+1)H,t}^{A,C})) &\le 
    2\varphi(R)^{-1} e^{-\tilde a \left(t-(i+1)H\right)} W_{1}(\cL(\bx_{iH,(i+1)H}^{A,C}),\cL(\bx_{(i+1)H,(i+1)H}^{A,C}))\\
    &\le  2\varphi(R)^{-1} e^{-\eta a \left(t-(i+1)H\right)}g(H) \\
    &\le 	2\varphi(R)^{-1} e^{- a \left(\hat k -i-1)/2\right)} g(\eta^{-1})
  \end{align*}
  where 
  \begin{align}
   g(r) = \left(
      \left(c_{\ref{AtoC1}}+c_{\ref{BoundCtoD2}}\sqrt{\varsigma + c_{\ref{AlgBound}}}\right) \eta \sqrt{r}+
    c_{\ref{BoundCtoD3}} \sqrt{\eta \log(4r)}
    \right)e^{\eta \ell r}.
  \end{align}
  The last inequality uses the facts that $1/2 \le \eta H \le 1 $ along with monotonicity of $g$. The lower bound of $\eta H$ arises because $H  \ge \eta^{-1} -1 $ and so $\eta H \ge 1- \eta \ge 1/2$, since $\eta \le 1/2$.
  Thus, the first $\hat k$ terms are bounded by:
  \begin{align*}
  \nonumber
  \sum_{i=0}^{\hat k-1} W_{1}(\cL(\bx_{iH,t}^{A,C}),\cL(\bx_{(i+1)H,t}^{A,C})) &\le \sum_{i=0}^{\hat k-1} 2\varphi(R)^{-1} e^{- a \left(\hat k-i-1)/2\right)} g(\eta^{-1}) \\
  &\le  2\varphi(R)^{-1} \frac{g(\eta^{-1})}{1-e^{-{a}/2}}
  \end{align*}
  For $ i =\hat{k}$, 
  \begin{align*}
  W_{1}(\cL(\bx_{iH,t}^{A,C}),\cL(\bx_{(i+1)H,t}^{A,C})) &= W_1(\cL(\bx_{\hat k H,t}^{A,C}),\cL(\bx_{t}^{A}))\\
  &\le g(t-\hat k H) \le g(\eta^{-1})
  \end{align*}
  By triangle inequality, adding all the $\hat k+1$ terms gives 
  \begin{align}
\nonumber
\MoveEqLeft[0]
  W_1(\cL(\bx_t^A),\cL(\bx_t^C))\\
  \nonumber
  &\le g(\eta^{-1}) \left( 1+  \frac{2\varphi(R)^{-1}}{1-e^{-{a}/2}}\right)\\
  &\le \left(
      \left(c_{\ref{AtoC1}}+c_{\ref{BoundCtoD2}}\sqrt{\varsigma + c_{\ref{AlgBound}}}\right) \eta \sqrt{\eta^{-1}}+
    c_{\ref{BoundCtoD3}} \sqrt{\eta \log(4\eta^{-1})}
    \right)e^{\ell} \left( 1+  \frac{2\varphi(R)^{-1}}{1-e^{-{a}/2}}\right)
    .
  \label{eq:AtoCGen}  
  \end{align}

  For $\eta^{-1} \ge 4$, we have $\log(4\eta^{-1}) \le 2 \log(\eta^{-1})$, and also $\log \eta^{-1}>1$. Thus, if $\eta \le 1/4$, then \eqref{eq:AtoCGen} can be further upper bounded by 
   \begin{align*}
    \nonumber
    \MoveEqLeft[0]
  W_1(\cL(\bx_t^A),\cL(\bx_t^C))\\
    &\le \left(
      \left(c_{\ref{AtoC1}}+c_{\ref{BoundCtoD2}}\sqrt{\varsigma + c_{\ref{AlgBound}}}\right) \eta \sqrt{\eta^{-1} \log(\eta^{-1})}+
    c_{\ref{BoundCtoD3}} \sqrt{2\eta \log( \eta^{-1})}
    \right)e^{\ell} \left( 1+  \frac{2\varphi(R)^{-1}}{1-e^{-{a}/2}}\right)\\
    &=\left(c_{\ref{AtoC1}}+c_{\ref{BoundCtoD2}}\sqrt{\varsigma + c_{\ref{AlgBound}}}+ \sqrt{2}c_{\ref{BoundCtoD3}} \right) e^{\ell}\left( 1+  \frac{2\varphi(R)^{-1}}{1-e^{-{a}/2}}\right)\sqrt{\eta \log( \eta^{-1})} \\
    &\le \left(c_{\ref{AtoC1}}+c_{\ref{BoundCtoD2}}\sqrt{ c_{\ref{AlgBound}}}+ \sqrt{2}c_{\ref{BoundCtoD3}}  + c_{\ref{BoundCtoD2}} \sqrt{\varsigma}\right) e^{\ell}\left( 1+  \frac{2\varphi(R)^{-1}}{1-e^{-{a}/2}}\right)\sqrt{\eta \log( \eta^{-1})}.
  \end{align*}
   So setting
 \begin{align*}	
c_{\ref{error_polyhedron1}} &= \left(c_{\ref{AtoC1}}+c_{\ref{BoundCtoD2}}\sqrt{c_{\ref{AlgBound}}}+ \sqrt{2} c_{\ref{BoundCtoD3}} \right) e^{\ell}\left( 1+  \frac{2\varphi(R)^{-1}}{1-e^{-{a}/2}}\right) 
\\
c_{\ref{error_polyhedron2}}  &= c_{\ref{BoundCtoD2}} e^{\ell}\left( 1+  \frac{2\varphi(R)^{-1}}{1-e^{-{a}/2}}\right)
 \end{align*}
 completes the proof.
\hfill$\blacksquare$


\section{Bounding the constants}
\label{app:constants}
In this section, we summarize all the constants in Table \ref{tb:constants}. The second column of the table points to the place where these values are defined or computed. Then we show the simplified bounds of the main constants $c_{\ref{contraction_const1}}, c_{\ref{contraction_const2}}, c_{\ref{error_polyhedron1}}, c_{\ref{error_polyhedron2}},a  $ in Theorem~{\ref{thm:nonconvexLangevin}} explicitly and also discuss their dependencies on state dimension $n$ and parameter $\beta$.

\begin{table}[ht]
  \caption{List of constants}
  \label{tb:constants}
  \centering
  \begin{tabular}{l l}
    \toprule
	Constant & Definition \\
		\midrule\\
		$a =  \frac{2\xi}{\beta}$ \\
$c_{\ref{contraction_const1}} = \contractionConstOneVal $ \\
$c_{\ref{contraction_const2}} =  \contractionConstTwoVal $
& \multirow{1}*{Appendix \ref{ss:proofLemmaconvergeToStationary} (Proof of Lemma~\ref{lem:convergeToStationary}}) \\
 \midrule\\
$c_{\ref{error_polyhedron1}} = \errorPolyhedronOneVal$
 \\
$c_{\ref{error_polyhedron2}}  = \errorPolyhedronTwoVal $ & \multirow{1}*{Appendix \ref{app:switching} (Proof of Lemma~\ref{lem:AtoC})} \\
  \midrule\\
$
c_{\ref{BoundCtoD1}} = \BoundCtoDOneVal $
\\
 $c_{\ref{BoundCtoD2}} = \BoundCtoDTwoVal $\\
  $c_{\ref{BoundCtoD3}} =  \BoundCtoDThreeVal $
 & \multirow{2}*{Appendix \ref{app:disBounds} (Proof of Lemma~\ref{lem:C2D}) }  \\ 
   \midrule\\
$
c_{\ref{AtoC1}} = \AtoCOneVal $
& Section \ref{ss:proofOverviewofLemmaAtoC} (Proof of Lemma \ref{lem:AtoCdependent}) \\ %
  \midrule\\
$
c_{\ref{diamBound}} = \diamBoundVal $
& Appendix \ref{app:diamProof} (Proof of Lemma~\ref{lem:boundingExistence})\\
  \midrule\\
$
  c_{\ref{LyapunovConst}}
    = \LyapunovConstVal $ 
& Appendix \ref{app:contBounds} (Proof of Lemma \ref{lem:GeometricDrift})\\ 
	  \midrule\\
$
c_{\ref{AlgBound}} = \AlgBoundVal  
$ 
& Appendix \ref{app:dis-timeBounds} (Proof of Lemma \ref{lem:algBound}) \\ %
    \bottomrule
  \end{tabular}
\end{table}

\begin{proposition}
\label{prop:mainConstantsBound}
The constants $c_{\ref{contraction_const2}}$ and $c_{\ref{error_polyhedron2}}$ grow linearly with $n$. The constants $c_{\ref{contraction_const1}}$ and $c_{\ref{error_polyhedron1}}$ have $O(\sqrt{n})$ and $O(n)$ dependencies respectively. So overall, the dimension dependency of convergence guarantee is $O(n)$. Constants $c_{\ref{contraction_const1}}, c_{\ref{contraction_const2}},c_{\ref{error_polyhedron1}}, c_{\ref{error_polyhedron2}}$ all grow exponentially with respect to $\frac{\beta \ell R^2}{2}$. And for all $\beta>0$, $a \ge \frac{2}{\frac{\beta R^2}{2}+\frac{16}{\mu}}  e^{-\frac{\beta \ell R^2}{4}}$.
\end{proposition}

\paragraph{Proof of Proposition~\ref{prop:mainConstantsBound}}

Recall that $a = 2\xi / \beta$, and from (\ref{eq:xi}) we have that 
from
$$
\xi^{-1} = \int_0^{R_1} \Phi(s)\varphi(s)^{-1} ds. 
$$
So, to get a lower bound on $\xi$, we need an upper bound on the right side. Recalling the definitions of the various functions for our scenario gives:
\begin{align*}
  h(s) &=  \frac{\ell\beta \min\{s^2,R^2\}}{8} \\
  \varphi(s) &= e^{-h(s)} \\
  \Phi(s)&= \int_0^s \varphi(r)dr.
\end{align*}
It follows that $\Phi(s)\le s$ and $\varphi(s)^{-1}=e^{h(s)}\le e^{\frac{\ell \beta R^2}{8}}$. Thus, we have that
$$
\xi^{-1}\le \frac{1}{2}R_1^2 e^{\frac{\ell\beta R^2}{8}}. 
$$

Now, note that in Corollary~\ref{cor:contraction} that we have set
$$
R_1 = \frac{R}{2} +\frac{1}{2} \sqrt{R^2 + \frac{32}{\mu
      \beta}e^{\frac{\beta \ell R^2}{8}}}.
  $$
  So, a bit of crude upper bounding gives:
  \begin{align*}
    \xi^{-1}&\le \frac{1}{2}R_1^2 e^{\frac{\ell\beta R^2}{8}}    \\
            &\le \frac{1}{2}\left( R^2 + \frac{32}{\mu
      \beta}e^{\frac{\beta \ell R^2}{8}}\right) 
              e^{\frac{\beta \ell R^2}{8}} \\
    &\le \left(\frac{R^2}{2}+\frac{16}{\mu\beta} \right) e^{\frac{\beta \ell R^2}{4}}
  \end{align*}

  The final bound on $a$ becomes:
  \begin{align*}
    a &= 2\xi/\beta \ge \frac{2}{\frac{\beta R^2}{2}+\frac{16}{\mu}}  e^{-\frac{\beta \ell R^2}{4}}
  \end{align*}

  The rest of focuses on bounding the other constants as $\beta$ grows large. For all sufficiently large $\beta$, we have that
  $$
  \frac{\frac{\beta R^2}{2}+\frac{16}{\mu}}{2}\le e^{\frac{\beta \ell R^2}{4}}
  $$
  so that
  \begin{equation}
    \label{eq:crudeA}
    a\ge e^{-\frac{\beta \ell R^2}{2}}
    .
  \end{equation}

We have the following inequality for all sufficiently large $\beta$:
\begin{align*}
\frac{1}{1-e^{-a/2}} &\le  \max\left\{ \frac{4}{a}, \frac{1}{1-e^{-1}}\right\}\\
                     &\le  \max\left\{ 4 e^{\frac{\beta \ell R^2}{2} }, \frac{1}{1-e^{-1}}\right\} \\
                     &= 4 e^{\frac{\beta \ell R^2}{2}}.
\end{align*}
The first inequality uses the fact that for all $y>0$, $\frac{1}{1-e^{-y}} \le \max\left\{ \frac{2}{y}, \frac{1}{1-e^{-1}}\right\}$, which is shown in \citep{lamperski2021projected}.

So 
\begin{align}
\label{eq:mainBoundExponential}
  1+ \frac{2 \varphi(R)^{-1}}{1-e^{-a/2}} \le 1+
 4 e^{\frac{\beta \ell R^2}{2}}.
\end{align}

Now we bound the growth of the other constants for large $\beta$. So,
without loss of generality, assume $\beta \ge 1$.
Then, plugging the definition of $\xi$ and $\varphi$ and (\ref{eq:mainBoundExponential}) gives  
\begin{align*}
c_{\ref{contraction_const1}} &= 2 e^{\frac{\beta \ell R^2}{8}}\sqrt{\frac{2}{\mu}\left( (\ell +\mu)R^2 + R \|\nabla_x \bar{f}(0)\| + \frac{n}{\beta}\right)} \\
&\le 2 e^{\frac{\beta \ell R^2}{8}}\sqrt{\frac{2}{\mu}\left( (\ell +\mu)R^2 + R \|\nabla_x \bar{f}(0)\| + n\right)}\\
c_{\ref{contraction_const2}} &= 4 e^{\frac{\beta \ell R^2}{8}}\\
c_{\ref{error_polyhedron1}} &= \left(c_{\ref{AtoC1}}+c_{\ref{BoundCtoD2}}\sqrt{c_{\ref{AlgBound}}}+ \sqrt{2} c_{\ref{BoundCtoD3}} \right) e^{\ell}\left( 1+  \frac{2 \varphi(R)^{-1}}{1-e^{-a/2}}\right) \\
& \le \left(c_{\ref{AtoC1}}+c_{\ref{BoundCtoD2}}\sqrt{c_{\ref{AlgBound}}}+ \sqrt{2} c_{\ref{BoundCtoD3}} \right) e^{\ell}\left(1+ 4e^{\frac{\beta\ell R^2}{2}}\right) \\
&\le r(\sqrt{n})  e^{\ell}\left(1+4e^{\frac{\beta \ell R^2}{2}}\right)\\
c_{\ref{error_polyhedron2}} &= \left( 6(\frac{1}{\alpha})^{\textrm{rank}(A)/2}+1\right) \sqrt{2\ell^2}e^{\ell}\left( 1+  \frac{2 \varphi(R)^{-1}}{1-e^{-a/2}}\right)\\
& \le \left( 6(\frac{1}{\alpha})^{\textrm{rank}(A)/2}+1\right) \sqrt{2\ell^2}e^{\ell} \left(1+ 4e^{\frac{\beta \ell R^2}{2}}1\right) .
\end{align*}

For constant $c_{\ref{error_polyhedron1}}$, $r(\sqrt{n})$ is a monotonically increasing function of order $\sqrt{n}$, (independent of $\eta$ and $\beta$). The upper bound of $c_{\ref{error_polyhedron1}}$ is derived by direct observation of the corresponding constants.
 
We can see neither $c_{\ref{contraction_const2}}$ nor $c_{\ref{error_polyhedron2}}$ depends on the state dimension, so the two constants grow linearly with $n$. The constant $c_{\ref{contraction_const1}}$ are $O(\sqrt{n})$ and $c_{\ref{error_polyhedron1}}$ are $O(n)$. As for the dependencies on $\beta$, we can see that all four constants are $O(e^{\frac{\beta \ell R^2}{2}})$.

\hfill $\blacksquare$


\section{Near-optimality of Gibbs distributions}

In this appendix, we prove Proposition~\ref{prop: app_optimization} which shows that $\bx_k$ can be near-optimal. The proof closely follows \cite{lamperski2021projected} and \cite{raginsky2017non}. The main difference is that in our case we have to deal with the unbounded polyhedral constraint, while in \cite{raginsky2017non} there is no constraint and in \cite{lamperski2021projected} the constraint is compact.

Firstly, we need a preliminary result shown as below.
\const{const_subopt}
\begin{lemma}
  \label{lem:gibbsSuboptimality}
Assume $\bx$ is drawn according to $\pi_{\beta \bar f}$. There exists a positive constant $c_{\ref{const_subopt}}$ such that the following bounds hold:
\begin{align*}
\bbE[\bar f(\bx)] \le \min_{x \in \cK} \bar f(x) +\frac{n}{ \beta} \left( 2\max\{0,\log\varsigma\} +c_{\ref{const_subopt}}\right)
\end{align*}
where $c_{\ref{const_subopt}} =\log n +  2 \log( 1 +\frac{1}{\mu} c_{\ref{LyapunovConst}} ) +  \frac{1}{6}\log 3 + \log 2 \sqrt{\pi} -   \log r_{\min}$ and $r_{\min}$ is a positive constant. 
\end{lemma}

\paragraph{Proof of Lemma~{\ref{lem:gibbsSuboptimality}}}
Recall that the probability measure $\pi_{\beta \bar f }(A)$ is defined by $\pi_{\beta \bar f (A)}(A) = \frac{\int_{A \cap \cK} e^{-\beta \bar f(x)}dx}{\int_{\cK} e^{- \beta \bar f(y)}dy}$.

Let $\Lambda = \int_{\cK} e^{-\beta \bar f(y)} dy $ and $p(x) = \frac{e^{-\beta \bar f(x)}}{\Lambda}$. So  $\log p(x) =  -\beta \bar f(x) - \log \Lambda$, which implies that $\bar f(x) = -\frac{1}{\beta} \log p(x) -\frac{1}{\beta} \log \Lambda$. Then we have
\begin{align}
\nonumber
\bbE_{\pi_{\beta \bar f}} [\bar f(\bx)] &=  \int_{\cK}  \bar f(x) p(x) dx \\
\label{eq:sub_optimal}
 &=  -\frac{1}{\beta} \int_{\cK} p(x) \log p(x) dx -\frac{1}{\beta} \log \Lambda.
\end{align}

We can bound the first term by maximizing the differential entropy. 

Let $h(x) = - \int_{\cK} p(x) \log p(x) dx$. Using the fact that the differential entropy of a distribution with finite moments is upper-bounded by that of a Gaussian density with the same second moment (see Theorem 8.6.5 in \cite{cover2012elements}), we have 
\begin{align}
\label{eq:diff_entropy_bound}
h(x) \le \frac{n}{2}\log(2 \pi e \sigma^2) \le \frac{n}{2}\log(2 \pi e ( \varsigma +\frac{1}{\mu} c_{\ref{LyapunovConst}})),
\end{align}
where $\sigma^2 = \bbE_{\pi_{\beta\bar f}}[\|\bx\|^2]$ and 
the second inequality uses Lemma \ref{lem:continuousBound}.

We aim to derive the upper bound of the second term of (\ref{eq:sub_optimal}).

First we show that there is a vector $x^\star \in \cK$ which minimizes $\bar f$ over $\cK$. In other words, an optimal solution exists. The bound (\ref{eq:quadraticLower}) from the proof of Lemma~\ref{lem:gibbs} implies that $\bar f(x)\ge  \bar f(0)+1$ for all sufficiently large $x$. This implies that there is a compact ball, $B$ such that if $x_n\in \cK$ is a sequence such that $\lim_{n\to \infty}\bar f(x_n)=\inf_{x\in\cK} \bar f(x)$, then $x_n$ must be in $B\cap \cK$ for all sufficiently large $n$. Then since $\bar f$ is continuous and $B\cap \cK$ is compact, there must be a limit point $x^\star \in B\cap \cK$ which minimizes $\bar f$.

Let $x^* \in \cK$ be a minimizer. The normalizing constant can be expressed as:
\begin{align*}
\label{eq:logNormalizationConstant}
\log \Lambda &= \log \int_{\cK} e^{-\beta \bar f(x)} dx\\
	&= \log e^{-\beta \bar f(x^*)} \int_{\cK} e^{\beta \left( \bar f(x^*) - \bar f(x)\right) } dx\\
	&= -\beta \bar f(x^*) + \log  \int_{\cK} e^{\beta \left( \bar f(x^*) - \bar f(x)\right) } dx
\end{align*}

So, to derive our desired upper bound on $-\log\Lambda$, 
it suffices to derive a lower bound on
\begin{equation}
  \label{eq:suboptIntegral}
\int_{\cK} e^{\beta \left( \bar f(x^*) - \bar f(x)\right) } dx.
\end{equation}

We have 
\begin{align*}
\bar f(x) - \bar f(x^*) =  \int_0^1 \nabla \bar f(x^* + t(x-x^*))^\top (x-x^*)dt.
\end{align*} 
Let $y = x^* + t(x-x^*)$, $t \in [0,1]$, then
\begin{align*}
\|\nabla \bar f(y)\| &= \| \nabla \bar f(y) - \nabla \bar f(x^*) +  \nabla \bar f(x^*) - \nabla \bar f(0) + \nabla \bar f(0)\| \\
 &\le \ell \| y- x^*\| + \ell \| x^*\|  + \| \nabla \bar f(0)\| \\
 & \le \ell \|x-x^*\| t + \ell \| x^*\|  + \| \nabla \bar f(0)\|.
\end{align*}

We can show $\|x^*\|$ is upper bounded by $\max \{R, \frac{\| \nabla \bar f(0)\|}{\mu}\}$. 

We have to find the bound for the case $\|x^*\| > R$.

The convexity outside a ball assumption gives
\begin{equation}
\label{eq:conv_opt}
\left( \nabla \bar f(x^*) - \nabla \bar f(0) \right)^\top x^* \ge \mu \|x^*\|^2.
\end{equation}

The optimality of $x^*$ gives $-\nabla \bar f(x^*) \in N_{\cK}(x^*)$, which is to say for all $y \in \cK$, $-\nabla \bar f(x^*)^\top (y- x^*) \le 0$. Since $0 \in \cK$, $\nabla \bar f(x^*)^\top x^* \le 0 $ holds. Applying the Cauchy-Schwartz inequality to the left side of \eqref{eq:conv_opt} gives 
\begin{align*}
 \| \nabla \bar f(0) \|  \|x^*\| \ge \mu \|x^*\|^2.
\end{align*}

\const{xoptBound}
This implies that $\|x^*\| \le  \frac{\|\nabla \bar f(0)\|}{\mu}  $. So we can conclude that $\|x^*\| \le \max \{R,  \frac{\|\nabla \bar f(0)\|}{\mu} \} = c_{\ref{xoptBound}}$.

Therefore,
\begin{align*}
\bar f(x) - \bar f(x^*) &\le \int_0^1 \|\nabla \bar f(x^*+t(x-x^*))\| \|x -x^*\| dt \\
&\le \frac{\ell}{2} \|x- x^*\|^2 + \left( \ell \| x^*\|  + \| \nabla \bar f(0)\|\right) \|x-x^*\| \\
&\le \frac{\ell}{2} \|x- x^*\|^2 + \left( \ell c_{\ref{xoptBound}} + \| \nabla \bar f(0)\|\right) \|x-x^*\|.
\end{align*}

To lower-bound the integral from (\ref{eq:suboptIntegral}), we restrict our attention to the points $x$ such that the integrand is at least $1/2$. 
For these values, we have the following implications:
\begin{align*}
&e^{\beta \left( \bar f(x^*) - \bar f(x)\right)}  \ge 1/2 \\
\iff &\beta \left( \bar f(x^*) - \bar f(x)\right) \ge -\log 2 \\
\impliedby & -\frac{\ell}{2} \|x- x^*\|^2 - \left( \ell c_{\ref{xoptBound}}  + \| \nabla \bar f(0)\|\right) \|x-x^*\| \ge -\frac{1}{\beta} \log 2.
\end{align*}

So solving the corresponding quadratic equation and taking the positive root gives an upper bound of $\|x- x^*\|$:
\begin{align*}
\|x - x^*\| \le - \frac{1}{\ell}\left( \ell c_{\ref{xoptBound}}  + \| \nabla \bar f(0)\|\right) + \frac{1}{\ell}\sqrt{\left( \ell c_{\ref{xoptBound}}  + \| \nabla \bar f(0)\|\right)^2 + 2 \ell \frac{1}{\beta} \log 2}.
\end{align*}

So let $\epsilon =  -\frac{1}{\ell}\left( \ell c_{\ref{xoptBound}}  + \| \nabla \bar f(0)\|\right) + \frac{1}{\ell}\sqrt{\left( \ell c_{\ref{xoptBound}}  + \| \nabla \bar f(0)\|\right)^2 + 2 \ell \frac{1}{\beta} \log 2}$ and let $\cB_{x^*}(\epsilon)$ be the ball of radius $\epsilon$ centered at $x^*$. Then we want to find a ball $\cS$ such that 
\begin{align*}
\int_{\cK} e^{\beta \left( \bar f(x^*) - \bar f(x)\right)} dx \ge \frac{1}{2} \textrm{vol} (\cK \cap \cB_{x^*}(\epsilon)) \ge \frac{1}{2} \textrm{vol}(\cS).
\end{align*}

To find the desired ball $\cS$, we consider the problem of finding the largest ball inscribed within $\cK\cap \cB_{x^\star}(\epsilon)$. This is a Chebyshev centering problem, and can be formulated 
as the following convex optimization problem.

  \begin{subequations}
    \label{eq:chebyOptimization}
  \begin{align}
    &\max_{r,y} && r \\
    \label{eq:ballInPolyhedron}
    &\textrm{subject to} && Ay \le b - r \boldsymbol{1} \\
    \label{eq:ballInBall}
    &&&\|x^* -y\| + r  \le \epsilon
  \end{align}
\end{subequations}
where $r$ and $y$ denotes the radius and the center of the Chebyshev ball respectively. The particular form arises because the rows of $A$ are unit vectors, and so the ball of radius $r$ around $y$ is inscribed in $\cK$ if and only if (\ref{eq:ballInPolyhedron}) holds, while this ball is contained in $\cB_{x^\star}(\epsilon)$ if and only if (\ref{eq:ballInBall}) holds. 
 

We rewrite this optimization problem as:
  \begin{subequations}
    \label{eq:chebyOptimization2}
  \begin{align}
    &\min_{r,y }&& -r + I_S\left(x^*, [\begin{smallmatrix}r\\y \end{smallmatrix}]\right)\\
    &\textrm{subject to} && Ay \le b - r \boldsymbol{1}
  \end{align}
\end{subequations}
where $S = \{(x^*, [\begin{smallmatrix}r\\y \end{smallmatrix}]) \vert \|x^* -y\| + r <\epsilon \}$.

Here, $I_S$ is defined by
\begin{equation}
I_S(x,[\begin{smallmatrix}r\\y \end{smallmatrix}]) =  \left\{ 
  \begin{array}{ c l }
    +\infty & \quad \textrm{if } (x,[\begin{smallmatrix}r\\y \end{smallmatrix}]) \notin S \\
    0                 & \quad \textrm{otherwise}.
  \end{array}
\right.
\end{equation}

Let $g(x^\star)$ denote the optimal value of (\ref{eq:chebyOptimization2}). We will show that there is a positive constant $r_{\min}>0$ such that $-g(x) \ge r_{\min}$ for all $x \in \cK$. As a result, for any $x^\star$ the corresponding Chebyshev centering solutions has radius at least $r_{\min}$. 


Let $F(x^*,[\begin{smallmatrix}r\\y \end{smallmatrix}])=  -r + I_s(x^*, [\begin{smallmatrix}r\\y \end{smallmatrix}])$.  We can see that $F$ is convex in $(x^*, [\begin{smallmatrix}r\\y \end{smallmatrix}])$  and $\textrm{dom}\:F = S$.

Let $C = \{ [\begin{smallmatrix}r\\y \end{smallmatrix}] \vert [\boldsymbol{1} \; A] [\begin{smallmatrix}r\\y \end{smallmatrix}]  \le b \}$. Then the optimal value of (\ref{eq:chebyOptimization2}) can be expressed as $g(x)= \inf_{[\begin{smallmatrix}r\\y \end{smallmatrix}] \in C} F(x, [\begin{smallmatrix}r\\y \end{smallmatrix}])$ and $\textrm{dom}\:g = \{ x \vert \exists [\begin{smallmatrix}r\\y \end{smallmatrix}] \in C \; \textrm{s.t.} \; (x, [\begin{smallmatrix}r\\y \end{smallmatrix}]) \in S\}$. 

The results of Section 3.2.5 of \cite{boyd2004convex} imply that if $F$ is convex, $S$ is convex, and $g(x) >-\infty$ for all $x$, then $g$ is also convex.


If $(x, [\begin{smallmatrix}r\\y \end{smallmatrix}])\in \textrm{dom} \:F$, then 
\begin{align*}
\|x-y\| + r \le \epsilon & \implies r \le \epsilon - \|x-y\|\\
& \implies -r \ge - \epsilon + \|x-y\| > - \infty.
\end{align*}
In particular, if there exist $ y,r $ such that $(x, [\begin{smallmatrix}r\\y \end{smallmatrix}]) \in \textrm{dom} \:F$, then $\inf_{[\begin{smallmatrix}r\\y \end{smallmatrix}] \in C} F(x, [\begin{smallmatrix}r\\y \end{smallmatrix}]) \ge -\epsilon $.

There are two cases:
\begin{itemize}
\item If there exists $[\begin{smallmatrix}r\\y \end{smallmatrix}] \in C$ such that $(x, [\begin{smallmatrix}r\\y \end{smallmatrix}]) \in \textrm{dom}\: F$, then $\inf_{[\begin{smallmatrix}r\\y \end{smallmatrix}] \in C} F(x, [\begin{smallmatrix}r\\y \end{smallmatrix}])  $ is finite and bounded below.
\item If there does not exist $[\begin{smallmatrix}r\\y \end{smallmatrix}] \in C$ such that $(x, [\begin{smallmatrix}r\\y \end{smallmatrix}]) \in \textrm{dom} \:F$, then for all $[\begin{smallmatrix}r\\y \end{smallmatrix}] \in C$, $F(x, [\begin{smallmatrix}r\\y \end{smallmatrix}]) = +\infty$. So $g(x) = \inf_{[\begin{smallmatrix}r\\y \end{smallmatrix}] \in C} F(x,[\begin{smallmatrix}r\\y \end{smallmatrix}]) = +\infty > - \infty$.
\end{itemize}

Hereby, we can conclude that for all $x$ , $g(x) >-\infty$, so $g(x)$ is convex. 

%
%

So, to found a lower bound on the inscribed radius, we want to maximize $g(x)$ over $\cK$.
Specifically, we analyze the following optimization problem
\begin{subequations}
    \label{eq:chebyOptimization3}
  \begin{align}
    &\max_{x \in \cK }&& g(x) 
  \end{align}
\end{subequations}
which corresponds to maximizing a convex function over a convex set. 

Note that $\cK \subset \dom{(g )}$. In particular, if $x\in\cK$, then $(x,[\begin{smallmatrix} 0\\x \end{smallmatrix}])\in S$, which implies that  $g(x) \le 0$. Thus, $g(x)\le 0$  for all $x\in \cK$. 
Therefore, using Theorem 32.2 \cite{rockafellar2015convex}, given $\cK$ is closed convex by our assumption and $g(x)$ is bounded above gives
\begin{align*}
\sup \left\{ g(x) \vert x \in \cK \right\} = \sup \left\{ g(x) \vert x \in E \right\}
\end{align*} 
where $E$ is a subset of $\cK$ consisting of the extreme points of $\cK \cap L^{\perp}$, where $L$ is the linearity space of $C$ and $L = \{x \vert Ax = 0  \} = \cN(A)$.

Now, we will show that $E$ is a finite set.

Let 
\begin{align*}
A = \begin{bmatrix}
U_1 & U_2
\end{bmatrix}
\begin{bmatrix}
\Sigma & 0 \\
0& 0 
\end{bmatrix}
\begin{bmatrix}
V_1^\top \\
 V_2^\top
\end{bmatrix}.
\end{align*}
Then $\cN(A) =  L = \cR(V_2)$
and $L^\perp = \cR(V_1)$, and 
\begin{align*}
K\cap L^\perp = \{ V_1 Z_1 \vert A V_1 Z_1 \le b \}. 
\end{align*}

This is a polyhedral with no lines so has a finite set of extreme points, i.e. $E$ is finite. In particular, they are contained in a compact subset of $\cK$. Then it is shown in the proof of Proposition 16 of \cite{lamperski2021projected} that the Chebyshev centering problem has a positive global lower bound, when restricted to a compact convex set with $0$ in its interior. Denote this value by $r_{\min}$.

%
Thus, we have that $\textrm{vol}(\cS)\ge \frac{\pi^{n/2}}{\Gamma(n/2+1)}r_{\min}^n $, using the fact that a ball of radius $\rho$ has volumn given by $\frac{\pi^{n/2}}{\Gamma(n/2+1)}\rho^n$

Then, utilizing an upper bound of Gamma function recorded in \cite{ramanujan1988lost} shown as below:  

\begin{equation}
\label{eq:stirling}
\Gamma(x+1) < \sqrt{\pi} \left( \frac{x}{e}\right)^x \left( 8x^3 + 4x^2 +x + \frac{1}{30}\right)^{1/6}, \; x \ge 0.
\end{equation}

Setting $x = \frac{n}{2}$ in (\ref{eq:stirling}) gives:
\begin{equation}
\Gamma(\frac{n}{2}+1) < \sqrt{\pi} \left( \frac{n}{2e}\right)^{\frac{n}{2}} \left( n^3 + n^2 + \frac{n}{2} + \frac{1}{30}\right)^{1/6}.
\end{equation}

Therefore, we can find the lower bound of $\log \frac{1}{2} \textrm{vol}(S)$:
\begin{align}
\nonumber
\log \frac{1}{2} \textrm{vol}(S) &= \log \frac{\pi^{n/2}}{\Gamma(n/2+1)}r_{\min}^n - \log 2\\
\nonumber
&> \frac{n}{2} \log \pi + n \log r_{\min} - \log \left\{ \sqrt{\pi} \left( \frac{n}{2e}\right)^{\frac{n}{2}} \left( n^3 + n^2 + \frac{n}{2} + \frac{1}{30}\right)^{1/6} \right\} - \log 2\\\nonumber
\label{eq:logvolumn_bound}
&= -\frac{1}{2}  \log \pi + n \log r_{\min} + \frac{n}{2} \log (2 \pi e)  - \frac{n}{2} \log n- \frac{1}{6} \log \left( n^3 + n^2 + \frac{n}{2} + \frac{1}{30}\right) - \log 2\\
& > n \log r_{\min} + \frac{n}{2} \log (2 \pi e)  - \frac{n}{2} \log n -\frac{1}{6} \log \left(  3 n^3 \right) - \log (2 \sqrt{\pi})
\end{align}

The last inequality holds because  $n \ge 1 $.

Plugging (\ref{eq:logvolumn_bound}) and (\ref{eq:diff_entropy_bound}) in (\ref{eq:sub_optimal}) gives
\begin{align*}
\bbE_{\pi_{\beta \bar f}} [\bar f(x)] & <  \min f(x) + \frac{n}{2 \beta} \log(2 \pi e ( \varsigma +\frac{1}{\mu} c_{\ref{LyapunovConst}})) \\
& -\frac{1}{\beta} \left(  n \log r_{\min} + \frac{n}{2} \log (2 \pi e)  - \frac{n}{2} \log n  -\frac{1}{2} \log n - \frac{1}{6}\log 3 - \log 2 \sqrt{\pi} \right)\\
&= \min f(x) +\frac{n}{2 \beta} \log( \varsigma +\frac{1}{\mu} c_{\ref{LyapunovConst}}) - \frac{1}{\beta} \left(  n \log r_{\min} - \frac{n}{2} \log n  - \frac{1}{2} \log n \right)\\
&  +
\frac{1}{\beta} ( \frac{1}{6}\log 3 + \log 2 \sqrt{\pi})\\
                                      &\le \min f(x) +\frac{n}{ \beta} \left(
2 \log( \varsigma +\frac{1}{\mu} c_{\ref{LyapunovConst}} ) +  \frac{1}{6}\log 3 + \log 2 \sqrt{\pi} -   \log r_{\min}
                                        + \log n \right).
\end{align*}
where last inequality holds because $n \ge 1$.

The final form of the bound holds because for any $c>0$
\begin{align*}
  \log\left(\varsigma + c \right) &\le \log\left( \max\{\varsigma,1\} + c\right) \\
                                  &= \log \max\{\varsigma,1\} + \log\left( 1 + \frac{c}{\max\{\varsigma,1\}}\right) \\
  &\le \max\{\log\varsigma,0\}+ \log\left( 1 + c\right).
\end{align*}
\hfill $\blacksquare$

Now we cover the case of compact sets for comparison with \cite{lamperski2021projected}.

\const{compactLipschitz}
\const{subOptIndep}
\const{TExponent}

\begin{proposition}
  \label{prop:suboptCompact}
  Assume that $\cK$ has diameter $D$ and $0\in \cK$ and let $c_{\ref{compactLipschitz}}=\ell D + \|\nabla \bar f(0)\|$. Then for all $k\ge 0$, the iterates of the algorithm satisfy
  \begin{equation}
    \label{eq:compactKantorovich}
    \bbE[\bar f(\bx^A_k)]\le \min_{x\in\cK} f(x) + c_{\ref{compactLipschitz}} W(\cL(\bx^A_k),\pi_{\beta \bar f}) + \frac{n}{ \beta} \left(\max\{\log\varsigma,0\}+ c_{\ref{const_subopt}} \right)
  \end{equation}
  In particular, there are constants $c_{\ref{subOptIndep}}$ and $c_{\ref{TExponent}}$ such that, for all sufficiently small $\epsilon$,  if
  \begin{subequations}
    \label{eq:betaOpt}
    \begin{align}
      \beta &= \frac{2n(\max\{\log\varsigma,0\}+c_{\ref{subOptIndep}})}{\epsilon}\\
      T&= e^{c_{\ref{TExponent}/\epsilon}}
    \end{align}
  \end{subequations}
  then 
   \begin{equation}
    \label{eq:compactSuboptimality}
    \bbE[\bar f(\bx^A_T)]\le \min_{x\in\cK} \bar f(x) +\epsilon. 
  \end{equation}
 
\end{proposition}

\paragraph{Proof of Proposition~{\ref{prop:suboptCompact}}}

  First, we show that $\bar f(x)$ is Lipschitz with Lipschitz constant $c_{\ref{compactLipschitz}}$. Indeed,
  $$
  \|\nabla \bar f(x)\| \le \|\nabla \bar f(x)-\nabla \bar f(0)\| + \| \nabla \bar f(0)\| \le \ell D + \|\nabla \bar f(0)\|.
  $$
  So, if $x$ and $y$ are in $\cK$, we have
  \begin{align*}
    |\bar f(x)-\bar f(y)|
    &= \left|
      \int_0^1 \nabla \bar f(y+t(x-y))^\top (x-y)dt 
      \right|\\
    &\le c_1 \|x-y\|.
  \end{align*}
  Then (\ref{eq:compactKantorovich}) follows by Kantorovich duality combined with Lemma~\ref{lem:gibbsSuboptimality}.

  Now, using our bound from Theorem~\ref{thm:nonconvexLangevin} gives that for $T\ge 4$:
  $$
  \bbE[\bar f(\bx^A_T)]\le \min_{x\in\cK} \bar f(x) + \frac{n}{ \beta} \left(\max\{\log\varsigma,0\}+ c_{\ref{const_subopt}}  \right)
  +c_{\ref{compactLipschitz}}
  \left( c_{\ref{contraction_const1}} + c_{\ref{contraction_const2}} \sqrt{\varsigma} + \frac{c_{\ref{error_polyhedron1}} + c_{\ref{error_polyhedron2}}\sqrt{\varsigma}}{(2a)^{1/2}} \right)  T^{-1/2} \log T
  $$

  Now, note that $c_{\ref{const_subopt}}$ is monotonically decreasing in $\beta$. In particular, for $\beta \ge 1$
  $$
  c_{\ref{LyapunovConst}}=\LyapunovConstVal \le (\ell+\mu)R^2 + R\|\nabla_x \bar f(0)\| + n,
  $$
  so that
  \begin{align*}
    c_{\ref{const_subopt}} &= \log n +  2 \log( 1 +\frac{1}{\mu} c_{\ref{LyapunovConst}} ) +  \frac{1}{6}\log 3 + \log 2 \sqrt{\pi} -   \log r_{\min} \\
                           &\le \log n +  2 \log\left( 1 +\frac{(\ell+\mu)R^2 + R\|\nabla_x \bar f(0)\| + n}{\mu}  \right) +  \frac{1}{6}\log 3 + \log 2 \sqrt{\pi} -   \log r_{\min} \\
    &=: c_{\ref{subOptIndep}}
  \end{align*}

  It follows that for $\beta\ge 1$ we have the bound 
    $$
  \bbE[\bar f(\bx^A_T)]\le \min_{x\in\cK} \bar f(x) + \frac{n}{ \beta} \left(\max\{\log\varsigma,0\}+ c_{\ref{subOptIndep}}  \right)
  +c_{\ref{compactLipschitz}}
  \left( c_{\ref{contraction_const1}} + c_{\ref{contraction_const2}} \sqrt{\varsigma} + \frac{c_{\ref{error_polyhedron1}} + c_{\ref{error_polyhedron2}}\sqrt{\varsigma}}{(2a)^{1/2}} \right)  T^{-1/2} \log T
  $$

  Now, picking $\beta$ as in (\ref{eq:betaOpt}) gives
  $$
  \frac{n}{ \beta} \left(\max\{\log\varsigma,0\}+ c_{\ref{subOptIndep}} \right) = \epsilon / 2.
  $$

  Proposition~\ref{prop:mainConstantsBound} implies that there is some constant, $c$ (independent of $\eta$ and $\beta$) such that $c_{\ref{contraction_const1}}, c_{\ref{contraction_const2}},c_{\ref{error_polyhedron1}}, c_{\ref{error_polyhedron2}}\le c e^{\frac{\beta \ell R^2}{2}}$. Furthermore, for all $\beta$ sufficiently large, we have from \eqref{eq:crudeA} that 
  \begin{equation}
  \frac{1}{\sqrt{a}} \le e^{\frac{\beta \ell R^2}{4}}.
\end{equation}
\const{crudeExp}
Thus, for all $\beta$ sufficiently large we have that 
  $$
c_{\ref{compactLipschitz}}
  \left( c_{\ref{contraction_const1}} + c_{\ref{contraction_const2}} \sqrt{\varsigma} + 
   \frac{c_{\ref{error_polyhedron1}} + c_{\ref{error_polyhedron2}}\sqrt{\varsigma}}{(2a)^{1/2}} \right) \le e^{\beta \ell R^2}. 
  $$

  Thus, for our choice of $\beta$ (which is large for sufficiently small $\epsilon$), we have that
  $$
  \bbE[\bar f(\bx^A_T)]\le \min_{x\in\cK} \bar f(x) + \frac{\epsilon}{2} +   e^{\beta \ell R^2} T^{-1/2} \log T
  $$

  For simple notation, let $\alpha$ be such that 
  $$
  \beta\ell R^2 = \frac{\alpha}{\epsilon}.
  $$
  In this case, $\alpha = 2n\left(\max\{\log \varsigma,0\}+c_{\ref{subOptIndep}}\right)\ell R^2$.

  We will choose $T=e^{\gamma/\epsilon}$ and choose $\gamma$ to ensure that
  $$
  e^{\beta \ell R^2} T^{-1/2}\log T = \exp\left(\frac{1}{\epsilon}\left(\alpha - \frac{\gamma}{2}\right)\right) \frac{\gamma}{\epsilon}\le \frac{\epsilon}{2}.
  $$

  The desired inequality holds if and only if:
  $$
\exp\left(\frac{1}{\epsilon}\left(\alpha - \frac{\gamma}{2}\right)\right) \frac{2\gamma}{\epsilon^2}\le 1
  $$
  Note that if $\gamma/2 > \alpha$, then the left side is maximized over $(0,\infty)$ at $\epsilon = \frac{\frac{\gamma}{2}-\alpha}{2}$. Thus, a sufficient condition for this inequality to hold is:
  $$
  \frac{8\gamma e^{-2}}{\left(\frac{\gamma}{2}-\alpha\right)^2}\le 1. 
  $$
  A clean sufficient condition is $T=e^{c_{\ref{TExponent}/\epsilon}}$, where
  $$c_{\ref{TExponent}}:=\gamma = 4\alpha + 32 e^{-2}=
  8n\left(\max\{\log \varsigma,0\}+c_{\ref{subOptIndep}}\right)\ell R^2 + 32 e^{-2}.
  $$
\hfill $\blacksquare$

Now we extend the analysis to the non-compact case. 

\const{optimization_const1}
\const{optimization_const2}
\const{TExpGeneral}
\const{2qmomentremainder1}
\const{2qmomentremainder2}
\begin{proposition} \label{prop: app_optimization}
Let $\bx_k$ be the iterates of the algorithms and assume $\eta \le \frac{\mu}{3\ell^2}$ and $\bbE[\|\bx_0\|^{2q}] < \infty$ for all $q>1$. For all $q>1$, there exist positive constants  $c_{\ref{optimization_const1}}, c_{\ref{optimization_const2}}$ such that for all integers $k \ge 0$, the following bound holds:
\begin{align}
  \label{eq:noncompactGenSuboptimality}
  \bbE[\bar f(\bx_k)] \le \min_{x \in \cK} \bar f(x) + c_{\ref{optimization_const1}} W_1(\cL(\bx_k), \pi_{\beta \bar f}) + c_{\ref{optimization_const2}} W_1 (\cL(\bx_k), \pi_{\beta \bar f})^{\frac{2-2q}{1-2q}} +
\frac{n}{ \beta} \left(\max\{\log\varsigma,0\}+ c_{\ref{const_subopt}} \right)
\end{align}
where
\begin{align*}
c_{\ref{optimization_const1}} &= \|\nabla \bar f(0)
\|
      + \ell \frac{2\left(\|\nabla \bar f(0)\|+\sqrt{\|\nabla \bar f(0)\|} + \sqrt{\mu \bar f(0)}\right)}{\mu} \\
c_{\ref{optimization_const2}} &=  \left( \frac{2 \ell }{\sqrt{\mu}} +\left( \bbE[\|\bx_0\|^{2q}] + c_{\ref{2qmomentremainder2}}\right) \frac{\ell^q 2^{q-1}}{(q-1)} \right) \left( \frac{\ell }{\sqrt{\mu} (q-1)\left( \bbE[\|\bx_0\|^{2q}] + c_{\ref{2qmomentremainder2}} \right) \frac{\ell^q 2^{q-1}}{(q-1)}}\right)^{\frac{2-2q}{-2q+1}}
\end{align*}
and $c_{\ref{2qmomentremainder2}}$ depends on $q$, the statistics of $\bz$, the parameters $\mu$,  $\ell$ and $\nabla \bar{f}(0)$ and decreases monotonically with respect to $\beta$.

Furthermore, there is a constant $c_{\ref{TExpGeneral}}$ such if $\epsilon$ is sufficiently small, $\beta$ is chosen as in (\ref{eq:betaOpt}), and $T=e^{c_{\ref{TExpGeneral}}/\epsilon}$, then
$$
\bbE[\bar f(\bx_T)]\le \min_{x\in\cK}\bar f(x) +\epsilon.
$$
\end{proposition}
\paragraph{Proof of Proposition~{\ref{prop: app_optimization}}}
  Let $\bx$ be drawn according to $\pi_{\beta \bar f}$. Then Lemma~\ref{lem:gibbsSuboptimality} implies:
  \begin{align}
  \label{eq:optimizationBound}
  \nonumber
    \bbE[\bar f(\bx_k)] &= \bbE[\bar f(\bx)] + \bbE[\bar f(\bx_k)-\bar f(\bx)]\\
                          &\le
    \min_{x \in \cK} \bar f(x) + \bbE[\bar f(\bx_k)-\bar f(\bx)] + \frac{n}{ \beta} \left(\max\{\log\varsigma,0\}+ c_{\ref{const_subopt}} \right)
  \end{align}
  So, it now suffices to bound $ \bbE[\bar f(\bx_k)-\bar f(\bx)]$. Ideally, we would bound this term via Kantorovich duality. The problem is that $\bar f$ may not be globally Lipschitz. So, we must approximate it with a Lipschitz function, and then bound the gap induced by this approximation. 

  Namely, fix a constant $m>\bar f(0)$ with $m$ to be chosen later. Set $g(x)=\min\{\bar f(x),m\}$. The inequality from (\ref{eq:quadraticLower}) implies that if $\|x\|\ge \hat R:=\frac{2\left(\|\nabla \bar f(0)\|+\sqrt{\|\nabla \bar f(0)\|+\mu (m-\bar f(0))}\right)}{\mu}$,  then $\bar f(x)\ge m$. We claim that $g$ is globally Lipschitz.

  For $\|x\|\le \hat R$, we have that
  $$
  \|\nabla \bar f(x)\|\le \|\nabla \bar f(0)\| + \ell \hat R=:u.
  $$
  We will show that $g$ is $u$-Lipschitz.

  In the case that $f(y)\ge m$ and $f(x)\ge m$, we have $|g(x)-g(y)|=0$, so the property holds.
  
  Now say that $\bar f(x) < m$ and $\bar f(y) < m$. Then we must have $\|x\|\le \hat R$ and $\|y\|\le \hat R$. Then for all $t\in [0,1]$, we have $\|(1-t)x+ty \|\le \hat R$. It follows that 
  \begin{align*}
    g(x)-g(y)&= 
    \bar f(x)-\bar f(y) \\
             &= \int_0^1 \nabla \bar f(x+t(y-x))^\top (y-x) dt \\
    &\le u \|x-y\|.
  \end{align*}
  Finally, consider the case that $\bar f(x) \ge m$ and $\bar f(y) < m$. Then there is some $\theta \in [0,1]$ such that $\bar f(y + \theta (x-y))=m$. Furthermore
  \begin{align*}
    |g(x)-g(y)| &= m-\bar f(y) \\
                &= \bar f(y+\theta (x-y))-\bar f(y) \\
                &=\int_0^\theta \nabla \bar f( y + t(x-y))^\top (x-y) dt\\
                &\le u \|x-y\|.
  \end{align*}
  It follows that $g$ is $u$-Lipschitz.

  Now noting that $g(x)\le \bar f(x)$ for all $x$ gives
  \begin{align}
    \nonumber
    \bbE[\bar f(\bx_k)-\bar f(\bx)]&\le \bbE[\bar f(\bx_k)-g(\bx)] \\
    \nonumber
                                   &=\bbE[g(\bx_k)-g(\bx)] + \bbE[\indic(\bar f(\bx_k) > m) (\bar f(\bx_k)-m)] \\
    \label{eq:kantorovichSplit}
                                   &\le u W_1(\cL(\bx_k),\pi_{\beta \bar f}) + \bbE[\indic(\bar f(\bx_k) > m) (\bar f(\bx_k)-m)].
  \end{align}
  The final inequality uses Kantorovich duality. Now, it remains to bound
  $
  \bbE[\indic(\bar f(\bx_k) > m) (\bar f(\bx_k)-m)].
  $

  Note that if $\by$ is a non-negative random variable, a standard identity gives that $\bbE[\by]=\int_0^\infty \bbP(\by > \epsilon)d\epsilon$. Thus, we  have
  $$
  \bbE[\indic(\bar f(\bx_k) > m) (\bar f(\bx_k)-m)] = \int_0^\infty \bbP(\bar f(\bx_k) -m > \epsilon)d\epsilon.
  $$

  For all $x\in \cK$, we have
  \begin{align*}
    \bar f(x) &= \bar f(0) - \nabla \bar f(0)^\top x + \int_0^1 (\nabla \bar f(tx)-\nabla \bar f(0))^\top x dt \\
    &\le \bar f(0) +  \|\nabla \bar f(0)\| \|x \| + \frac{1}{2} \ell \|x\|^2 \\
    &\le \bar f(0) + \frac{\|\nabla \bar f(0)\|^2}{2\ell} + \ell \|x\|^2.
  \end{align*}

  So, 
  \begin{align*}
    \bar f(x)-m > \epsilon &\implies  \bar f(0) + \frac{\|\nabla \bar f(0)\|^2}{2\ell} + \ell \|x\|^2 > m+\epsilon \\
    &\iff \|x\|^2 > \frac{m+\epsilon-  \left(\bar f(0) + \frac{\|\nabla \bar f(0)\|^2}{2\ell}\right)}{\ell}.
  \end{align*}

  Now assume that $m/2 > \bar f(0) + \frac{\|\nabla \bar f(0)\|^2}{2\ell}$. Then the right side implies $\|x\|^2 \ge \frac{\frac{m}{2}+\epsilon}{\ell}$.  
  It follows that for any $q>1$, we have, via Markov's inequality and direct computation:
  \begin{align*}
    \bbE[\indic(\bar f(\bx_k) > m) (\bar f(\bx_k)-m)] & \le \int_0^{\infty} \bbP\left(\|\bx_k\|^2 >  \frac{\frac{m}{2}+\epsilon}{\ell} \right) d\epsilon \\
                                                      &= \int_0^{\infty} \bbP\left(\|\bx_k\|^{2q} >  \left(\frac{\frac{m}{2}+\epsilon}{\ell}\right)^q \right) d\epsilon  \\
                                                      &\le \bbE[\|\bx_k\|^{2q}] \int_0^{\infty} \left(\frac{\frac{m}{2}+\epsilon}{\ell}\right)^{-q} d\epsilon \\
    &= \bbE[\|\bx_k\|^{2q}] \frac{\ell^q 2^{q-1}}{(q-1)m^{q-1}}. 
  \end{align*}

  Plugging this expression into (\ref{eq:kantorovichSplit}) and using the definition of $u$ gives
  \begin{align}
  \label{eq:kantorovichSplit2}
    \MoveEqLeft
    \nonumber
    \bbE[\bar f(\bx_k)-\bar f(\bx)] \\
    & \nonumber
      \le \left(\|\nabla \bar f(0)
\|
      + \ell \frac{2\left(\|\nabla \bar f(0)\|+\sqrt{\|\nabla \bar f(0)\|+\mu (m-\bar f(0))}\right)}{\mu} \right) W_1(\cL(\bx_k),\pi_{\beta \bar f}) \\
    &\qquad +  \bbE[\|\bx_k\|^{2q}] \frac{\ell^q 2^{q-1}}{(q-1)m^{q-1}}.
  \end{align}

We want to derive the bound of $\bbE[\|\bx_k\|^{2q}]$.

We have
\begin{align*}
\|\bx_{k+1}\|^{2q} \le \|\bx_k - \eta \nabla f(\bx_k, \bz_k) + \sqrt{\frac{2\eta}{\beta}} \hat{\bw}_k\|^2.
\end{align*}

For notational simplicity, let $\by = \frac{\bx_k - \eta \nabla f(\bx_k, \bz_k)}{\sqrt{2\eta /\beta}}$ and $\bw = \hat{\bw}_k$, then the above inequality can be expressed as 
\begin{align}
\label{eq:binomial_expand} \nonumber
\|\bx_{k+1}\|^{2q} &\le \left( \frac{2 \eta}{\beta}\right)^q\|\by + \bw\|^{2q}\\ \nonumber
&= \left( \frac{2 \eta}{\beta}\right)^q \left( \|\by\|^2 + \|\bw\|^2 +2 \by^\top \bw \right)^q\\ \nonumber
&= \left( \frac{2 \eta}{\beta}\right)^q \sum_{k=0}^q  \binom qk \left( 2 \by^\top \bw \right)^{q-k}\left( \|\by\|^2 + \|\bw\|^2 \right)^k   \\
& = \left( \frac{2 \eta}{\beta}\right)^q \sum_{k=0}^q   \binom qk \left( 2 \by^\top  \bw \right)^{q-k} \sum_{i=0}^k \binom ki\left( \|\by\|^{2i} \|\bw\|^{2(k-i)}    \right).
\end{align}

The last two equalities use the binomial theorem. Here, we construct an orthogonal matrix $U =  \begin{bmatrix}\frac{1}{\|\by\|  } \by^\top \bw\\ s \end{bmatrix}$  such that we can linearly transform the Gaussian noise $\bw$ into $\bv = U\bw =  \begin{bmatrix}\bv_1\\ \bv_2 \end{bmatrix}$, where $\bv_1 = \frac{1}{\|\by\|} \by^\top \bw$ and $\bv_2 = s\bw$. And the orthogonality of the matrix $U$ gives $\bv_1 \perp \bv_2$ and thus $\bv_1^2 + \bv_2^\top \bv_2$ follows a chi-squared distribution with $n$ degrees of freedom. Furthermore, we have $\|\bw\|^2 = \bv_1^2 + \bv_2^\top \bv_2$ and $\by^\top \bw = \|\by\| \bv_1$.

Therefore, with the change of variables, (\ref{eq:binomial_expand}) can be expressed as 
\begin{align*}
\|\bx_{k+1}\|^{2q} \le \left( \frac{2 \eta}{\beta}\right)^q \sum_{k=0}^q   \binom qk \left( 2 \|\by\| \bv_1 \right)^{q-k} \sum_{i=0}^k \binom ki\left( \|\by\|^{2i} (\bv_1^2 + \bv_2^\top \bv_2)^{(k-i)}    \right).
\end{align*}

Taking the expectation of the above inequality gives 
\begin{align}
\label{eq:expectaion_x_2q}
\bbE[\|\bx_{k+1}\|^{2q}] &\le \left( \frac{2 \eta}{\beta}\right)^q \bbE \left[ \sum_{k=0}^q   \binom qk \left( 2 \|\by\|\bv_1 \right)^{q-k} \sum_{i=0}^k \binom ki\left( \|\by\|^{2i} (\bv_1^2 + \bv_2^\top \bv_2)^{(k-i)} \right) \right]\\ \label{eq:simplified_momentBound}
&\le \bbE\left[ \|\bx_k - \eta \nabla f(\bx_k, \bz_k )\|^{2q} \right] + \eta \bbE\left[ p(\|\bx_k - \eta \nabla f(\bx_k, \bz_k ) \|^{2})\right]
\end{align}


where $p(\|\bx_k - \eta \nabla f(\bx_k, \bz_k )\|^2)$ is a polynomial in $\|\bx_k - \eta \nabla f(\bx_k, \bz_k )\|^2$ with order strictly lower than $q$ and the coefficients of $\bbE[p(\|\bx_k - \eta \nabla f(\bx_k, \bz_k )\|^2)]$ depend on the moments of the chi-squared distributions and $q$. (Additionally, note that the coefficients of $p$ can be taken to be monotonically decreasing with respect to $\beta$.) And the reason the polynomial only have even order terms in $\|\bx_k - \eta \nabla f(\bx_k, \bz_k )\|$ is that in (\ref{eq:expectaion_x_2q}), when $q-k$ is odd, the expectation is zero since $\bv_1 \sim \cN(0,1)$ whose odd order moments are all zero. 



Then we firstly aim to bound $\bbE[\|\bx_k - \eta \nabla \bar{f}(\bx_k)\|^{2q}]$.
 
We have 
\begin{align}
\label{eq:x_2q_first_term}
\|\bx_k - \eta \nabla f(\bx_k, \bz_k)\|^{2q} &= \left( \|\bx_k\|^2 - 2 \eta \bx_k^\top \nabla f(\bx_k, \bz_k) + \eta^2 \|\nabla f(\bx_k, \bz_k)\|^2\right)^q.
\end{align}
We examine the second term:
\begin{align*}
\bx_k^\top \nabla f(\bx_k, \bz_k) &= \bx_k^\top (\nabla \bar{f}(\bx_k) - \nabla \bar{f}(0)) + \bx_k^\top( \nabla \bar{f}(0) - \nabla \bar{f}(\bx_k) + \nabla f(\bx_k, \bz_k) ) \\
&\ge \mu \|\bx_k\|^2 - (\ell+ \mu) R^2 + \bx_k^\top(\nabla \bar{f}(0) + \bbE_{\hat \bz}[\nabla f(\bx_k, \bz_k) - \nabla f(\bx_k, \hat \bz_k) ])
\end{align*} 
where the first term is bounded by the assumption of the strong convexity outside a ball and the detailed statement is shown below:

If $\|x\| \ge R$, then $x^\top (\nabla \bar f(x) - \nabla \bar  f(0)) \ge \mu \|x\|^2$.

If $\|x\| \le R$, then $x^\top (\nabla \bar f(x) - \nabla \bar  f(0)) \ge - \ell \|x\|^2 \ge - \ell R^2$.

Therefore, we have for all $x \in \cK$, $x^\top (\nabla \bar f(x) - \nabla \bar f(0)) \ge \mu \|x\|^2 - (\ell+\mu) R^2$.

Note here and below $\hat \bz$ and $\bz$ are IID.

Taking expectation of (\ref{eq:x_2q_first_term}) gives
\begin{align}
\nonumber
& \bbE\left[ \|\bx_k - \eta \nabla f(\bx_k, \bz_k)\|^{2q} \right] \nonumber \\
\nonumber
&= \bbE\left[\left( \|\bx_k\|^2 - 2 \eta \bx_k^\top \nabla f(\bx_k, \bz_k) + \eta^2 \|\nabla f(\bx_k, \bz_k)\|^2\right)^q\right] \\
\nonumber
&\le \bbE\left[\left((1- 2 \mu \eta) \|\bx_k\|^2 + 2 \eta (\ell+ \mu) R^2  \right.\right. \\ \nonumber & \left. \left. \qquad - 2 \eta\bx_k^\top(\nabla \bar{f}(0) + \bbE_{\hat \bz}[\nabla f(\bx_k, \bz_k) - \nabla f(\bx_k, \hat \bz_k) ]) + \eta^2 \|\nabla f(\bx_k, \bz_k)\|^2\right)^q\right] \\
\nonumber
&\le \bbE\left[\left((1- 2 \mu \eta) \|\bx_k\|^2 + 2 \eta (\ell+ \mu) R^2  \right.\right. \\ \nonumber & \left. \left. \qquad - 2 \eta\bx_k^\top(\nabla \bar{f}(0) + \nabla f(\bx_k, \bz_k) - \nabla f(\bx_k, \hat \bz_k) ) + \eta^2 \|\nabla f(\bx_k, \bz_k)\|^2\right)^q\right] \\
 \nonumber
&\le \bbE\left[\left((1- 2 \mu \eta) \|\bx_k\|^2 + 2 \eta (\ell+ \mu) R^2  \right.\right. \\  & \left. \left. \qquad + 2 \eta \|\bx_k\|(\|\nabla \bar{f}(0)\| +  \ell\|\bz_k -  \hat \bz_k\| ) + \eta^2 \|\nabla f(\bx_k, \bz_k)\|^2\right)^q\right]. 
\label{eq:Bound2qmoment}
\end{align}

The second inequality uses Jensen's inequality, and the last inequality uses Cauchy-Schwartz inequality together with $\ell$-Lipschitzness of $\nabla f(x,z)$ in $z$. 

Now we examine the last term of (\ref{eq:Bound2qmoment}).

Firstly, we have
\begin{align*}
\|\nabla f(x,z)\| &= \|\nabla f(x,z) - \bbE_{\hat{\bz}}[\nabla f(x, \hat \bz)] + \bbE_{\hat \bz}[\nabla f(x, \hat \bz)]\| \\
&\le \|\bbE_{\hat{\bz}}[ \nabla f(x,z) - \nabla f(x, \hat \bz)]\| + \| \bar{f}(x)\| \\
&\le \ell \bbE_{\hat{\bz}}[ \|z - \hat \bz\|] + \|\nabla \bar{f}(0)\| + \ell \|x\|.
\end{align*}
So
\begin{align}
\label{eq:fsquareBound}
\|\nabla f(x,z)\|^2 \le 3 \left( \ell^2 \left(\bbE_{\hat{\bz}}[ \|z - \hat \bz\|] \right)^2 + \|\nabla \bar{f}(0)\|^2 + \ell^2 \|x\|^2\right).
\end{align}

Then, we can group the square terms in (\ref{eq:Bound2qmoment}) together and simplify it:
\begin{align*}
&(1-2\mu \eta) \|x\|^2 + \eta^2 3 \ell^2 \|x\|^2 \le (1-\eta \mu) \|x\|^2\\
\iff & 1-2\mu \eta + \eta^2 3 \ell^2 \le 1- \eta \mu \\
 \iff& \eta \le \frac{\mu}{3\ell^2}.
\end{align*}

So, if $\eta \le \frac{\mu}{3\ell^2}$,  plugging (\ref{eq:fsquareBound}) into (\ref{eq:Bound2qmoment}) gives 
\begin{align}
  \nonumber
  \MoveEqLeft
\bbE\left[ \|\bx_k - \eta \nabla f(\bx_k, \bz_k)\|^{2q} \right] \le \bbE\left[\left((1- \mu \eta) \|\bx_k\|^2 + 2 \eta (\ell+ \mu) R^2  \right.\right. \\   & \left. \left. + 2 \eta \|\bx_k\|(\|\nabla \bar{f}(0)\| +  \ell\| \bz_k -  \hat \bz_k \| ) + \eta^2 3 \left( \ell^2 \left(\bbE_{\hat{\bz}}[ \|\bz_k - \hat \bz_k\|] \right)^2 + \|\nabla \bar{f}(0)\|^2 \right) \right)^q\right].
\label{eq:Bound2qmoment2}
\end{align}
We want to further group the first and third terms above together.

 For all $\epsilon \ge 0$, $2ab = 2 (\epsilon a) (\frac{1}{\epsilon} b) \le (\epsilon a)^2 + (\frac{1}{\epsilon}b)^2$ . Let $a= \|\bx_k\|$, $b = \|\nabla \bar{f}(0)\| +  \ell \|\bz_k -  \hat \bz_k\| $, then we can see the third term of the right  side of (\ref{eq:Bound2qmoment2}) can be upper bounded by a summation of two parts. The first part can be grouped with the first term of the right side of (\ref{eq:Bound2qmoment2}):
\begin{align*}
&(1- \mu \eta) \|\bx_k\|^2 + \eta \epsilon^2 \|\bx_k\|^2
\le (1- \frac{\mu \eta}{2}) \|\bx_k\|^2 \\
\iff & 1- \mu \eta +\eta \epsilon^2 \le 1- \frac{\mu \eta}{2} \\
\iff & \epsilon \le \sqrt{\frac{\mu}{2}}.
\end{align*}

So let $\epsilon = \sqrt{\frac{\mu}{2}}$, we have
\begin{align}
\nonumber
  \bbE\left[ \|\bx_k - \eta \nabla f(\bx_k, \bz_k)\|^{2q} \right] &\le \bbE\left[\left((1- \frac{\mu \eta}{2}) \|\bx_k\|^2 + 2 \eta (\ell+ \mu) R^2  \right.\right. \\
  \nonumber & \hspace{-40pt}                                                                                                                        \left. \left. + \frac{2\eta}{\mu}(\|\nabla \bar{f}(0)\| +  \ell \| \bz_k -  \hat \bz_k \| )^2 + \eta^2 3 \left( \ell^2 \left(\bbE_{\hat{\bz}}[ \|\bz_k - \hat \bz_k\|] \right)^2 + \|\nabla \bar{f}(0)\|^2 \right) \right)^q\right]  \\ \nonumber 
                                                                  &\le \bbE\left[\left((1- \frac{\mu \eta}{2}) \|\bx_k\|^2 + 2 \eta (\ell+ \mu) R^2  \right.\right. \\ \label{eq:second_inequality}
                                                                  &
                                                                    \hspace{-30pt}
                                                                    \left. \left. + \frac{2\eta}{\mu}(\|\nabla \bar{f}(0)\| +  \ell\| \bz_k -  \hat \bz_k \| )^2 + \eta^2 3 \left( \ell^2 \left( \|\bz_k - \hat \bz_k\| \right)^2 +\|\nabla \bar{f}(0)\|^2 \right) \right)^q\right] \\ \nonumber
& = (1-\frac{\mu \eta}{2})^q \bbE[\|\bx_k\|^{2q}] + \eta \bbE[p_2(\|\bx_k\|^2, \|\bz_k -\hat{\bz}_k\|)]\\ 
&\le (1-\frac{\mu \eta}{2}) \bbE[\|\bx_k\|^{2q}] + \eta \bbE[p_2(\|\bx_k\|^2, \|\bz_k -\hat{\bz}_k\|)].
\label{eq:y2q_upperbound}
 \end{align}

The inequality (\ref{eq:second_inequality}) uses Jensen's inequality twice. The polynomial $p_2(\|\bx_k\|^2, \|\bz_k -\hat{\bz}_k\|)$ is with order strictly lower than $q$ in $\|\bx_k\|^2$ and with the highest order of $2q$ in $\|\bz_k - \hat{\bz}_k\|$.

Similarly, we can obtain for all $i <q$,
\begin{align*}
  \MoveEqLeft
\bbE\left[ \|\bx_k - \eta \nabla f(\bx_k, \bz_k)\|^{2i} \right] \le 
\bbE\left[ \left((1- \frac{\mu \eta}{2}) \|\bx_k\|^2 + 2 \eta (\ell+ \mu) R^2  \right. \right. \\   & \left. \left. + \frac{2\eta}{\mu}(\|\nabla \bar{f}(0)\| +  \ell\| \bz_k -  \hat \bz_k \| )^2 + \eta^2 3 \left( \ell^2 \left( \|\bz_k - \hat \bz_k\| \right)^2 +\|\nabla \bar{f}(0)\|^2 \right)\right)^i \right].
\end{align*}
This implies that $ \bbE[p(\|\bx_k - \eta \nabla f(\bx_k, \bz_k ) \|^{2})]$ can be upper bounded by $\bbE[p_1(\|\bx_k\|^2, \|\bz_k - \hat{\bz}_k\|)]$ where $p_1(\|\bx_k\|^2, \|\bz_k - \hat{\bz}_k\|)$ is a polynomial with the order strictly lower than q in $\|\bx_k\|^2$ and the highest order of $2q-2$ in $\|\bz_k - \hat{\bz}_k \|$. 

So 
(\ref{eq:simplified_momentBound}) can be further upper bounded as below: 
\begin{align}
\nonumber
 \bbE[\|\bx_{k+1}\|^{2q}] &\le (1-\frac{\mu \eta}{2}) \bbE[\|\bx_k\|^{2q}] + \eta \bbE[p_2(\|\bx_k\|^2, \|\bz_k -\hat{\bz}_k\|)] + \eta \bbE[p_1(\|\bx_k\|^2, \|\bz_k -\hat{\bz}_k\|)] \\
 \nonumber
 &= (1-\frac{\mu \eta}{2}) \bbE[\|\bx_k\|^{2q}] + \eta \bbE[p_3(\|\bx_k\|^2, \|\bz_k -\hat{\bz}_k\|)] \\
 \nonumber
 & \le (1- \frac{\mu \eta}{4}) \bbE[\|\bx_k\|^{2q}] + \eta \bbE[-\frac{\mu}{4} \|\bx_k\|^{2q} + p_3(\|\bx_k\|^2, \|\bz_k -\hat{\bz}_k\|)] \\
 \label{eq:x_2q_momentBound}
  & \le (1- \frac{\mu \eta}{4}) \bbE[\|\bx_k\|^{2q}] + \eta \bbE[\frac{\mu}{4} \left( - \|\bx_k\|^{2q} +\tilde{p}(\|\bx_k\|^2, \|\bz_k -\hat{\bz}_k\|) ] \right)
\end{align}

To get the upper bound of the second term of (\ref{eq:x_2q_momentBound}), we examine the following polynomial with $x \ge 0$
\begin{align*}
- x^{q} + \sum_{i=0}^{q-1} a_{q,i}x^{i},
\end{align*}
where the $a_{q,i}$'s depend on the value of $q$, the statistics of the external random variables $\bz$ and some other parameters including $\ell$, $\mu$ and $\|\nabla \bar{f}(0)\|$ and $a_{q,i}$'s decrease monotonically with respect to $\beta$.

To find the upper bound of such a polynomial, we consider two cases
\begin{itemize}
\item Assume $0 \le  x \le 1$, then $- x^{q} + \sum_{i=0}^{q-1} a_{q,i}x^{i} \le \sum_{i=0}^{q-1} |a_{q,i}| $; 
\item Assume $x >1$, then $- x^{q} + \sum_{i=0}^{q-1} a_{q,i} x^{i} \le \left( \sum_{i=0}^{q-1} |a_{q,i}| \right) \left( \sum_{i=0}^{q-1} |a_{q,i}| + 1 \right)^{q-1} $.
\end{itemize}

Combining the two cases gives that for all $x \ge 0$,
\begin{equation}
\nonumber
- x^{q} + \sum_{i=0}^{q-1} a_{q,i} x^{i} \le \left( \sum_{i=0}^{q-1} |a_{q,i}| \right) \left( \sum_{i=0}^{q-1} |a_{q,i}| + 1 \right)^{q-1}.
\end{equation}
The first case is a direct result of dropping the negative term and using Cauchy-Schwartz inequality. The second case is obtained by firstly showing the sufficient condition of the polynomial being non-positive. The detail is shown below:
\begin{align}
- x^{q} + \sum_{i=0}^{q-1} a_{q,i}x^{i} \le 0 
&\iff -1 + \sum_{i=0}^{q-1} \frac{a_{q,i}}{x^{q-i}} \le 0 \nonumber \\
& \impliedby -1 + \sum_{i=0}^{q-1} \frac{|a_{q,i}|}{x} \le 0  \label{eq:poly_sufficient_1}\\ \nonumber
& \iff -1 + \frac{1}{x} \sum_{i=0}^{q-1} |a_{q,i}| \le 0 \\
& \iff x \ge \max \{ \sum_{i=0}^{q-1} |a_{q,i}|, 1 \} \label{eq:poly_sufficient_2} \\ \nonumber
&\impliedby x \ge  \sum_{i=0}^{q-1} |a_{q,i}| + 1
\end{align}

Both (\ref{eq:poly_sufficient_1}) and (\ref{eq:poly_sufficient_2}) use the assumption that $x>1$.

Besides, for $1<x \le \sum_{i=0}^{q-1} |a_{q,i}| + 1 $,
\begin{align*}
- x^{q} + \sum_{i=0}^{q-1} a_{q,i} x^{i} &\le \sum_{i=0}^{q-1} |a_{q,i}| x^i \\
&\le \sum_{i=0}^{q-1} |a_{q,i}| x_{max}^{q-1}\\
&= \sum_{i=0}^{q-1} |a_{q,i}| \left( \sum_{i=0}^{q-1} |a_{q,i}|+1 \right)^{q-1}.
\end{align*}

Therefore, we can conclude that 
\begin{align*}
\bbE[-\frac{\mu}{4} \|\bx_k\|^{2q} + \tilde{p} (\|\bx_k\|^2, \|\bz_k -\hat{\bz}_k\|)] \le
\bbE \left[\frac{\mu}{4}  \sum_{i=0}^{q-1} |a_{q,i}| \left( \sum_{i=0}^{q-1} |a_{q,i}|+1 \right)^{q-1}  \right].
\end{align*}

The L-mixing property ensures that the right side of the inequality is bounded.  Then,  we achieve the upper bound of equation (\ref{eq:x_2q_momentBound}). 
\begin{align*}
\bbE[\|\bx_{k+1}\|^{2q}] &\le (1- \frac{\mu \eta}{4}) \bbE[\|\bx_k\|^{2q}] + \eta \bbE \left[\frac{\mu}{4}  \sum_{i=0}^{q-1} |a_{q,i}| \left( \sum_{i=0}^{q-1} |a_{q,i}|+1 \right)^{q-1} \right]
\end{align*}

Iterating the inequality above and letting 
$ \tilde{a}_q = \bbE \left[\frac{\mu}{4}  \sum_{i=0}^{q-1} |a_{q,i}| \left( \sum_{i=0}^{q-1} |a_{q,i}|+1 \right)^{q-1}  \right]$ give
\begin{align*}
\bbE[\|\bx_k\|^q] &\le \left( 1- \frac{\mu \eta}{4} \right)^k \bbE[\|\bx_0\|^{2q}] + \eta \tilde{a}_q \sum_{i=0}^{k-1}(1-\frac{\mu \eta}{4})^i \\
&\le \bbE[\|\bx_0\|^{2q}] + \eta \tilde{a}_q \frac{1- \left( 1- \frac{\mu \eta}{4}\right)^k}{1-\left( 1- \frac{\mu \eta}{4}\right) } \\
&\le \bbE[\|\bx_0\|^{2q}] + \frac{4}{\mu} \tilde{a}_q \left( 1-\left(1- \frac{\mu \eta}{4} \right)^k \right) \\
& \le \bbE[\|\bx_0\|^{2q}] + \frac{4}{\mu} \tilde{a}_q .
\end{align*}


Now as long as $\bbE[\|\bx_0\|^{2q}] < \infty$ and $\eta <1$, we have
\begin{align*}
\bbE[\|\bx_k\|^{2q}] \le \bbE[\|\bx_0\|^{2q}] + c_{\ref{2qmomentremainder2}},
\end{align*}
where $c_{\ref{2qmomentremainder2}} = \frac{4}{\mu} \tilde{a}_q$. More specifically, $c_{\ref{2qmomentremainder2}}$ depends on $q$, the statistics of $\bz$, the parameters $\mu$,  $\ell$ and $\nabla \bar{f}(0)$.

Plugging the above result into (\ref{eq:kantorovichSplit2}) gives 
  \begin{align*}
  \label{eq:kantorovichSplit3}
    \MoveEqLeft
    \bbE[\bar f(\bx_k)-\bar f(\bx)] \\
    &
      \le \left(\|\nabla \bar f(0)
\|
      + \ell \frac{2\left(\|\nabla \bar f(0)\|+\sqrt{\|\nabla \bar f(0)\|+\mu (m-\bar f(0))}\right)}{\mu} \right) W_1(\cL(\bx_k),\pi_{\beta \bar f}) \\
    &\qquad +  \left( \bbE[\|\bx_0\|^{2q}] + c_{\ref{2qmomentremainder2}} \right) \frac{\ell^q 2^{q-1}}{(q-1)m^{q-1}}.
  \end{align*}

The remaining work is to optimize the right side of the above inequality with respect to $m$ so that we can make a choice of the value of $m$ mentioned earlier in the proof.

Let 
\begin{align*}
g(m)&=\left(\|\nabla \bar f(0)
\|
      + \ell \frac{2\left(\|\nabla \bar f(0)\|+\sqrt{\|\nabla \bar f(0)\|} + \sqrt{\mu m} + \sqrt{\mu \bar f(0)}\right)}{\mu} \right) W_1(\cL(\bx_k),\pi_{\beta \bar f}) \\
    &\qquad +  \left( \bbE[\|\bx_0\|^{2q}] + c_{\ref{2qmomentremainder2}} \right) \frac{\ell^q 2^{q-1}}{(q-1)m^{q-1}}.
  \end{align*}
We can see that $g(m)$ is an upper bound of the right side of (\ref{eq:kantorovichSplit2}). 
  
Setting $g^\prime(m) = 0$ leads to $m^* = \left( \frac{\ell W_1}{\sqrt{\mu} (q-1)C}\right)^{\frac{2}{-2q+1}}$, where $C = \left(\bbE[\|\bx_0\|^{2q}] + c_{\ref{2qmomentremainder2}} \right) \frac{\ell^q 2^{q-1}}{(q-1)}$ for notation simplicity. 
So 
\begin{align*}
\max_{m \ge 0}g(m) &= g(m^*)\\
&\le \left(\|\nabla \bar f(0)
\|
      + \ell \frac{2\left(\|\nabla \bar f(0)\|+\sqrt{\|\nabla \bar f(0)\|} + \sqrt{\mu \bar f(0)}\right)}{\mu} \right) W_1(\cL(\bx_k),\pi_{\beta \bar f}) \\
      & + \frac{2 \ell }{\sqrt{\mu}} \left( \frac{\ell W_1(\cL(\bx_k),\pi_{\beta \bar f})}{\sqrt{\mu} (q-1)C}\right)^{\frac{2-2q}{-2q+1}}
       +  C \left( \frac{\ell W_1(\cL(\bx_k),\pi_{\beta \bar f})}{\sqrt{\mu} (q-1)C}\right)^{\frac{2-2q}{-2q+1}}\\
       & =  \left(\|\nabla \bar f(0)
\|
      + \ell \frac{2\left(\|\nabla \bar f(0)\|+\sqrt{\|\nabla \bar f(0)\|} + \sqrt{\mu \bar f(0)}\right)}{\mu} \right) W_1(\cL(\bx_k),\pi_{\beta \bar f}) \\
      & + \left( \frac{2 \ell }{\sqrt{\mu}} +C \right) \left( \frac{\ell }{\sqrt{\mu} (q-1)C}\right)^{\frac{2-2q}{-2q+1}}
       W_1(\cL(\bx_k),\pi_{\beta \bar f})^{\frac{2-2q}{-2q+1}}
\end{align*}

Setting 
\begin{align*}
c_{\ref{optimization_const1}} &= \|\nabla \bar f(0)
\|
      + \ell \frac{2\left(\|\nabla \bar f(0)\|+\sqrt{\|\nabla \bar f(0)\|} + \sqrt{\mu \bar f(0)}\right)}{\mu} \\
c_{\ref{optimization_const2}} &=  \left( \frac{2 \ell }{\sqrt{\mu}} +\left( \bbE[\|\bx_0\|^{2q}] + c_{\ref{2qmomentremainder2}}\right) \frac{\ell^q 2^{q-1}}{(q-1)} \right) \left( \frac{\ell }{\sqrt{\mu} (q-1)\left( \bbE[\|\bx_0\|^{2q}] + c_{\ref{2qmomentremainder2}} \right) \frac{\ell^q 2^{q-1}}{(q-1)}}\right)^{\frac{2-2q}{-2q+1}}.
\end{align*}
and plugging this bound into (\ref{eq:optimizationBound}) give the suboptimality bound from \eqref{eq:noncompactGenSuboptimality}.

In particular, if $q=4$, $\beta \ge 1$ and $W_1(\cL(\bx_k),\pi_{\beta\bar f})\le 1$
we get a bound of the form:
$$
\bbE[\bar f(\bx_k)] \le \min_{x \in \cK} \bar f(x) + c W_1 (\cL(\bx_k), \pi_{\beta \bar f})^{\frac{2}{3}} +
\frac{n}{ \beta} \left(\max\{\log\varsigma,0\}+ c_{\ref{const_subopt}} \right)
$$
for some constant $c$ independent of $\beta$.

Indeed, $c_{\ref{2qmomentremainder2}}$ decreases monotonically with respect to $\beta$, and thus so does $c_{\ref{optimization_const2}}$. So, assuming $\beta\ge 1$, we can take $c\ge c_{\ref{optimization_const1}}+c_{\ref{optimization_const2}}$ to be a fixed value independent of $\beta$.

Setting $\beta$ as in (\ref{eq:betaOpt}) gives
$$
\bbE[\bar f(\bx_k)] \le \min_{x \in \cK} \bar f(x) + c W_1 (\cL(\bx_k), \pi_{\beta \bar f})^{\frac{2}{3}} +
\frac{\epsilon}{2}
$$

Then, arguing as in the proof of Proposition~\ref{prop:suboptCompact}, for sufficiently large $\beta$ and $T\ge 4$, we have that
\begin{align*}
  c W_1 (\cL(\bx_k), \pi_{\beta \bar f})^{\frac{2}{3}}
  &
    \le c  \left( c_{\ref{contraction_const1}} + c_{\ref{contraction_const2}} \sqrt{\varsigma} + \frac{c_{\ref{error_polyhedron1}} + c_{\ref{error_polyhedron2}}\sqrt{\varsigma}}{(2a)^{1/2}} \right)^{2/3}  T^{-1/3} \log T \\
  &\le e^{\frac{2\beta\ell R^2}{3}} T^{-1/3}\log T
\end{align*}

Then setting, $\alpha = 2n\left(\max\{\log \varsigma,0\}+c_{\ref{subOptIndep}}\right)\ell R^2$, $\beta$ from (\ref{eq:betaOpt}), and $T=e^{\gamma/\epsilon}$ gives
$$
c W_1 (\cL(\bx_k), \pi_{\beta \bar f})^{\frac{2}{3}}\le \exp\left(
  \frac{1}\epsilon\left(\frac{2\alpha}{3}-\frac{\gamma}{3} \right)
\right) \frac{\gamma}{\epsilon}
$$
So, we seek a sufficient condition for
$$
\exp\left(
  \frac{1}\epsilon\left(\frac{2\alpha}{3}-\frac{\gamma}{3} \right)
\right) \frac{\gamma}{\epsilon}\le \frac{\epsilon}{2} \iff 
\exp\left(
  \frac{1}\epsilon\left(\frac{2\alpha}{3}-\frac{\gamma}{3} \right)
\right) \frac{2\gamma}{\epsilon^2}\le 1.
$$
Then, similar to the compact case, we have that when $\gamma > 2\alpha$,  the left side is maximized over $(0,\infty)$ at $\epsilon = \frac{\gamma - 2\alpha}{6}$.  Plugging in the maximizer gives the sufficient condition:
$$
\frac{72 e^{-2} \gamma}{(\gamma-2\alpha)^2}\le 1
$$
This is satisfied in particular at
$$
c_{\ref{TExpGeneral}}=\gamma = 8\alpha + 72 e^{-2}. 
$$
\hfill $\blacksquare$


\end{document}